\documentclass{article}
\usepackage{iclr2027_conference}
\iclrfinalcopy
\usepackage{fontspec}
\usepackage{amsmath,amssymb,booktabs,tabularx,array}
\usepackage{xcolor,graphicx,microtype,url}
\usepackage{tikz}
\usetikzlibrary{arrows.meta,calc,positioning}
\usepackage[colorlinks=true,allcolors=blue]{hyperref}
\usepackage{xurl}
\newcommand{\slashbreaks}{\Urlmuskip=0mu\relax\def\UrlBreaks{\do\/}\def\UrlBigBreaks{}%
  \def\UrlOrds{\do\*\do\-\do\~\do\'\do\"\do\.\do\@\do\\\do\!\do\_\do\|\do\;\do\>\do\]%
    \do\)\do\,\do\?\do\&\do+\do\=\do\#\do\:}}
\newcommand{\code}[1]{\texttt{\slashbreaks\nolinkurl{#1}}}
\DeclareUrlCommand\path{\urlstyle{tt}\slashbreaks}
\newcommand{\N}[1]{\ifcsname papernum#1\endcsname\ifmmode\begingroup\mathcode`\,="013B\relax\csname papernum#1\endcsname\endgroup\else\csname papernum#1\endcsname\fi\else\PackageError{numbers}{Undefined number #1}{numbers.tex defines every number}\fi}
\expandafter\def\csname papernumLK_all_zero\endcsname{151}
\expandafter\def\csname papernumLK_all_zero_changed_entries\endcsname{2,473,984}
\expandafter\def\csname papernumLK_changed_entries\endcsname{126,696,155}
\expandafter\def\csname papernumLK_forgiven\endcsname{9,982}
\expandafter\def\csname papernumLK_forgiven_share\endcsname{79.37}
\expandafter\def\csname papernumLK_mixed\endcsname{9,831}
\expandafter\def\csname papernumLK_mixed_changed_entries\endcsname{124,222,171}
\expandafter\def\csname papernumLK_mixed_entry_percent\endcsname{1.0487}
\expandafter\def\csname papernumLK_mixed_max\endcsname{16,383}
\expandafter\def\csname papernumLK_mixed_median\endcsname{14,323}
\expandafter\def\csname papernumLK_mixed_median_percent\endcsname{87.42}
\expandafter\def\csname papernumLK_mixed_min\endcsname{306}
\expandafter\def\csname papernumLK_mixed_percent\endcsname{9.52}
\expandafter\def\csname papernumLK_pooled_median\endcsname{14,391}
\expandafter\def\csname papernumP1_count\endcsname{12,576}
\expandafter\def\csname papernumcache_residuals\endcsname{8}
\expandafter\def\csname papernumcontrol_pool_comparisons\endcsname{540}
\expandafter\def\csname papernumdiv_denominator\endcsname{103,288}
\expandafter\def\csname papernumdiv_tokens\endcsname{0}
\expandafter\def\csname papernumdiv_worst_deviation\endcsname{5.53}
\expandafter\def\csname papernumdiv_worst_step\endcsname{2.27}
\expandafter\def\csname papernumexact_DDC\endcsname{13,319,450}
\expandafter\def\csname papernumexact_NC\endcsname{158,216}
\expandafter\def\csname papernumexact_TLC\endcsname{29,606,430}
\expandafter\def\csname papernumexact_TTC\endcsname{268,967}
\expandafter\def\csname papernumexposure_mean_probability\endcsname{0.998323}
\expandafter\def\csname papernumexposure_p_one\endcsname{11,084}
\expandafter\def\csname papernumexposure_population\endcsname{11,434}
\expandafter\def\csname papernumflip_T_dac_fail_percent\endcsname{65.7}
\expandafter\def\csname papernumflip_T_devC_gt_human_percent\endcsname{84.7}
\expandafter\def\csname papernumflip_T_devC_median_m\endcsname{6.09}
\expandafter\def\csname papernumflip_T_human_bin_a_percent\endcsname{31.6}
\expandafter\def\csname papernumflip_T_human_bin_b_percent\endcsname{27.8}
\expandafter\def\csname papernumflip_T_human_bin_c_percent\endcsname{31.5}
\expandafter\def\csname papernumflip_T_human_bin_d_percent\endcsname{9.1}
\expandafter\def\csname papernumflip_T_human_devC_median_m\endcsname{1.43}
\expandafter\def\csname papernumflip_T_lkH_dac_fail_percent\endcsname{53.0}
\expandafter\def\csname papernumflip_T_lkH_fail_percent\endcsname{64.0}
\expandafter\def\csname papernumflip_T_other_fail_percent\endcsname{82.4}
\expandafter\def\csname papernumflip_T_runC_median_steps\endcsname{33}
\expandafter\def\csname papernumflip_T_within_half_m_percent\endcsname{9.1}
\expandafter\def\csname papernumflip_U_dac_fail_percent\endcsname{78.1}
\expandafter\def\csname papernumflip_U_devC_median_m\endcsname{7.96}
\expandafter\def\csname papernumflip_U_fail_entries\endcsname{481,727,638}
\expandafter\def\csname papernumflip_U_lkH_fail_percent\endcsname{88.1}
\expandafter\def\csname papernumflip_U_other_fail_percent\endcsname{91.8}
\expandafter\def\csname papernumflip_U_runC_median_steps\endcsname{26}
\expandafter\def\csname papernumflip_U_within_half_m_percent\endcsname{1.3}
\expandafter\def\csname papernumflip_human_bin_a_m\endcsname{1.0}
\expandafter\def\csname papernumflip_human_bin_b_m\endcsname{1.75}
\expandafter\def\csname papernumflip_human_bin_c_m\endcsname{3.5}
\expandafter\def\csname papernumflip_lk_limit_m\endcsname{0.5}
\expandafter\def\csname papernumflip_lk_steps_required\endcsname{20}
\expandafter\def\csname papernumflip_rollout_candidates\endcsname{1,574,322,176}
\expandafter\def\csname papernumflip_rollout_steps\endcsname{41}
\expandafter\def\csname papernumflip_rome_tokens\endcsname{67,035}
\expandafter\def\csname papernumflip_step_s\endcsname{0.1}
\expandafter\def\csname papernumforgiven_token_columns\endcsname{16,009}
\expandafter\def\csname papernumheadline_count\endcsname{11,237}
\expandafter\def\csname papernumheadline_percent\endcsname{10.8793}
\expandafter\def\csname papernumhmdp_nt_R1\endcsname{0.6627}
\expandafter\def\csname papernumhmdp_nt_R1_hi\endcsname{0.6878}
\expandafter\def\csname papernumhmdp_nt_R1_lo\endcsname{0.6276}
\expandafter\def\csname papernumhmdp_nt_R2\endcsname{0.9761}
\expandafter\def\csname papernumhmdp_nt_R2_hi\endcsname{0.9784}
\expandafter\def\csname papernumhmdp_nt_R2_lo\endcsname{0.9732}
\expandafter\def\csname papernumhmdp_nt_R3\endcsname{0.2571}
\expandafter\def\csname papernumhmdp_nt_R3_hi\endcsname{0.2829}
\expandafter\def\csname papernumhmdp_nt_R3_lo\endcsname{0.2362}
\expandafter\def\csname papernumhmdp_tr_R1\endcsname{0.6060}
\expandafter\def\csname papernumhmdp_tr_R1_hi\endcsname{0.6144}
\expandafter\def\csname papernumhmdp_tr_R1_lo\endcsname{0.5958}
\expandafter\def\csname papernumhmdp_tr_R2\endcsname{0.9891}
\expandafter\def\csname papernumhmdp_tr_R2_hi\endcsname{0.9895}
\expandafter\def\csname papernumhmdp_tr_R2_lo\endcsname{0.9888}
\expandafter\def\csname papernumhmdp_tr_R3\endcsname{0.3310}
\expandafter\def\csname papernumhmdp_tr_R3_hi\endcsname{0.3390}
\expandafter\def\csname papernumhmdp_tr_R3_lo\endcsname{0.3241}
\expandafter\def\csname papernumnt_A0_R1\endcsname{0.7150}
\expandafter\def\csname papernumnt_A0_R2\endcsname{0.9781}
\expandafter\def\csname papernumnt_A0_R3\endcsname{0.2268}
\expandafter\def\csname papernumnt_A0_R4\endcsname{0.2327}
\expandafter\def\csname papernumnt_A1_R1\endcsname{0.7184}
\expandafter\def\csname papernumnt_A1_R2\endcsname{0.9813}
\expandafter\def\csname papernumnt_A1_R3\endcsname{0.2249}
\expandafter\def\csname papernumnt_A1_R4\endcsname{0.2285}
\expandafter\def\csname papernumnt_A2_R1\endcsname{0.7105}
\expandafter\def\csname papernumnt_A2_R2\endcsname{0.9788}
\expandafter\def\csname papernumnt_A2_R3\endcsname{0.2341}
\expandafter\def\csname papernumnt_A2_R4\endcsname{0.2380}
\expandafter\def\csname papernumnt_A_R3_max\endcsname{0.2341}
\expandafter\def\csname papernumnt_A_R3_min\endcsname{0.2249}
\expandafter\def\csname papernumnt_L0_R1\endcsname{0.4778}
\expandafter\def\csname papernumnt_L0_R2\endcsname{0.2426}
\expandafter\def\csname papernumnt_L0_R3\endcsname{-0.2291}
\expandafter\def\csname papernumnt_L0_R4\endcsname{-0.2192}
\expandafter\def\csname papernumnt_M0_R1\endcsname{0.7308}
\expandafter\def\csname papernumnt_M0_R2\endcsname{0.9815}
\expandafter\def\csname papernumnt_M0_R3\endcsname{0.2126}
\expandafter\def\csname papernumnt_M0_R4\endcsname{0.2190}
\expandafter\def\csname papernumnt_M1_R1\endcsname{0.7293}
\expandafter\def\csname papernumnt_M1_R2\endcsname{0.9833}
\expandafter\def\csname papernumnt_M1_R3\endcsname{0.2138}
\expandafter\def\csname papernumnt_M1_R4\endcsname{0.2180}
\expandafter\def\csname papernumnt_M2_R1\endcsname{0.7204}
\expandafter\def\csname papernumnt_M2_R2\endcsname{0.9818}
\expandafter\def\csname papernumnt_M2_R3\endcsname{0.2170}
\expandafter\def\csname papernumnt_M2_R4\endcsname{0.2249}
\expandafter\def\csname papernumnt_P0_R1\endcsname{0.8236}
\expandafter\def\csname papernumnt_P0_R2\endcsname{0.9834}
\expandafter\def\csname papernumnt_P0_R3\endcsname{0.1399}
\expandafter\def\csname papernumnt_P0_R4\endcsname{0.1421}
\expandafter\def\csname papernumnt_P1_R1\endcsname{0.8397}
\expandafter\def\csname papernumnt_P1_R2\endcsname{0.9850}
\expandafter\def\csname papernumnt_P1_R3\endcsname{0.1234}
\expandafter\def\csname papernumnt_P1_R4\endcsname{0.1327}
\expandafter\def\csname papernumnt_P2_R1\endcsname{0.8356}
\expandafter\def\csname papernumnt_P2_R2\endcsname{0.9830}
\expandafter\def\csname papernumnt_P2_R3\endcsname{0.1275}
\expandafter\def\csname papernumnt_P2_R4\endcsname{0.1323}
\expandafter\def\csname papernumnt_PA_R1_max\endcsname{0.1251}
\expandafter\def\csname papernumnt_PA_R1_min\endcsname{0.1086}
\expandafter\def\csname papernumnt_P_R3_max\endcsname{0.1399}
\expandafter\def\csname papernumnt_P_R3_min\endcsname{0.1234}
\expandafter\def\csname papernumnt_T\endcsname{1,496}
\expandafter\def\csname papernumnt_pairs\endcsname{1,287}
\expandafter\def\csname papernumnt_released_R1\endcsname{0.7095}
\expandafter\def\csname papernumnt_released_R1_hi\endcsname{0.7241}
\expandafter\def\csname papernumnt_released_R1_lo\endcsname{0.6847}
\expandafter\def\csname papernumnt_released_R2\endcsname{0.9807}
\expandafter\def\csname papernumnt_released_R2_hi\endcsname{0.9820}
\expandafter\def\csname papernumnt_released_R2_lo\endcsname{0.9780}
\expandafter\def\csname papernumnt_released_R3\endcsname{0.2368}
\expandafter\def\csname papernumnt_released_R3_hi\endcsname{0.2544}
\expandafter\def\csname papernumnt_released_R3_lo\endcsname{0.2206}
\expandafter\def\csname papernumnt_released_R4\endcsname{0.2441}
\expandafter\def\csname papernumob_default_differing_max\endcsname{77.7}
\expandafter\def\csname papernumob_default_differing_median\endcsname{51.5}
\expandafter\def\csname papernumob_allpin_nodispatch_tokens\endcsname{4,992}
\expandafter\def\csname papernumob_pin_reproduced\endcsname{4,991}
\expandafter\def\csname papernumob_pin_tokens\endcsname{4,992}
\expandafter\def\csname papernumob_skx_tokens\endcsname{260}
\expandafter\def\csname papernumother_changed_entries_exact\endcsname{43,353,063}
\expandafter\def\csname papernump530nt_A0_R1\endcsname{0.7208}
\expandafter\def\csname papernump530nt_A0_R2\endcsname{0.9815}
\expandafter\def\csname papernump530nt_A0_R3\endcsname{0.2171}
\expandafter\def\csname papernump530nt_A0_R4\endcsname{0.2277}
\expandafter\def\csname papernump530nt_A1_R1\endcsname{0.7302}
\expandafter\def\csname papernump530nt_A1_R2\endcsname{0.9801}
\expandafter\def\csname papernump530nt_A1_R3\endcsname{0.2245}
\expandafter\def\csname papernump530nt_A1_R4\endcsname{0.2267}
\expandafter\def\csname papernump530nt_A2_R1\endcsname{0.7101}
\expandafter\def\csname papernump530nt_A2_R2\endcsname{0.9814}
\expandafter\def\csname papernump530nt_A2_R3\endcsname{0.2271}
\expandafter\def\csname papernump530nt_A2_R4\endcsname{0.2372}
\expandafter\def\csname papernump530nt_A_R3_max\endcsname{0.2271}
\expandafter\def\csname papernump530nt_A_R3_min\endcsname{0.2171}
\expandafter\def\csname papernump530nt_M0_R1\endcsname{0.7200}
\expandafter\def\csname papernump530nt_M0_R2\endcsname{0.9833}
\expandafter\def\csname papernump530nt_M0_R3\endcsname{0.2145}
\expandafter\def\csname papernump530nt_M0_R4\endcsname{0.2224}
\expandafter\def\csname papernump530nt_M1_R1\endcsname{0.7331}
\expandafter\def\csname papernump530nt_M1_R2\endcsname{0.9832}
\expandafter\def\csname papernump530nt_M1_R3\endcsname{0.2151}
\expandafter\def\csname papernump530nt_M1_R4\endcsname{0.2157}
\expandafter\def\csname papernump530nt_M2_R1\endcsname{0.7287}
\expandafter\def\csname papernump530nt_M2_R2\endcsname{0.9831}
\expandafter\def\csname papernump530nt_M2_R3\endcsname{0.2114}
\expandafter\def\csname papernump530nt_M2_R4\endcsname{0.2196}
\expandafter\def\csname papernump530nt_P0_R1\endcsname{0.8531}
\expandafter\def\csname papernump530nt_P0_R2\endcsname{0.9852}
\expandafter\def\csname papernump530nt_P0_R3\endcsname{0.1129}
\expandafter\def\csname papernump530nt_P0_R4\endcsname{0.1166}
\expandafter\def\csname papernump530nt_P1_R1\endcsname{0.8566}
\expandafter\def\csname papernump530nt_P1_R2\endcsname{0.9854}
\expandafter\def\csname papernump530nt_P1_R3\endcsname{0.1100}
\expandafter\def\csname papernump530nt_P1_R4\endcsname{0.1136}
\expandafter\def\csname papernump530nt_P2_R1\endcsname{0.8478}
\expandafter\def\csname papernump530nt_P2_R2\endcsname{0.9848}
\expandafter\def\csname papernump530nt_P2_R3\endcsname{0.1156}
\expandafter\def\csname papernump530nt_P2_R4\endcsname{0.1212}
\expandafter\def\csname papernump530nt_PA_R1_max\endcsname{0.1377}
\expandafter\def\csname papernump530nt_PA_R1_min\endcsname{0.1264}
\expandafter\def\csname papernump530nt_P_R3_max\endcsname{0.1156}
\expandafter\def\csname papernump530nt_P_R3_min\endcsname{0.1100}
\expandafter\def\csname papernump530tr_A0_R1\endcsname{0.6316}
\expandafter\def\csname papernump530tr_A0_R2\endcsname{0.9928}
\expandafter\def\csname papernump530tr_A0_R3\endcsname{0.3092}
\expandafter\def\csname papernump530tr_A1_R1\endcsname{0.6164}
\expandafter\def\csname papernump530tr_A1_R2\endcsname{0.9924}
\expandafter\def\csname papernump530tr_A1_R3\endcsname{0.3275}
\expandafter\def\csname papernump530tr_A2_R1\endcsname{0.6359}
\expandafter\def\csname papernump530tr_A2_R2\endcsname{0.9928}
\expandafter\def\csname papernump530tr_A2_R3\endcsname{0.3060}
\expandafter\def\csname papernump530tr_M0_R1\endcsname{0.7231}
\expandafter\def\csname papernump530tr_M0_R2\endcsname{0.9932}
\expandafter\def\csname papernump530tr_M0_R3\endcsname{0.2083}
\expandafter\def\csname papernump530tr_M1_R1\endcsname{0.7010}
\expandafter\def\csname papernump530tr_M1_R2\endcsname{0.9931}
\expandafter\def\csname papernump530tr_M1_R3\endcsname{0.2326}
\expandafter\def\csname papernump530tr_M2_R1\endcsname{0.7193}
\expandafter\def\csname papernump530tr_M2_R2\endcsname{0.9928}
\expandafter\def\csname papernump530tr_M2_R3\endcsname{0.2145}
\expandafter\def\csname papernump530tr_P0_R1\endcsname{0.9419}
\expandafter\def\csname papernump530tr_P0_R2\endcsname{0.9929}
\expandafter\def\csname papernump530tr_P0_R3\endcsname{0.0416}
\expandafter\def\csname papernump530tr_P1_R1\endcsname{0.9425}
\expandafter\def\csname papernump530tr_P1_R2\endcsname{0.9927}
\expandafter\def\csname papernump530tr_P1_R3\endcsname{0.0417}
\expandafter\def\csname papernump530tr_P2_R1\endcsname{0.9393}
\expandafter\def\csname papernump530tr_P2_R2\endcsname{0.9924}
\expandafter\def\csname papernump530tr_P2_R3\endcsname{0.0437}
\expandafter\def\csname papernumpaired_pool\endcsname{12,576}
\expandafter\def\csname papernumpdm_reproduced\endcsname{103,280}
\expandafter\def\csname papernumpopulation_tokens\endcsname{103,288}
\expandafter\def\csname papernumprobe_matched_lk_ap\endcsname{0.956}
\expandafter\def\csname papernumprobe_matched_lk_auc\endcsname{0.9905}
\expandafter\def\csname papernumprobe_matched_lk_auc_hi\endcsname{0.9908}
\expandafter\def\csname papernumprobe_matched_lk_auc_lo\endcsname{0.9901}
\expandafter\def\csname papernumprobe_matched_lk_pauc\endcsname{0.953}
\expandafter\def\csname papernumprobe_T_dac_auc\endcsname{0.6241}
\expandafter\def\csname papernumprobe_T_dac_share\endcsname{81}
\expandafter\def\csname papernumprobe_T_lk_ap\endcsname{0.201}
\expandafter\def\csname papernumprobe_T_lk_auc\endcsname{0.6538}
\expandafter\def\csname papernumprobe_T_lk_auc_hi\endcsname{0.6609}
\expandafter\def\csname papernumprobe_T_lk_auc_lo\endcsname{0.6465}
\expandafter\def\csname papernumprobe_T_lk_pauc\endcsname{0.516}
\expandafter\def\csname papernumprobe_Tmatched_lk_auc\endcsname{0.7046}
\expandafter\def\csname papernumprobe_U_dac_auc\endcsname{0.8489}
\expandafter\def\csname papernumprobe_U_dac_share\endcsname{71}
\expandafter\def\csname papernumprobe_U_lk_auc\endcsname{0.9940}
\expandafter\def\csname papernumprobe_U_scenes\endcsname{86,107}
\expandafter\def\csname papernumprobe_asd_after\endcsname{0.016}
\expandafter\def\csname papernumprobe_delta_match\endcsname{-0.2859}
\expandafter\def\csname papernumprobe_delta_match_hi\endcsname{-0.2797}
\expandafter\def\csname papernumprobe_delta_match_lo\endcsname{-0.2939}
\expandafter\def\csname papernumprobe_delta_strat\endcsname{-0.3733}
\expandafter\def\csname papernumprobe_delta_strat_hi\endcsname{-0.3679}
\expandafter\def\csname papernumprobe_delta_strat_lo\endcsname{-0.3787}
\expandafter\def\csname papernumprobe_did_match\endcsname{0.2850}
\expandafter\def\csname papernumprobe_did_match_hi\endcsname{0.2922}
\expandafter\def\csname papernumprobe_did_match_lo\endcsname{0.2770}
\expandafter\def\csname papernumprobe_did_strat\endcsname{0.3710}
\expandafter\def\csname papernumprobe_did_strat_hi\endcsname{0.3766}
\expandafter\def\csname papernumprobe_did_strat_lo\endcsname{0.3658}
\expandafter\def\csname papernumprobe_pairs\endcsname{7,940}
\expandafter\def\csname papernumprobe_pauc_floor\endcsname{0.474}
\expandafter\def\csname papernumprobe_rank_trapezoid_max\endcsname{0.00000763}
\expandafter\def\csname papernumprobe_unmatched\endcsname{1,891}
\expandafter\def\csname papernumprt_A0_R1\endcsname{0.6257}
\expandafter\def\csname papernumprt_A0_R2\endcsname{0.9922}
\expandafter\def\csname papernumprt_A0_R3\endcsname{0.3225}
\expandafter\def\csname papernumprt_A1_R1\endcsname{0.6423}
\expandafter\def\csname papernumprt_A1_R2\endcsname{0.9925}
\expandafter\def\csname papernumprt_A1_R3\endcsname{0.3007}
\expandafter\def\csname papernumprt_A2_R1\endcsname{0.6141}
\expandafter\def\csname papernumprt_A2_R2\endcsname{0.9919}
\expandafter\def\csname papernumprt_A2_R3\endcsname{0.3219}
\expandafter\def\csname papernumprt_AMP_R2_max\endcsname{0.9927}
\expandafter\def\csname papernumprt_AMP_R2_min\endcsname{0.9919}
\expandafter\def\csname papernumprt_A_R1_max\endcsname{0.6423}
\expandafter\def\csname papernumprt_A_R1_min\endcsname{0.6141}
\expandafter\def\csname papernumprt_A_R3_max\endcsname{0.3225}
\expandafter\def\csname papernumprt_A_R3_min\endcsname{0.3007}
\expandafter\def\csname papernumprt_L0_R1\endcsname{0.4447}
\expandafter\def\csname papernumprt_L0_R2\endcsname{0.3003}
\expandafter\def\csname papernumprt_L0_R3\endcsname{-0.1318}
\expandafter\def\csname papernumprt_M0_R1\endcsname{0.6962}
\expandafter\def\csname papernumprt_M0_R2\endcsname{0.9925}
\expandafter\def\csname papernumprt_M0_R3\endcsname{0.2378}
\expandafter\def\csname papernumprt_M1_R1\endcsname{0.7044}
\expandafter\def\csname papernumprt_M1_R2\endcsname{0.9927}
\expandafter\def\csname papernumprt_M1_R3\endcsname{0.2239}
\expandafter\def\csname papernumprt_M2_R1\endcsname{0.6965}
\expandafter\def\csname papernumprt_M2_R2\endcsname{0.9923}
\expandafter\def\csname papernumprt_M2_R3\endcsname{0.2363}
\expandafter\def\csname papernumprt_P0_R1\endcsname{0.9225}
\expandafter\def\csname papernumprt_P0_R2\endcsname{0.9920}
\expandafter\def\csname papernumprt_P0_R3\endcsname{0.0570}
\expandafter\def\csname papernumprt_P1_R1\endcsname{0.9233}
\expandafter\def\csname papernumprt_P1_R2\endcsname{0.9921}
\expandafter\def\csname papernumprt_P1_R3\endcsname{0.0563}
\expandafter\def\csname papernumprt_P2_R1\endcsname{0.9230}
\expandafter\def\csname papernumprt_P2_R2\endcsname{0.9919}
\expandafter\def\csname papernumprt_P2_R3\endcsname{0.0567}
\expandafter\def\csname papernumprt_P_R1_max\endcsname{0.9233}
\expandafter\def\csname papernumprt_P_R1_min\endcsname{0.9225}
\expandafter\def\csname papernumprt_P_R3_max\endcsname{0.0570}
\expandafter\def\csname papernumprt_P_R3_min\endcsname{0.0563}
\expandafter\def\csname papernumranks_DDC\endcsname{5,919, 14,857, 16,317, 16,384, 16,384}
\expandafter\def\csname papernumranks_LK\endcsname{306, 10,484, 14,391, 15,604, 16,384}
\expandafter\def\csname papernumranks_NC\endcsname{11,343, 16,191, 16,384, 16,384, 16,384}
\expandafter\def\csname papernumranks_TLC\endcsname{650, 12,365, 16,380, 16,384, 16,384}
\expandafter\def\csname papernumranks_TTC\endcsname{5,726, 15,384, 16,384, 16,384, 16,384}
\expandafter\def\csname papernumreproduced\endcsname{103,278}
\expandafter\def\csname papernumresidual_total\endcsname{10}
\expandafter\def\csname papernumrome_expected\endcsname{260}
\expandafter\def\csname papernumrome_tokens\endcsname{260}
\expandafter\def\csname papernumrt530_IM0_off_EC\endcsname{-1.073}
\expandafter\def\csname papernumrt530_IM0_off_EPDMS\endcsname{+0.303}
\expandafter\def\csname papernumrt530_IM0_off_LK\endcsname{-0.100}
\expandafter\def\csname papernumrt530_IM0_on_EPDMS\endcsname{+0.284}
\expandafter\def\csname papernumrt530_IM0_on_LK\endcsname{-0.210}
\expandafter\def\csname papernumrt530_IM1_off_EC\endcsname{+0.257}
\expandafter\def\csname papernumrt530_IM1_off_EPDMS\endcsname{+0.111}
\expandafter\def\csname papernumrt530_IM1_off_LK\endcsname{+0.343}
\expandafter\def\csname papernumrt530_IM1_on_EPDMS\endcsname{-0.213}
\expandafter\def\csname papernumrt530_IM1_on_LK\endcsname{-0.042}
\expandafter\def\csname papernumrt530_IM2_off_EC\endcsname{-1.507}
\expandafter\def\csname papernumrt530_IM2_off_EPDMS\endcsname{-0.285}
\expandafter\def\csname papernumrt530_IM2_off_LK\endcsname{+0.173}
\expandafter\def\csname papernumrt530_IM2_on_EPDMS\endcsname{-0.548}
\expandafter\def\csname papernumrt530_IM2_on_LK\endcsname{-0.378}
\expandafter\def\csname papernumrt530_IP0_off_EC\endcsname{-5.948}
\expandafter\def\csname papernumrt530_IP0_off_EPDMS\endcsname{+0.553}
\expandafter\def\csname papernumrt530_IP0_off_LK\endcsname{+2.870}
\expandafter\def\csname papernumrt530_IP0_on_EPDMS\endcsname{+0.054}
\expandafter\def\csname papernumrt530_IP0_on_LK\endcsname{-0.875}
\expandafter\def\csname papernumrt530_IP1_off_EC\endcsname{-5.115}
\expandafter\def\csname papernumrt530_IP1_off_EPDMS\endcsname{+0.152}
\expandafter\def\csname papernumrt530_IP1_off_LK\endcsname{+3.008}
\expandafter\def\csname papernumrt530_IP1_on_EPDMS\endcsname{-0.617}
\expandafter\def\csname papernumrt530_IP1_on_LK\endcsname{-0.516}
\expandafter\def\csname papernumrt530_IP2_off_EC\endcsname{-6.800}
\expandafter\def\csname papernumrt530_IP2_off_EPDMS\endcsname{-0.234}
\expandafter\def\csname papernumrt530_IP2_off_LK\endcsname{+2.218}
\expandafter\def\csname papernumrt530_IP2_on_EPDMS\endcsname{-0.657}
\expandafter\def\csname papernumrt530_IP2_on_LK\endcsname{-0.865}
\expandafter\def\csname papernumrt530_L0A0_off_EC\endcsname{-0.349}
\expandafter\def\csname papernumrt530_L0A0_off_EC_hi\endcsname{+0.302}
\expandafter\def\csname papernumrt530_L0A0_off_EC_lo\endcsname{-1.011}
\expandafter\def\csname papernumrt530_L0A0_off_EPDMS\endcsname{-0.256}
\expandafter\def\csname papernumrt530_L0A0_off_EPDMS_hi\endcsname{-0.051}
\expandafter\def\csname papernumrt530_L0A0_off_EPDMS_lo\endcsname{-0.470}
\expandafter\def\csname papernumrt530_L0A0_off_LK\endcsname{-1.169}
\expandafter\def\csname papernumrt530_L0A0_off_LK_hi\endcsname{-0.835}
\expandafter\def\csname papernumrt530_L0A0_off_LK_lo\endcsname{-1.516}
\expandafter\def\csname papernumrt530_L0A0_on_EPDMS\endcsname{-0.265}
\expandafter\def\csname papernumrt530_L0A0_on_EPDMS_hi\endcsname{-0.044}
\expandafter\def\csname papernumrt530_L0A0_on_EPDMS_lo\endcsname{-0.494}
\expandafter\def\csname papernumrt530_L0A0_on_LK\endcsname{-1.021}
\expandafter\def\csname papernumrt530_L0A0_on_LK_hi\endcsname{-0.731}
\expandafter\def\csname papernumrt530_L0A0_on_LK_lo\endcsname{-1.318}
\expandafter\def\csname papernumrt530_M0A0_off_EC_all\endcsname{-0.060}
\expandafter\def\csname papernumrt530_M0A0_off_EC_human_fail\endcsname{-0.980}
\expandafter\def\csname papernumrt530_M0A0_off_EC_rest\endcsname{+0.093}
\expandafter\def\csname papernumrt530_M0A0_off_EPDMS_all\endcsname{+0.075}
\expandafter\def\csname papernumrt530_M0A0_off_EPDMS_human_fail\endcsname{+0.332}
\expandafter\def\csname papernumrt530_M0A0_off_EPDMS_rest\endcsname{+0.030}
\expandafter\def\csname papernumrt530_M0A0_off_LK_all\endcsname{+0.140}
\expandafter\def\csname papernumrt530_M0A0_off_LK_human_fail\endcsname{+0.055}
\expandafter\def\csname papernumrt530_M0A0_off_LK_rest\endcsname{+0.155}
\expandafter\def\csname papernumrt530_M0A0_on_EPDMS_all\endcsname{+0.072}
\expandafter\def\csname papernumrt530_M0A0_on_EPDMS_human_fail\endcsname{+0.314}
\expandafter\def\csname papernumrt530_M0A0_on_EPDMS_rest\endcsname{+0.030}
\expandafter\def\csname papernumrt530_M0A0_on_LK_all\endcsname{+0.123}
\expandafter\def\csname papernumrt530_M0A0_on_LK_human_fail\endcsname{-0.055}
\expandafter\def\csname papernumrt530_M0A0_on_LK_rest\endcsname{+0.155}
\expandafter\def\csname papernumrt530_M1A1_off_EC_all\endcsname{-0.010}
\expandafter\def\csname papernumrt530_M1A1_off_EC_human_fail\endcsname{+0.210}
\expandafter\def\csname papernumrt530_M1A1_off_EC_rest\endcsname{-0.046}
\expandafter\def\csname papernumrt530_M1A1_off_EPDMS_all\endcsname{-0.218}
\expandafter\def\csname papernumrt530_M1A1_off_EPDMS_human_fail\endcsname{-0.123}
\expandafter\def\csname papernumrt530_M1A1_off_EPDMS_rest\endcsname{-0.235}
\expandafter\def\csname papernumrt530_M1A1_off_LK_all\endcsname{-0.016}
\expandafter\def\csname papernumrt530_M1A1_off_LK_human_fail\endcsname{+0.275}
\expandafter\def\csname papernumrt530_M1A1_off_LK_rest\endcsname{-0.068}
\expandafter\def\csname papernumrt530_M1A1_on_EPDMS_all\endcsname{-0.266}
\expandafter\def\csname papernumrt530_M1A1_on_EPDMS_human_fail\endcsname{-0.448}
\expandafter\def\csname papernumrt530_M1A1_on_EPDMS_rest\endcsname{-0.235}
\expandafter\def\csname papernumrt530_M1A1_on_LK_all\endcsname{-0.074}
\expandafter\def\csname papernumrt530_M1A1_on_LK_human_fail\endcsname{-0.110}
\expandafter\def\csname papernumrt530_M1A1_on_LK_rest\endcsname{-0.068}
\expandafter\def\csname papernumrt530_M2A2_off_EC_all\endcsname{-0.528}
\expandafter\def\csname papernumrt530_M2A2_off_EC_human_fail\endcsname{-1.821}
\expandafter\def\csname papernumrt530_M2A2_off_EC_rest\endcsname{-0.314}
\expandafter\def\csname papernumrt530_M2A2_off_EPDMS_all\endcsname{+0.059}
\expandafter\def\csname papernumrt530_M2A2_off_EPDMS_human_fail\endcsname{-0.184}
\expandafter\def\csname papernumrt530_M2A2_off_EPDMS_rest\endcsname{+0.102}
\expandafter\def\csname papernumrt530_M2A2_off_LK_all\endcsname{+0.403}
\expandafter\def\csname papernumrt530_M2A2_off_LK_human_fail\endcsname{+0.551}
\expandafter\def\csname papernumrt530_M2A2_off_LK_rest\endcsname{+0.378}
\expandafter\def\csname papernumrt530_M2A2_on_EPDMS_all\endcsname{+0.020}
\expandafter\def\csname papernumrt530_M2A2_on_EPDMS_human_fail\endcsname{-0.446}
\expandafter\def\csname papernumrt530_M2A2_on_EPDMS_rest\endcsname{+0.102}
\expandafter\def\csname papernumrt530_M2A2_on_LK_all\endcsname{+0.321}
\expandafter\def\csname papernumrt530_M2A2_on_LK_human_fail\endcsname{+0.000}
\expandafter\def\csname papernumrt530_M2A2_on_LK_rest\endcsname{+0.378}
\expandafter\def\csname papernumrt530_P0A0_off_EC_all\endcsname{-2.321}
\expandafter\def\csname papernumrt530_P0A0_off_EC_human_fail\endcsname{-7.423}
\expandafter\def\csname papernumrt530_P0A0_off_EC_rest\endcsname{-1.475}
\expandafter\def\csname papernumrt530_P0A0_off_EPDMS_all\endcsname{-0.259}
\expandafter\def\csname papernumrt530_P0A0_off_EPDMS_human_fail\endcsname{+0.212}
\expandafter\def\csname papernumrt530_P0A0_off_EPDMS_rest\endcsname{-0.341}
\expandafter\def\csname papernumrt530_P0A0_off_LK_all\endcsname{+1.194}
\expandafter\def\csname papernumrt530_P0A0_off_LK_human_fail\endcsname{+3.634}
\expandafter\def\csname papernumrt530_P0A0_off_LK_rest\endcsname{+0.765}
\expandafter\def\csname papernumrt530_P0A0_on_EPDMS_all\endcsname{-0.333}
\expandafter\def\csname papernumrt530_P0A0_on_EPDMS_human_fail\endcsname{-0.287}
\expandafter\def\csname papernumrt530_P0A0_on_EPDMS_rest\endcsname{-0.341}
\expandafter\def\csname papernumrt530_P0A0_on_LK_all\endcsname{+0.634}
\expandafter\def\csname papernumrt530_P0A0_on_LK_human_fail\endcsname{-0.110}
\expandafter\def\csname papernumrt530_P0A0_on_LK_rest\endcsname{+0.765}
\expandafter\def\csname papernumrt530_P1A1_off_EC_all\endcsname{-1.285}
\expandafter\def\csname papernumrt530_P1A1_off_EC_human_fail\endcsname{-5.672}
\expandafter\def\csname papernumrt530_P1A1_off_EC_rest\endcsname{-0.557}
\expandafter\def\csname papernumrt530_P1A1_off_EPDMS_all\endcsname{+0.064}
\expandafter\def\csname papernumrt530_P1A1_off_EPDMS_human_fail\endcsname{+0.193}
\expandafter\def\csname papernumrt530_P1A1_off_EPDMS_rest\endcsname{+0.041}
\expandafter\def\csname papernumrt530_P1A1_off_LK_all\endcsname{+1.021}
\expandafter\def\csname papernumrt530_P1A1_off_LK_human_fail\endcsname{+3.579}
\expandafter\def\csname papernumrt530_P1A1_off_LK_rest\endcsname{+0.571}
\expandafter\def\csname papernumrt530_P1A1_on_EPDMS_all\endcsname{-0.051}
\expandafter\def\csname papernumrt530_P1A1_on_EPDMS_human_fail\endcsname{-0.576}
\expandafter\def\csname papernumrt530_P1A1_on_EPDMS_rest\endcsname{+0.041}
\expandafter\def\csname papernumrt530_P1A1_on_LK_all\endcsname{+0.494}
\expandafter\def\csname papernumrt530_P1A1_on_LK_human_fail\endcsname{+0.055}
\expandafter\def\csname papernumrt530_P1A1_on_LK_rest\endcsname{+0.571}
\expandafter\def\csname papernumrt530_P2A2_off_EC_all\endcsname{-2.291}
\expandafter\def\csname papernumrt530_P2A2_off_EC_human_fail\endcsname{-8.123}
\expandafter\def\csname papernumrt530_P2A2_off_EC_rest\endcsname{-1.324}
\expandafter\def\csname papernumrt530_P2A2_off_EPDMS_all\endcsname{-0.302}
\expandafter\def\csname papernumrt530_P2A2_off_EPDMS_human_fail\endcsname{-0.500}
\expandafter\def\csname papernumrt530_P2A2_off_EPDMS_rest\endcsname{-0.267}
\expandafter\def\csname papernumrt530_P2A2_off_LK_all\endcsname{+1.087}
\expandafter\def\csname papernumrt530_P2A2_off_LK_human_fail\endcsname{+2.974}
\expandafter\def\csname papernumrt530_P2A2_off_LK_rest\endcsname{+0.755}
\expandafter\def\csname papernumrt530_P2A2_on_EPDMS_all\endcsname{-0.365}
\expandafter\def\csname papernumrt530_P2A2_on_EPDMS_human_fail\endcsname{-0.924}
\expandafter\def\csname papernumrt530_P2A2_on_EPDMS_rest\endcsname{-0.267}
\expandafter\def\csname papernumrt530_P2A2_on_LK_all\endcsname{+0.626}
\expandafter\def\csname papernumrt530_P2A2_on_LK_human_fail\endcsname{-0.110}
\expandafter\def\csname papernumrt530_P2A2_on_LK_rest\endcsname{+0.755}
\expandafter\def\csname papernumrt530_steps\endcsname{530}
\expandafter\def\csname papernumrt_A0released_off_EC_all\endcsname{+3.635}
\expandafter\def\csname papernumrt_A0released_off_EC_all_hi\endcsname{+4.498}
\expandafter\def\csname papernumrt_A0released_off_EC_all_lo\endcsname{+2.842}
\expandafter\def\csname papernumrt_A0released_off_EPDMS_all\endcsname{-0.277}
\expandafter\def\csname papernumrt_A0released_off_EPDMS_all_hi\endcsname{+0.151}
\expandafter\def\csname papernumrt_A0released_off_EPDMS_all_lo\endcsname{-0.743}
\expandafter\def\csname papernumrt_A0released_off_LK_all\endcsname{+0.313}
\expandafter\def\csname papernumrt_A0released_off_LK_all_hi\endcsname{+0.648}
\expandafter\def\csname papernumrt_A0released_off_LK_all_lo\endcsname{-0.019}
\expandafter\def\csname papernumrt_A0released_on_EPDMS_all\endcsname{-0.195}
\expandafter\def\csname papernumrt_A0released_on_EPDMS_all_hi\endcsname{+0.224}
\expandafter\def\csname papernumrt_A0released_on_EPDMS_all_lo\endcsname{-0.658}
\expandafter\def\csname papernumrt_A0released_on_LK_all\endcsname{+0.123}
\expandafter\def\csname papernumrt_A0released_on_LK_all_hi\endcsname{+0.396}
\expandafter\def\csname papernumrt_A0released_on_LK_all_lo\endcsname{-0.150}
\expandafter\def\csname papernumrt_unchanged_EC_drop_max\endcsname{2.829}
\expandafter\def\csname papernumrt_unchanged_EC_drop_min\endcsname{1.448}
\expandafter\def\csname papernumrt_IP0_off_EC\endcsname{-5.740}
\expandafter\def\csname papernumrt_IP0_off_EC_hi\endcsname{-3.385}
\expandafter\def\csname papernumrt_IP0_off_EC_lo\endcsname{-8.112}
\expandafter\def\csname papernumrt_IP0_off_LK\endcsname{+1.820}
\expandafter\def\csname papernumrt_IP0_off_LK_hi\endcsname{+3.134}
\expandafter\def\csname papernumrt_IP0_off_LK_lo\endcsname{+0.619}
\expandafter\def\csname papernumrt_IP1_off_EC\endcsname{-5.583}
\expandafter\def\csname papernumrt_IP1_off_EC_hi\endcsname{-3.109}
\expandafter\def\csname papernumrt_IP1_off_EC_lo\endcsname{-8.070}
\expandafter\def\csname papernumrt_IP1_off_LK\endcsname{+3.243}
\expandafter\def\csname papernumrt_IP1_off_LK_hi\endcsname{+4.861}
\expandafter\def\csname papernumrt_IP1_off_LK_lo\endcsname{+1.704}
\expandafter\def\csname papernumrt_IP2_off_EC\endcsname{-4.236}
\expandafter\def\csname papernumrt_IP2_off_EC_hi\endcsname{-1.658}
\expandafter\def\csname papernumrt_IP2_off_EC_lo\endcsname{-6.772}
\expandafter\def\csname papernumrt_IP2_off_LK\endcsname{+2.795}
\expandafter\def\csname papernumrt_IP2_off_LK_hi\endcsname{+4.152}
\expandafter\def\csname papernumrt_IP2_off_LK_lo\endcsname{+1.377}
\expandafter\def\csname papernumrt_L0A0_off_EC_all\endcsname{-0.458}
\expandafter\def\csname papernumrt_L0A0_off_EC_all_hi\endcsname{+0.049}
\expandafter\def\csname papernumrt_L0A0_off_EC_all_lo\endcsname{-0.953}
\expandafter\def\csname papernumrt_L0A0_off_EPDMS_all\endcsname{-0.047}
\expandafter\def\csname papernumrt_L0A0_off_EPDMS_all_hi\endcsname{+0.237}
\expandafter\def\csname papernumrt_L0A0_off_EPDMS_all_lo\endcsname{-0.311}
\expandafter\def\csname papernumrt_L0A0_off_LK_all\endcsname{-1.219}
\expandafter\def\csname papernumrt_L0A0_off_LK_all_hi\endcsname{-0.872}
\expandafter\def\csname papernumrt_L0A0_off_LK_all_lo\endcsname{-1.561}
\expandafter\def\csname papernumrt_L0A0_on_EPDMS_all\endcsname{-0.072}
\expandafter\def\csname papernumrt_L0A0_on_EPDMS_all_hi\endcsname{+0.195}
\expandafter\def\csname papernumrt_L0A0_on_EPDMS_all_lo\endcsname{-0.323}
\expandafter\def\csname papernumrt_L0A0_on_LK_all\endcsname{-0.972}
\expandafter\def\csname papernumrt_L0A0_on_LK_all_hi\endcsname{-0.684}
\expandafter\def\csname papernumrt_L0A0_on_LK_all_lo\endcsname{-1.259}
\expandafter\def\csname papernumrt_LK_forgiven_max\endcsname{3.524}
\expandafter\def\csname papernumrt_LK_forgiven_min\endcsname{2.478}
\expandafter\def\csname papernumrt_M0A0_off_EC_all\endcsname{-0.309}
\expandafter\def\csname papernumrt_M0A0_off_EC_all_hi\endcsname{+0.220}
\expandafter\def\csname papernumrt_M0A0_off_EC_all_lo\endcsname{-0.860}
\expandafter\def\csname papernumrt_M0A0_off_EC_human_fail\endcsname{-1.821}
\expandafter\def\csname papernumrt_M0A0_off_EC_human_fail_hi\endcsname{+0.065}
\expandafter\def\csname papernumrt_M0A0_off_EC_human_fail_lo\endcsname{-3.773}
\expandafter\def\csname papernumrt_M0A0_off_EC_rest\endcsname{-0.058}
\expandafter\def\csname papernumrt_M0A0_off_EC_rest_hi\endcsname{+0.471}
\expandafter\def\csname papernumrt_M0A0_off_EC_rest_lo\endcsname{-0.608}
\expandafter\def\csname papernumrt_M0A0_off_EPDMS_all\endcsname{+0.173}
\expandafter\def\csname papernumrt_M0A0_off_EPDMS_all_hi\endcsname{+0.389}
\expandafter\def\csname papernumrt_M0A0_off_EPDMS_all_lo\endcsname{-0.031}
\expandafter\def\csname papernumrt_M0A0_off_EPDMS_human_fail\endcsname{+0.350}
\expandafter\def\csname papernumrt_M0A0_off_EPDMS_human_fail_hi\endcsname{+0.872}
\expandafter\def\csname papernumrt_M0A0_off_EPDMS_human_fail_lo\endcsname{-0.141}
\expandafter\def\csname papernumrt_M0A0_off_EPDMS_rest\endcsname{+0.142}
\expandafter\def\csname papernumrt_M0A0_off_EPDMS_rest_hi\endcsname{+0.371}
\expandafter\def\csname papernumrt_M0A0_off_EPDMS_rest_lo\endcsname{-0.076}
\expandafter\def\csname papernumrt_M0A0_off_LK_all\endcsname{+0.305}
\expandafter\def\csname papernumrt_M0A0_off_LK_all_hi\endcsname{+0.541}
\expandafter\def\csname papernumrt_M0A0_off_LK_all_lo\endcsname{+0.083}
\expandafter\def\csname papernumrt_M0A0_off_LK_human_fail\endcsname{+0.551}
\expandafter\def\csname papernumrt_M0A0_off_LK_human_fail_hi\endcsname{+1.272}
\expandafter\def\csname papernumrt_M0A0_off_LK_human_fail_lo\endcsname{-0.138}
\expandafter\def\csname papernumrt_M0A0_off_LK_rest\endcsname{+0.261}
\expandafter\def\csname papernumrt_M0A0_off_LK_rest_hi\endcsname{+0.493}
\expandafter\def\csname papernumrt_M0A0_off_LK_rest_lo\endcsname{+0.043}
\expandafter\def\csname papernumrt_M0A0_on_EPDMS_all\endcsname{+0.047}
\expandafter\def\csname papernumrt_M0A0_on_EPDMS_all_hi\endcsname{+0.263}
\expandafter\def\csname papernumrt_M0A0_on_EPDMS_all_lo\endcsname{-0.158}
\expandafter\def\csname papernumrt_M0A0_on_EPDMS_human_fail\endcsname{-0.493}
\expandafter\def\csname papernumrt_M0A0_on_EPDMS_human_fail_hi\endcsname{+0.129}
\expandafter\def\csname papernumrt_M0A0_on_EPDMS_human_fail_lo\endcsname{-1.144}
\expandafter\def\csname papernumrt_M0A0_on_EPDMS_rest\endcsname{+0.142}
\expandafter\def\csname papernumrt_M0A0_on_EPDMS_rest_hi\endcsname{+0.371}
\expandafter\def\csname papernumrt_M0A0_on_EPDMS_rest_lo\endcsname{-0.076}
\expandafter\def\csname papernumrt_M0A0_on_LK_all\endcsname{+0.247}
\expandafter\def\csname papernumrt_M0A0_on_LK_all_hi\endcsname{+0.453}
\expandafter\def\csname papernumrt_M0A0_on_LK_all_lo\endcsname{+0.049}
\expandafter\def\csname papernumrt_M0A0_on_LK_human_fail\endcsname{+0.165}
\expandafter\def\csname papernumrt_M0A0_on_LK_human_fail_hi\endcsname{+0.452}
\expandafter\def\csname papernumrt_M0A0_on_LK_human_fail_lo\endcsname{-0.095}
\expandafter\def\csname papernumrt_M0A0_on_LK_rest\endcsname{+0.261}
\expandafter\def\csname papernumrt_M0A0_on_LK_rest_hi\endcsname{+0.493}
\expandafter\def\csname papernumrt_M0A0_on_LK_rest_lo\endcsname{+0.043}
\expandafter\def\csname papernumrt_M1A1_off_EC_all\endcsname{-0.040}
\expandafter\def\csname papernumrt_M1A1_off_EC_all_hi\endcsname{+0.482}
\expandafter\def\csname papernumrt_M1A1_off_EC_all_lo\endcsname{-0.594}
\expandafter\def\csname papernumrt_M1A1_off_EC_human_fail\endcsname{-1.120}
\expandafter\def\csname papernumrt_M1A1_off_EC_human_fail_hi\endcsname{+0.928}
\expandafter\def\csname papernumrt_M1A1_off_EC_human_fail_lo\endcsname{-3.253}
\expandafter\def\csname papernumrt_M1A1_off_EC_rest\endcsname{+0.139}
\expandafter\def\csname papernumrt_M1A1_off_EC_rest_hi\endcsname{+0.665}
\expandafter\def\csname papernumrt_M1A1_off_EC_rest_lo\endcsname{-0.392}
\expandafter\def\csname papernumrt_M1A1_off_EPDMS_all\endcsname{-0.034}
\expandafter\def\csname papernumrt_M1A1_off_EPDMS_all_hi\endcsname{+0.177}
\expandafter\def\csname papernumrt_M1A1_off_EPDMS_all_lo\endcsname{-0.238}
\expandafter\def\csname papernumrt_M1A1_off_EPDMS_human_fail\endcsname{-0.037}
\expandafter\def\csname papernumrt_M1A1_off_EPDMS_human_fail_hi\endcsname{+0.432}
\expandafter\def\csname papernumrt_M1A1_off_EPDMS_human_fail_lo\endcsname{-0.470}
\expandafter\def\csname papernumrt_M1A1_off_EPDMS_rest\endcsname{-0.033}
\expandafter\def\csname papernumrt_M1A1_off_EPDMS_rest_hi\endcsname{+0.173}
\expandafter\def\csname papernumrt_M1A1_off_EPDMS_rest_lo\endcsname{-0.237}
\expandafter\def\csname papernumrt_M1A1_off_LK_all\endcsname{+0.231}
\expandafter\def\csname papernumrt_M1A1_off_LK_all_hi\endcsname{+0.435}
\expandafter\def\csname papernumrt_M1A1_off_LK_all_lo\endcsname{+0.042}
\expandafter\def\csname papernumrt_M1A1_off_LK_human_fail\endcsname{+0.661}
\expandafter\def\csname papernumrt_M1A1_off_LK_human_fail_hi\endcsname{+1.463}
\expandafter\def\csname papernumrt_M1A1_off_LK_human_fail_lo\endcsname{-0.119}
\expandafter\def\csname papernumrt_M1A1_off_LK_rest\endcsname{+0.155}
\expandafter\def\csname papernumrt_M1A1_off_LK_rest_hi\endcsname{+0.354}
\expandafter\def\csname papernumrt_M1A1_off_LK_rest_lo\endcsname{-0.030}
\expandafter\def\csname papernumrt_M1A1_on_EPDMS_all\endcsname{-0.015}
\expandafter\def\csname papernumrt_M1A1_on_EPDMS_all_hi\endcsname{+0.198}
\expandafter\def\csname papernumrt_M1A1_on_EPDMS_all_lo\endcsname{-0.223}
\expandafter\def\csname papernumrt_M1A1_on_EPDMS_human_fail\endcsname{+0.090}
\expandafter\def\csname papernumrt_M1A1_on_EPDMS_human_fail_hi\endcsname{+0.658}
\expandafter\def\csname papernumrt_M1A1_on_EPDMS_human_fail_lo\endcsname{-0.442}
\expandafter\def\csname papernumrt_M1A1_on_EPDMS_rest\endcsname{-0.033}
\expandafter\def\csname papernumrt_M1A1_on_EPDMS_rest_hi\endcsname{+0.173}
\expandafter\def\csname papernumrt_M1A1_on_EPDMS_rest_lo\endcsname{-0.237}
\expandafter\def\csname papernumrt_M1A1_on_LK_all\endcsname{+0.165}
\expandafter\def\csname papernumrt_M1A1_on_LK_all_hi\endcsname{+0.335}
\expandafter\def\csname papernumrt_M1A1_on_LK_all_lo\endcsname{+0.008}
\expandafter\def\csname papernumrt_M1A1_on_LK_human_fail\endcsname{+0.220}
\expandafter\def\csname papernumrt_M1A1_on_LK_human_fail_hi\endcsname{+0.460}
\expandafter\def\csname papernumrt_M1A1_on_LK_human_fail_lo\endcsname{+0.048}
\expandafter\def\csname papernumrt_M1A1_on_LK_rest\endcsname{+0.155}
\expandafter\def\csname papernumrt_M1A1_on_LK_rest_hi\endcsname{+0.354}
\expandafter\def\csname papernumrt_M1A1_on_LK_rest_lo\endcsname{-0.030}
\expandafter\def\csname papernumrt_M2A2_off_EC_all\endcsname{-0.588}
\expandafter\def\csname papernumrt_M2A2_off_EC_all_hi\endcsname{-0.042}
\expandafter\def\csname papernumrt_M2A2_off_EC_all_lo\endcsname{-1.111}
\expandafter\def\csname papernumrt_M2A2_off_EC_human_fail\endcsname{-0.070}
\expandafter\def\csname papernumrt_M2A2_off_EC_human_fail_hi\endcsname{+1.427}
\expandafter\def\csname papernumrt_M2A2_off_EC_human_fail_lo\endcsname{-1.488}
\expandafter\def\csname papernumrt_M2A2_off_EC_rest\endcsname{-0.673}
\expandafter\def\csname papernumrt_M2A2_off_EC_rest_hi\endcsname{-0.129}
\expandafter\def\csname papernumrt_M2A2_off_EC_rest_lo\endcsname{-1.193}
\expandafter\def\csname papernumrt_M2A2_off_EPDMS_all\endcsname{+0.144}
\expandafter\def\csname papernumrt_M2A2_off_EPDMS_all_hi\endcsname{+0.343}
\expandafter\def\csname papernumrt_M2A2_off_EPDMS_all_lo\endcsname{-0.039}
\expandafter\def\csname papernumrt_M2A2_off_EPDMS_human_fail\endcsname{-0.151}
\expandafter\def\csname papernumrt_M2A2_off_EPDMS_human_fail_hi\endcsname{+0.343}
\expandafter\def\csname papernumrt_M2A2_off_EPDMS_human_fail_lo\endcsname{-0.677}
\expandafter\def\csname papernumrt_M2A2_off_EPDMS_rest\endcsname{+0.196}
\expandafter\def\csname papernumrt_M2A2_off_EPDMS_rest_hi\endcsname{+0.420}
\expandafter\def\csname papernumrt_M2A2_off_EPDMS_rest_lo\endcsname{-0.014}
\expandafter\def\csname papernumrt_M2A2_off_LK_all\endcsname{+0.280}
\expandafter\def\csname papernumrt_M2A2_off_LK_all_hi\endcsname{+0.500}
\expandafter\def\csname papernumrt_M2A2_off_LK_all_lo\endcsname{+0.064}
\expandafter\def\csname papernumrt_M2A2_off_LK_human_fail\endcsname{+0.771}
\expandafter\def\csname papernumrt_M2A2_off_LK_human_fail_hi\endcsname{+1.532}
\expandafter\def\csname papernumrt_M2A2_off_LK_human_fail_lo\endcsname{+0.000}
\expandafter\def\csname papernumrt_M2A2_off_LK_rest\endcsname{+0.194}
\expandafter\def\csname papernumrt_M2A2_off_LK_rest_hi\endcsname{+0.407}
\expandafter\def\csname papernumrt_M2A2_off_LK_rest_lo\endcsname{-0.010}
\expandafter\def\csname papernumrt_M2A2_on_EPDMS_all\endcsname{+0.139}
\expandafter\def\csname papernumrt_M2A2_on_EPDMS_all_hi\endcsname{+0.345}
\expandafter\def\csname papernumrt_M2A2_on_EPDMS_all_lo\endcsname{-0.051}
\expandafter\def\csname papernumrt_M2A2_on_EPDMS_human_fail\endcsname{-0.186}
\expandafter\def\csname papernumrt_M2A2_on_EPDMS_human_fail_hi\endcsname{+0.341}
\expandafter\def\csname papernumrt_M2A2_on_EPDMS_human_fail_lo\endcsname{-0.724}
\expandafter\def\csname papernumrt_M2A2_on_EPDMS_rest\endcsname{+0.196}
\expandafter\def\csname papernumrt_M2A2_on_EPDMS_rest_hi\endcsname{+0.420}
\expandafter\def\csname papernumrt_M2A2_on_EPDMS_rest_lo\endcsname{-0.014}
\expandafter\def\csname papernumrt_M2A2_on_LK_all\endcsname{+0.156}
\expandafter\def\csname papernumrt_M2A2_on_LK_all_hi\endcsname{+0.349}
\expandafter\def\csname papernumrt_M2A2_on_LK_all_lo\endcsname{-0.025}
\expandafter\def\csname papernumrt_M2A2_on_LK_human_fail\endcsname{-0.055}
\expandafter\def\csname papernumrt_M2A2_on_LK_human_fail_hi\endcsname{+0.128}
\expandafter\def\csname papernumrt_M2A2_on_LK_human_fail_lo\endcsname{-0.240}
\expandafter\def\csname papernumrt_M2A2_on_LK_rest\endcsname{+0.194}
\expandafter\def\csname papernumrt_M2A2_on_LK_rest_hi\endcsname{+0.407}
\expandafter\def\csname papernumrt_M2A2_on_LK_rest_lo\endcsname{-0.010}
\expandafter\def\csname papernumrt_P0A0_off_EC_all\endcsname{-2.709}
\expandafter\def\csname papernumrt_P0A0_off_EC_all_hi\endcsname{-1.943}
\expandafter\def\csname papernumrt_P0A0_off_EC_all_lo\endcsname{-3.476}
\expandafter\def\csname papernumrt_P0A0_off_EC_human_fail\endcsname{-7.633}
\expandafter\def\csname papernumrt_P0A0_off_EC_human_fail_hi\endcsname{-5.288}
\expandafter\def\csname papernumrt_P0A0_off_EC_human_fail_lo\endcsname{-9.986}
\expandafter\def\csname papernumrt_P0A0_off_EC_rest\endcsname{-1.893}
\expandafter\def\csname papernumrt_P0A0_off_EC_rest_hi\endcsname{-1.135}
\expandafter\def\csname papernumrt_P0A0_off_EC_rest_lo\endcsname{-2.645}
\expandafter\def\csname papernumrt_P0A0_off_EPDMS_all\endcsname{-0.245}
\expandafter\def\csname papernumrt_P0A0_off_EPDMS_all_hi\endcsname{+0.044}
\expandafter\def\csname papernumrt_P0A0_off_EPDMS_all_lo\endcsname{-0.526}
\expandafter\def\csname papernumrt_P0A0_off_EPDMS_human_fail\endcsname{-0.167}
\expandafter\def\csname papernumrt_P0A0_off_EPDMS_human_fail_hi\endcsname{+0.496}
\expandafter\def\csname papernumrt_P0A0_off_EPDMS_human_fail_lo\endcsname{-0.752}
\expandafter\def\csname papernumrt_P0A0_off_EPDMS_rest\endcsname{-0.259}
\expandafter\def\csname papernumrt_P0A0_off_EPDMS_rest_hi\endcsname{+0.049}
\expandafter\def\csname papernumrt_P0A0_off_EPDMS_rest_lo\endcsname{-0.562}
\expandafter\def\csname papernumrt_P0A0_off_LK_all\endcsname{+0.930}
\expandafter\def\csname papernumrt_P0A0_off_LK_all_hi\endcsname{+1.256}
\expandafter\def\csname papernumrt_P0A0_off_LK_all_lo\endcsname{+0.629}
\expandafter\def\csname papernumrt_P0A0_off_LK_human_fail\endcsname{+2.478}
\expandafter\def\csname papernumrt_P0A0_off_LK_human_fail_hi\endcsname{+3.758}
\expandafter\def\csname papernumrt_P0A0_off_LK_human_fail_lo\endcsname{+1.304}
\expandafter\def\csname papernumrt_P0A0_off_LK_rest\endcsname{+0.658}
\expandafter\def\csname papernumrt_P0A0_off_LK_rest_hi\endcsname{+0.944}
\expandafter\def\csname papernumrt_P0A0_off_LK_rest_lo\endcsname{+0.383}
\expandafter\def\csname papernumrt_P0A0_on_EPDMS_all\endcsname{-0.328}
\expandafter\def\csname papernumrt_P0A0_on_EPDMS_all_hi\endcsname{-0.031}
\expandafter\def\csname papernumrt_P0A0_on_EPDMS_all_lo\endcsname{-0.616}
\expandafter\def\csname papernumrt_P0A0_on_EPDMS_human_fail\endcsname{-0.723}
\expandafter\def\csname papernumrt_P0A0_on_EPDMS_human_fail_hi\endcsname{+0.008}
\expandafter\def\csname papernumrt_P0A0_on_EPDMS_human_fail_lo\endcsname{-1.404}
\expandafter\def\csname papernumrt_P0A0_on_EPDMS_rest\endcsname{-0.259}
\expandafter\def\csname papernumrt_P0A0_on_EPDMS_rest_hi\endcsname{+0.049}
\expandafter\def\csname papernumrt_P0A0_on_EPDMS_rest_lo\endcsname{-0.562}
\expandafter\def\csname papernumrt_P0A0_on_LK_all\endcsname{+0.609}
\expandafter\def\csname papernumrt_P0A0_on_LK_all_hi\endcsname{+0.851}
\expandafter\def\csname papernumrt_P0A0_on_LK_all_lo\endcsname{+0.370}
\expandafter\def\csname papernumrt_P0A0_on_LK_human_fail\endcsname{+0.330}
\expandafter\def\csname papernumrt_P0A0_on_LK_human_fail_hi\endcsname{+0.624}
\expandafter\def\csname papernumrt_P0A0_on_LK_human_fail_lo\endcsname{+0.064}
\expandafter\def\csname papernumrt_P0A0_on_LK_rest\endcsname{+0.658}
\expandafter\def\csname papernumrt_P0A0_on_LK_rest_hi\endcsname{+0.944}
\expandafter\def\csname papernumrt_P0A0_on_LK_rest_lo\endcsname{+0.383}
\expandafter\def\csname papernumrt_P1A1_off_EC_all\endcsname{-1.584}
\expandafter\def\csname papernumrt_P1A1_off_EC_all_hi\endcsname{-0.864}
\expandafter\def\csname papernumrt_P1A1_off_EC_all_lo\endcsname{-2.329}
\expandafter\def\csname papernumrt_P1A1_off_EC_human_fail\endcsname{-6.373}
\expandafter\def\csname papernumrt_P1A1_off_EC_human_fail_hi\endcsname{-3.911}
\expandafter\def\csname papernumrt_P1A1_off_EC_human_fail_lo\endcsname{-8.834}
\expandafter\def\csname papernumrt_P1A1_off_EC_rest\endcsname{-0.790}
\expandafter\def\csname papernumrt_P1A1_off_EC_rest_hi\endcsname{-0.149}
\expandafter\def\csname papernumrt_P1A1_off_EC_rest_lo\endcsname{-1.448}
\expandafter\def\csname papernumrt_P1A1_off_EPDMS_all\endcsname{-0.210}
\expandafter\def\csname papernumrt_P1A1_off_EPDMS_all_hi\endcsname{+0.012}
\expandafter\def\csname papernumrt_P1A1_off_EPDMS_all_lo\endcsname{-0.431}
\expandafter\def\csname papernumrt_P1A1_off_EPDMS_human_fail\endcsname{-0.237}
\expandafter\def\csname papernumrt_P1A1_off_EPDMS_human_fail_hi\endcsname{+0.401}
\expandafter\def\csname papernumrt_P1A1_off_EPDMS_human_fail_lo\endcsname{-0.896}
\expandafter\def\csname papernumrt_P1A1_off_EPDMS_rest\endcsname{-0.205}
\expandafter\def\csname papernumrt_P1A1_off_EPDMS_rest_hi\endcsname{+0.025}
\expandafter\def\csname papernumrt_P1A1_off_EPDMS_rest_lo\endcsname{-0.444}
\expandafter\def\csname papernumrt_P1A1_off_LK_all\endcsname{+0.766}
\expandafter\def\csname papernumrt_P1A1_off_LK_all_hi\endcsname{+1.095}
\expandafter\def\csname papernumrt_P1A1_off_LK_all_lo\endcsname{+0.467}
\expandafter\def\csname papernumrt_P1A1_off_LK_human_fail\endcsname{+3.524}
\expandafter\def\csname papernumrt_P1A1_off_LK_human_fail_hi\endcsname{+5.107}
\expandafter\def\csname papernumrt_P1A1_off_LK_human_fail_lo\endcsname{+2.008}
\expandafter\def\csname papernumrt_P1A1_off_LK_rest\endcsname{+0.281}
\expandafter\def\csname papernumrt_P1A1_off_LK_rest_hi\endcsname{+0.505}
\expandafter\def\csname papernumrt_P1A1_off_LK_rest_lo\endcsname{+0.060}
\expandafter\def\csname papernumrt_P1A1_on_EPDMS_all\endcsname{-0.249}
\expandafter\def\csname papernumrt_P1A1_on_EPDMS_all_hi\endcsname{-0.025}
\expandafter\def\csname papernumrt_P1A1_on_EPDMS_all_lo\endcsname{-0.473}
\expandafter\def\csname papernumrt_P1A1_on_EPDMS_human_fail\endcsname{-0.497}
\expandafter\def\csname papernumrt_P1A1_on_EPDMS_human_fail_hi\endcsname{+0.203}
\expandafter\def\csname papernumrt_P1A1_on_EPDMS_human_fail_lo\endcsname{-1.208}
\expandafter\def\csname papernumrt_P1A1_on_EPDMS_rest\endcsname{-0.205}
\expandafter\def\csname papernumrt_P1A1_on_EPDMS_rest_hi\endcsname{+0.025}
\expandafter\def\csname papernumrt_P1A1_on_EPDMS_rest_lo\endcsname{-0.444}
\expandafter\def\csname papernumrt_P1A1_on_LK_all\endcsname{+0.313}
\expandafter\def\csname papernumrt_P1A1_on_LK_all_hi\endcsname{+0.513}
\expandafter\def\csname papernumrt_P1A1_on_LK_all_lo\endcsname{+0.111}
\expandafter\def\csname papernumrt_P1A1_on_LK_human_fail\endcsname{+0.496}
\expandafter\def\csname papernumrt_P1A1_on_LK_human_fail_hi\endcsname{+0.875}
\expandafter\def\csname papernumrt_P1A1_on_LK_human_fail_lo\endcsname{+0.169}
\expandafter\def\csname papernumrt_P1A1_on_LK_rest\endcsname{+0.281}
\expandafter\def\csname papernumrt_P1A1_on_LK_rest_hi\endcsname{+0.505}
\expandafter\def\csname papernumrt_P1A1_on_LK_rest_lo\endcsname{+0.060}
\expandafter\def\csname papernumrt_P2A2_off_EC_all\endcsname{-2.669}
\expandafter\def\csname papernumrt_P2A2_off_EC_all_hi\endcsname{-2.007}
\expandafter\def\csname papernumrt_P2A2_off_EC_all_lo\endcsname{-3.373}
\expandafter\def\csname papernumrt_P2A2_off_EC_human_fail\endcsname{-6.303}
\expandafter\def\csname papernumrt_P2A2_off_EC_human_fail_hi\endcsname{-3.776}
\expandafter\def\csname papernumrt_P2A2_off_EC_human_fail_lo\endcsname{-8.842}
\expandafter\def\csname papernumrt_P2A2_off_EC_rest\endcsname{-2.067}
\expandafter\def\csname papernumrt_P2A2_off_EC_rest_hi\endcsname{-1.458}
\expandafter\def\csname papernumrt_P2A2_off_EC_rest_lo\endcsname{-2.717}
\expandafter\def\csname papernumrt_P2A2_off_EPDMS_all\endcsname{+0.101}
\expandafter\def\csname papernumrt_P2A2_off_EPDMS_all_hi\endcsname{+0.357}
\expandafter\def\csname papernumrt_P2A2_off_EPDMS_all_lo\endcsname{-0.141}
\expandafter\def\csname papernumrt_P2A2_off_EPDMS_human_fail\endcsname{+0.416}
\expandafter\def\csname papernumrt_P2A2_off_EPDMS_human_fail_hi\endcsname{+1.163}
\expandafter\def\csname papernumrt_P2A2_off_EPDMS_human_fail_lo\endcsname{-0.290}
\expandafter\def\csname papernumrt_P2A2_off_EPDMS_rest\endcsname{+0.045}
\expandafter\def\csname papernumrt_P2A2_off_EPDMS_rest_hi\endcsname{+0.313}
\expandafter\def\csname papernumrt_P2A2_off_EPDMS_rest_lo\endcsname{-0.212}
\expandafter\def\csname papernumrt_P2A2_off_LK_all\endcsname{+1.037}
\expandafter\def\csname papernumrt_P2A2_off_LK_all_hi\endcsname{+1.381}
\expandafter\def\csname papernumrt_P2A2_off_LK_all_lo\endcsname{+0.724}
\expandafter\def\csname papernumrt_P2A2_off_LK_human_fail\endcsname{+3.414}
\expandafter\def\csname papernumrt_P2A2_off_LK_human_fail_hi\endcsname{+4.694}
\expandafter\def\csname papernumrt_P2A2_off_LK_human_fail_lo\endcsname{+2.131}
\expandafter\def\csname papernumrt_P2A2_off_LK_rest\endcsname{+0.620}
\expandafter\def\csname papernumrt_P2A2_off_LK_rest_hi\endcsname{+0.983}
\expandafter\def\csname papernumrt_P2A2_off_LK_rest_lo\endcsname{+0.322}
\expandafter\def\csname papernumrt_P2A2_on_EPDMS_all\endcsname{+0.018}
\expandafter\def\csname papernumrt_P2A2_on_EPDMS_all_hi\endcsname{+0.281}
\expandafter\def\csname papernumrt_P2A2_on_EPDMS_all_lo\endcsname{-0.235}
\expandafter\def\csname papernumrt_P2A2_on_EPDMS_human_fail\endcsname{-0.139}
\expandafter\def\csname papernumrt_P2A2_on_EPDMS_human_fail_hi\endcsname{+0.648}
\expandafter\def\csname papernumrt_P2A2_on_EPDMS_human_fail_lo\endcsname{-0.892}
\expandafter\def\csname papernumrt_P2A2_on_EPDMS_rest\endcsname{+0.045}
\expandafter\def\csname papernumrt_P2A2_on_EPDMS_rest_hi\endcsname{+0.313}
\expandafter\def\csname papernumrt_P2A2_on_EPDMS_rest_lo\endcsname{-0.212}
\expandafter\def\csname papernumrt_P2A2_on_LK_all\endcsname{+0.543}
\expandafter\def\csname papernumrt_P2A2_on_LK_all_hi\endcsname{+0.853}
\expandafter\def\csname papernumrt_P2A2_on_LK_all_lo\endcsname{+0.286}
\expandafter\def\csname papernumrt_P2A2_on_LK_human_fail\endcsname{+0.110}
\expandafter\def\csname papernumrt_P2A2_on_LK_human_fail_hi\endcsname{+0.381}
\expandafter\def\csname papernumrt_P2A2_on_LK_human_fail_lo\endcsname{-0.142}
\expandafter\def\csname papernumrt_P2A2_on_LK_rest\endcsname{+0.620}
\expandafter\def\csname papernumrt_P2A2_on_LK_rest_hi\endcsname{+0.983}
\expandafter\def\csname papernumrt_P2A2_on_LK_rest_lo\endcsname{+0.322}
\expandafter\def\csname papernumrt_abs_A0_off_EC\endcsname{79.562}
\expandafter\def\csname papernumrt_abs_A0_off_EPDMS\endcsname{86.283}
\expandafter\def\csname papernumrt_abs_A0_off_LK\endcsname{86.102}
\expandafter\def\csname papernumrt_abs_A0_on_EPDMS\endcsname{89.842}
\expandafter\def\csname papernumrt_abs_A0_on_LK\endcsname{96.583}
\expandafter\def\csname papernumrt_abs_A1_off_EC\endcsname{78.357}
\expandafter\def\csname papernumrt_abs_A1_off_EPDMS\endcsname{86.480}
\expandafter\def\csname papernumrt_abs_A1_off_LK\endcsname{86.218}
\expandafter\def\csname papernumrt_abs_A1_on_EPDMS\endcsname{89.988}
\expandafter\def\csname papernumrt_abs_A1_on_LK\endcsname{96.666}
\expandafter\def\csname papernumrt_abs_A2_off_EC\endcsname{78.934}
\expandafter\def\csname papernumrt_abs_A2_off_EPDMS\endcsname{86.218}
\expandafter\def\csname papernumrt_abs_A2_off_LK\endcsname{86.078}
\expandafter\def\csname papernumrt_abs_A2_on_EPDMS\endcsname{89.747}
\expandafter\def\csname papernumrt_abs_A2_on_LK\endcsname{96.715}
\expandafter\def\csname papernumrt_abs_L0_off_EC\endcsname{79.104}
\expandafter\def\csname papernumrt_abs_L0_off_EPDMS\endcsname{86.237}
\expandafter\def\csname papernumrt_abs_L0_off_LK\endcsname{84.884}
\expandafter\def\csname papernumrt_abs_L0_on_EPDMS\endcsname{89.770}
\expandafter\def\csname papernumrt_abs_L0_on_LK\endcsname{95.612}
\expandafter\def\csname papernumrt_abs_M0_off_EC\endcsname{79.253}
\expandafter\def\csname papernumrt_abs_M0_off_EPDMS\endcsname{86.456}
\expandafter\def\csname papernumrt_abs_M0_off_LK\endcsname{86.407}
\expandafter\def\csname papernumrt_abs_M0_on_EPDMS\endcsname{89.889}
\expandafter\def\csname papernumrt_abs_M0_on_LK\endcsname{96.830}
\expandafter\def\csname papernumrt_abs_M1_off_EC\endcsname{78.317}
\expandafter\def\csname papernumrt_abs_M1_off_EPDMS\endcsname{86.446}
\expandafter\def\csname papernumrt_abs_M1_off_LK\endcsname{86.448}
\expandafter\def\csname papernumrt_abs_M1_on_EPDMS\endcsname{89.973}
\expandafter\def\csname papernumrt_abs_M1_on_LK\endcsname{96.830}
\expandafter\def\csname papernumrt_abs_M2_off_EC\endcsname{78.347}
\expandafter\def\csname papernumrt_abs_M2_off_EPDMS\endcsname{86.362}
\expandafter\def\csname papernumrt_abs_M2_off_LK\endcsname{86.358}
\expandafter\def\csname papernumrt_abs_M2_on_EPDMS\endcsname{89.886}
\expandafter\def\csname papernumrt_abs_M2_on_LK\endcsname{96.871}
\expandafter\def\csname papernumrt_abs_P0_off_EC\endcsname{76.853}
\expandafter\def\csname papernumrt_abs_P0_off_EPDMS\endcsname{86.039}
\expandafter\def\csname papernumrt_abs_P0_off_LK\endcsname{87.033}
\expandafter\def\csname papernumrt_abs_P0_on_EPDMS\endcsname{89.514}
\expandafter\def\csname papernumrt_abs_P0_on_LK\endcsname{97.192}
\expandafter\def\csname papernumrt_abs_P1_off_EC\endcsname{76.773}
\expandafter\def\csname papernumrt_abs_P1_off_EPDMS\endcsname{86.270}
\expandafter\def\csname papernumrt_abs_P1_off_LK\endcsname{86.983}
\expandafter\def\csname papernumrt_abs_P1_on_EPDMS\endcsname{89.739}
\expandafter\def\csname papernumrt_abs_P1_on_LK\endcsname{96.978}
\expandafter\def\csname papernumrt_abs_P2_off_EC\endcsname{76.265}
\expandafter\def\csname papernumrt_abs_P2_off_EPDMS\endcsname{86.319}
\expandafter\def\csname papernumrt_abs_P2_off_LK\endcsname{87.115}
\expandafter\def\csname papernumrt_abs_P2_on_EPDMS\endcsname{89.765}
\expandafter\def\csname papernumrt_abs_P2_on_LK\endcsname{97.258}
\expandafter\def\csname papernumrt_abs_released_off_EC\endcsname{75.926}
\expandafter\def\csname papernumrt_abs_released_off_EPDMS\endcsname{86.560}
\expandafter\def\csname papernumrt_abs_released_off_LK\endcsname{85.790}
\expandafter\def\csname papernumrt_abs_released_on_EPDMS\endcsname{90.038}
\expandafter\def\csname papernumrt_abs_released_on_LK\endcsname{96.460}
\expandafter\def\csname papernumrt_bootstrap\endcsname{10,000}
\expandafter\def\csname papernumrt_confidence\endcsname{95}
\expandafter\def\csname papernumrt_dist_F_ADE_max\endcsname{0.084}
\expandafter\def\csname papernumrt_dist_F_ADE_min\endcsname{0.050}
\expandafter\def\csname papernumrt_dist_P0A0_ADE_F\endcsname{+0.084}
\expandafter\def\csname papernumrt_dist_P0A0_ADE_FR\endcsname{+0.049}
\expandafter\def\csname papernumrt_dist_P0A0_ADE_FR_hi\endcsname{+0.075}
\expandafter\def\csname papernumrt_dist_P0A0_ADE_FR_lo\endcsname{+0.024}
\expandafter\def\csname papernumrt_dist_P0A0_ADE_F_hi\endcsname{+0.108}
\expandafter\def\csname papernumrt_dist_P0A0_ADE_F_lo\endcsname{+0.062}
\expandafter\def\csname papernumrt_dist_P0A0_ADE_R\endcsname{+0.034}
\expandafter\def\csname papernumrt_dist_P0A0_ADE_R_hi\endcsname{+0.048}
\expandafter\def\csname papernumrt_dist_P0A0_ADE_R_lo\endcsname{+0.023}
\expandafter\def\csname papernumrt_dist_P0A0_FDE_F\endcsname{+0.201}
\expandafter\def\csname papernumrt_dist_P0A0_FDE_FR\endcsname{+0.106}
\expandafter\def\csname papernumrt_dist_P0A0_FDE_FR_hi\endcsname{+0.177}
\expandafter\def\csname papernumrt_dist_P0A0_FDE_FR_lo\endcsname{+0.041}
\expandafter\def\csname papernumrt_dist_P0A0_FDE_F_hi\endcsname{+0.267}
\expandafter\def\csname papernumrt_dist_P0A0_FDE_F_lo\endcsname{+0.142}
\expandafter\def\csname papernumrt_dist_P0A0_FDE_R\endcsname{+0.095}
\expandafter\def\csname papernumrt_dist_P0A0_FDE_R_hi\endcsname{+0.125}
\expandafter\def\csname papernumrt_dist_P0A0_FDE_R_lo\endcsname{+0.068}
\expandafter\def\csname papernumrt_dist_P0A0_MAXLAT_F\endcsname{+0.124}
\expandafter\def\csname papernumrt_dist_P0A0_MAXLAT_FR\endcsname{+0.087}
\expandafter\def\csname papernumrt_dist_P0A0_MAXLAT_FR_hi\endcsname{+0.131}
\expandafter\def\csname papernumrt_dist_P0A0_MAXLAT_FR_lo\endcsname{+0.047}
\expandafter\def\csname papernumrt_dist_P0A0_MAXLAT_F_hi\endcsname{+0.166}
\expandafter\def\csname papernumrt_dist_P0A0_MAXLAT_F_lo\endcsname{+0.086}
\expandafter\def\csname papernumrt_dist_P0A0_MAXLAT_R\endcsname{+0.036}
\expandafter\def\csname papernumrt_dist_P0A0_MAXLAT_R_hi\endcsname{+0.050}
\expandafter\def\csname papernumrt_dist_P0A0_MAXLAT_R_lo\endcsname{+0.024}
\expandafter\def\csname papernumrt_dist_P1A1_ADE_F\endcsname{+0.054}
\expandafter\def\csname papernumrt_dist_P1A1_ADE_FR\endcsname{+0.047}
\expandafter\def\csname papernumrt_dist_P1A1_ADE_FR_hi\endcsname{+0.070}
\expandafter\def\csname papernumrt_dist_P1A1_ADE_FR_lo\endcsname{+0.026}
\expandafter\def\csname papernumrt_dist_P1A1_ADE_F_hi\endcsname{+0.074}
\expandafter\def\csname papernumrt_dist_P1A1_ADE_F_lo\endcsname{+0.035}
\expandafter\def\csname papernumrt_dist_P1A1_ADE_R\endcsname{+0.007}
\expandafter\def\csname papernumrt_dist_P1A1_ADE_R_hi\endcsname{+0.017}
\expandafter\def\csname papernumrt_dist_P1A1_ADE_R_lo\endcsname{-0.003}
\expandafter\def\csname papernumrt_dist_P1A1_FDE_F\endcsname{+0.123}
\expandafter\def\csname papernumrt_dist_P1A1_FDE_FR\endcsname{+0.105}
\expandafter\def\csname papernumrt_dist_P1A1_FDE_FR_hi\endcsname{+0.161}
\expandafter\def\csname papernumrt_dist_P1A1_FDE_FR_lo\endcsname{+0.051}
\expandafter\def\csname papernumrt_dist_P1A1_FDE_F_hi\endcsname{+0.175}
\expandafter\def\csname papernumrt_dist_P1A1_FDE_F_lo\endcsname{+0.075}
\expandafter\def\csname papernumrt_dist_P1A1_FDE_R\endcsname{+0.018}
\expandafter\def\csname papernumrt_dist_P1A1_FDE_R_hi\endcsname{+0.041}
\expandafter\def\csname papernumrt_dist_P1A1_FDE_R_lo\endcsname{-0.005}
\expandafter\def\csname papernumrt_dist_P1A1_MAXLAT_F\endcsname{+0.131}
\expandafter\def\csname papernumrt_dist_P1A1_MAXLAT_FR\endcsname{+0.117}
\expandafter\def\csname papernumrt_dist_P1A1_MAXLAT_FR_hi\endcsname{+0.157}
\expandafter\def\csname papernumrt_dist_P1A1_MAXLAT_FR_lo\endcsname{+0.076}
\expandafter\def\csname papernumrt_dist_P1A1_MAXLAT_F_hi\endcsname{+0.168}
\expandafter\def\csname papernumrt_dist_P1A1_MAXLAT_F_lo\endcsname{+0.092}
\expandafter\def\csname papernumrt_dist_P1A1_MAXLAT_R\endcsname{+0.014}
\expandafter\def\csname papernumrt_dist_P1A1_MAXLAT_R_hi\endcsname{+0.029}
\expandafter\def\csname papernumrt_dist_P1A1_MAXLAT_R_lo\endcsname{-0.002}
\expandafter\def\csname papernumrt_dist_P2A2_ADE_F\endcsname{+0.050}
\expandafter\def\csname papernumrt_dist_P2A2_ADE_FR\endcsname{+0.027}
\expandafter\def\csname papernumrt_dist_P2A2_ADE_FR_hi\endcsname{+0.050}
\expandafter\def\csname papernumrt_dist_P2A2_ADE_FR_lo\endcsname{+0.001}
\expandafter\def\csname papernumrt_dist_P2A2_ADE_F_hi\endcsname{+0.070}
\expandafter\def\csname papernumrt_dist_P2A2_ADE_F_lo\endcsname{+0.030}
\expandafter\def\csname papernumrt_dist_P2A2_ADE_R\endcsname{+0.023}
\expandafter\def\csname papernumrt_dist_P2A2_ADE_R_hi\endcsname{+0.041}
\expandafter\def\csname papernumrt_dist_P2A2_ADE_R_lo\endcsname{+0.009}
\expandafter\def\csname papernumrt_dist_P2A2_FDE_F\endcsname{+0.090}
\expandafter\def\csname papernumrt_dist_P2A2_FDE_FR\endcsname{+0.055}
\expandafter\def\csname papernumrt_dist_P2A2_FDE_FR_hi\endcsname{+0.113}
\expandafter\def\csname papernumrt_dist_P2A2_FDE_FR_lo\endcsname{-0.009}
\expandafter\def\csname papernumrt_dist_P2A2_FDE_F_hi\endcsname{+0.141}
\expandafter\def\csname papernumrt_dist_P2A2_FDE_F_lo\endcsname{+0.038}
\expandafter\def\csname papernumrt_dist_P2A2_FDE_R\endcsname{+0.036}
\expandafter\def\csname papernumrt_dist_P2A2_FDE_R_hi\endcsname{+0.074}
\expandafter\def\csname papernumrt_dist_P2A2_FDE_R_lo\endcsname{+0.004}
\expandafter\def\csname papernumrt_dist_P2A2_MAXLAT_F\endcsname{+0.039}
\expandafter\def\csname papernumrt_dist_P2A2_MAXLAT_FR\endcsname{+0.080}
\expandafter\def\csname papernumrt_dist_P2A2_MAXLAT_FR_hi\endcsname{+0.121}
\expandafter\def\csname papernumrt_dist_P2A2_MAXLAT_FR_lo\endcsname{+0.037}
\expandafter\def\csname papernumrt_dist_P2A2_MAXLAT_F_hi\endcsname{+0.078}
\expandafter\def\csname papernumrt_dist_P2A2_MAXLAT_F_lo\endcsname{-0.001}
\expandafter\def\csname papernumrt_dist_P2A2_MAXLAT_R\endcsname{-0.041}
\expandafter\def\csname papernumrt_dist_P2A2_MAXLAT_R_hi\endcsname{-0.028}
\expandafter\def\csname papernumrt_dist_P2A2_MAXLAT_R_lo\endcsname{-0.053}
\expandafter\def\csname papernumrt_ec_n_all\endcsname{10,040}
\expandafter\def\csname papernumrt_ec_n_human_fail\endcsname{1,428}
\expandafter\def\csname papernumrt_ec_n_rest\endcsname{8,612}
\expandafter\def\csname papernumrt_initial_step\endcsname{4,800}
\expandafter\def\csname papernumrt_logs\endcsname{136}
\expandafter\def\csname papernumrt_n_human_fail\endcsname{1,816}
\expandafter\def\csname papernumrt_n_rest\endcsname{10,330}
\expandafter\def\csname papernumrt_seed0\endcsname{0}
\expandafter\def\csname papernumrt_seed1\endcsname{1}
\expandafter\def\csname papernumrt_seed2\endcsname{2}
\expandafter\def\csname papernumrt_seeds\endcsname{3}
\expandafter\def\csname papernumrt_spread_three_seed_off_EC_all\endcsname{1.205}
\expandafter\def\csname papernumrt_spread_three_seed_off_EPDMS_all\endcsname{0.280}
\expandafter\def\csname papernumrt_spread_three_seed_off_LK_all\endcsname{0.140}
\expandafter\def\csname papernumrt_spread_three_seed_on_EPDMS_all\endcsname{0.251}
\expandafter\def\csname papernumrt_spread_three_seed_on_LK_all\endcsname{0.280}
\expandafter\def\csname papernumrt_spread_two_seed_off_EC_all\endcsname{1.205}
\expandafter\def\csname papernumrt_spread_two_seed_off_EPDMS_all\endcsname{0.231}
\expandafter\def\csname papernumrt_spread_two_seed_off_LK_all\endcsname{0.115}
\expandafter\def\csname papernumrt_spread_two_seed_on_EPDMS_all\endcsname{0.225}
\expandafter\def\csname papernumrt_spread_two_seed_on_LK_all\endcsname{0.214}
\expandafter\def\csname papernumrt_steps\endcsname{300}
\expandafter\def\csname papernumrt_tokens\endcsname{12,146}
\expandafter\def\csname papernumsolver_identical\endcsname{103,288}
\expandafter\def\csname papernumsolver_worst_conditioned\endcsname{1,000}
\expandafter\def\csname papernumspr_sample_reproduced\endcsname{0}
\expandafter\def\csname papernumspr_sample_tokens\endcsname{260}
\expandafter\def\csname papernumspr_default_reproduced\endcsname{0}
\expandafter\def\csname papernumspr_default_tokens\endcsname{4,992}
\expandafter\def\csname papernumspr_noforgiveness_tokens\endcsname{832}
\expandafter\def\csname papernumsupervision_entries_exact\endcsname{170,049,218}
\expandafter\def\csname papernumsupervision_share_exact_percent\endcsname{1.4355}
\expandafter\def\csname papernumtarget_entries\endcsname{11,845,894,144}
\expandafter\def\csname papernumties\endcsname{2}
\expandafter\def\csname papernumties_draws\endcsname{100}
\expandafter\def\csname papernumties_zero_released\endcsname{2,533}
\expandafter\def\csname papernumvocab_dense\endcsname{16,384}

\title{When an Evaluation Rule Writes Training Labels:\\ Measuring Human-Reference\\ Forgiveness in NAVSIM}
\author{Jiaxuan Guo$^{1}$\thanks{Corresponding author.}\hphantom{$^{*}$} \enspace Jingxin Yang$^{1}$ \enspace Jiaqi Ye$^{1}$ \enspace Youran Sun$^{2}$ \\
\textbf{Shuo Xin$^{1}$ \enspace Kejia Zhang$^{2}$ \enspace Haizhao Yang$^{2}$} \\
$^{1}$Stanford University \quad $^{2}$University of Maryland, College Park \\
\texttt{guojx@stanford.edu}}
\begin{document}
\maketitle
\lhead{}\renewcommand{\headrulewidth}{0pt}
\begin{abstract}
When the human reference scores zero on a metric, the released GTRS-Dense label generator for NAVSIM marks every candidate trajectory in the scene as passing it.
NAVSIM's authors introduced this human-reference forgiveness to avoid penalizing contextually justified maneuvers when scoring one trajectory, and warned that it could overlook important failures.
In label generation it sets a whole column of \N{vocab_dense} candidate targets to passing.
To measure the consequences for supervision, we re-run the generator with the overwrite disabled and compare the pre-overwrite targets with the released labels on all \N{population_tokens} navtrain scenes.
The rule erases a candidate distinction that the training loss reads on \N{headline_count} of them (\N{headline_percent}\%).
Firing usually changes most of a column: lane keeping carries \N{LK_forgiven} of the 13,042 forgiven loss columns, and its median forgiven column had \N{LK_pooled_median} of \N{vocab_dense} candidates failing before the overwrite.
On held-out navtest scenes forgiven on lane keeping, the released lane-keeping head's median AUC against the pre-overwrite outcome is \N{nt_released_R1}; on unforgiven scenes matched on failing-candidate count it is \N{nt_released_R2}.
For the Hydra-MDP checkpoint released with GTRS, whose configuration takes the same label file, the two values are \N{hmdp_nt_R1} and \N{hmdp_nt_R2}.
Continuing the released GTRS-Dense checkpoint for \N{rt_steps} optimizer steps with three paired seeds, we observe the forgiven-scene AUC \N{nt_PA_R1_min}--\N{nt_PA_R1_max} higher with pre-overwrite than with published targets, and a narrower gap between matched groups, still above zero.
Scoring with forgiveness disabled, we observe lane keeping higher by \N{rt_LK_forgiven_min}--\N{rt_LK_forgiven_max} points on navtest scenes forgiven on any of five loss metrics, with lower adjacent-frame plan consistency.
Both changes are larger there than on the rest.
EPDMS, scored the same way, does not separate the two target sets.
\end{abstract}
\section{Introduction}
\label{sec:introduction}
NAVSIM's authors introduced human-reference filtering to avoid penalizing contextually justified maneuvers and warned that it could overlook important failures and edge cases \citep[Section~3.1; main-text Limitations, ``Human Flag Filtering'']{pseudosim}.
They called for further study of that trade-off, which concerns evaluation reliability, without measuring its effect on training labels; we measure that effect for the released GTRS-Dense trajectory labels.
The measurement has two parts: how often the rule fires, and how many of the \N{vocab_dense} candidate targets in a column it changes when it does, which on this split is most of them.

The same rule serves a different consumer when a scorer generates training targets.
GTRS, whose dense scorer builds on Hydra-MDP \citep[Sections~2.3--2.4]{hydramdp}, learns component scores for candidate trajectories and
uses their predictions to select a plan \citep[Sections~2.2--2.3 and~3]{gtrs}.
Its label generator replaces a component by passing labels throughout a scene's vocabulary when the human reference fails that component.
An overwrite can change a target level without removing a target distinction: a column that already assigns every candidate the same failing value had no candidate ordering to erase.
We therefore distinguish a human-reference zero, an overwrite that changes a value, and an overwrite of a previously mixed column.
The last predicate is evaluated after the training consumer's target mapping and restricted to the columns its loss reads.

We also measure whether vocabulary dropout would retain the pre-overwrite distinction and compare distinct label rows before and after overwrite within each token.
We test the earlier split criterion, a v1 screen of navtrain on the human's PDM score (PDMS) rather than the rule measured here, with a separate v1 human census in Appendix~\ref{sec:splitcounterexamples}, and compare the top-ranked candidate sets of the two label files under a stated rule in Appendix~\ref{sec:labelranking}.

\paragraph{Contributions.}
We recover the pre-overwrite side of the released label file by re-running its generator with the forgiveness assignment disabled, and check that reconstruction against the published file on every column the overwrite did not touch (Section~\ref{sec:method}).
That pair of arrays carries a per-column magnitude, which the single scored trajectory per agent of \citet{scorebasis} cannot (Section~\ref{sec:related}), and separates the human-zero trigger from a changed target level and from an erased distinction, after the training consumer's target mapping and inside the columns its loss reads.
We read the released checkpoint itself on the forgiven scenes and report what it still recovers of the pre-overwrite distinction, beside a base-rate-matched control and the drivable-area head, whose column the rule never fires on (Section~\ref{sec:probe}); Appendix~\ref{sec:probeprecedents} gives the prior work behind the head comparison.
Continuing that checkpoint with matched seeds, we observe higher lane keeping on forgiven navtest scenes and lower adjacent-frame plan consistency under pre-overwrite than published targets, while the extended PDM score (EPDMS) with forgiveness disabled does not separate the supervision policies (Section~\ref{sec:retraining}).

\section{Related work}
\label{sec:related}
\paragraph{The rule and its evaluation-side audit.}
\citet[Section~3.1]{pseudosim} introduce and call novel the EPDMS filtering mechanism: a rule violation is forgiven when the human expert also commits it in the same scene.
\citet[Sections~3.3 and~4.2]{scorebasis} subsequently formalize reference-conditioned forgiveness and measure its evaluation-side interaction with numerical rollout failures.
Their published audit protocol recommends the failure-mask and overwrite-rate measurements used here, and they measure evaluation-side overwrites for DAC, DDC, LK, TTC, and NC on 12,146 navtest tokens \citep[Sections~4.2 and~5]{scorebasis}.
The per-channel overwrite rates below are that recommended measurement, carried out on a different artifact.

Their unit is one scored trajectory per agent, so a channel holds a single value; the label generator writes 16,384 candidate values per column, and their rate is the one-entry case of the per-column magnitude measured here.
The recommendation's stated purpose is to precede the interpretation of a ranking, whereas the object here is a released training-label file; their rates are measured under the rollout divergence they audit, and, with their thresholds read as deviation or step limits, no stock-solver human-reference rollout of this split exceeds them (Section~\ref{sec:method} and Appendix~\ref{app:solver}).

\paragraph{Label-error audits.}
Confident learning flags likely label errors from out-of-sample predicted probabilities \citep[Sections~2--3]{confidentlearning}, and \citet[Sections~3--4]{labelerrors} validate such flags in ten benchmarks with crowdworkers; there the unit is one example judged against a latent true class, here a column of candidate values against its regenerated pre-overwrite side.

\paragraph{Constant groups and ranking supervision.}
The phenomenon class is established in both reinforcement learning with verifiable rewards and learning-to-rank.
In GRPO, identical rewards yield zero group-relative advantages, as described by \citet[Section~3.4, ``Standard GRPO'' and ``Dynamic Pool Management'']{zerovariance}.
XGBoost skips equal-label pairs and defines effective pairs by their contribution to the ranking gradient \citep[``Constructing Pairs'' and ``Obtaining Good Result'']{xgboostltr}.
Its default nDCG convention assigns 1 to a query with no positive labels, illustrating why a passing aggregate value need not attest to a ranking distinction \citep[\code{eval_metric}, \code{ndcg-}]{xgboostparams}.
Within driving, FlowR2A replaces nearly constant v1 comfort labels with history comfort to obtain more ranking signal \citep[Appendix~F.2.4]{flowr2a}.
Unlike the group-relative reward term, GTRS's component BCE can still have prediction-dependent gradients when its targets are constant.

\paragraph{Abstention and positive supervision.}
If forgiveness expresses that a rule cannot judge a scene, abstaining from that rule and assigning a positive target express different supervision decisions.
Treating abstention as a first-class output of a labeling function is prior weak-supervision tooling \citep[Section~2]{ratner2017snorkel}.

\paragraph{Whether an item still carries a target distinction.}
\citet[Section~3, ``Data filtering'']{hiddenbiases} prune an end-to-end driving dataset by a heuristic that estimates whether a frame changes the model's target labels relative to the previous frame.
The predicate used here asks the same question of a different object: whether a forgiven column still separated candidates within one scene's vocabulary before it was overwritten, not whether a frame changes the targets relative to the previous frame.
Both test target variation to decide whether an item still carries supervision, there to drop samples and here to read what an overwrite removed; the constant-group logic itself is the learning-to-rank one named above.
The same work reports that adjusting how its CARLA expert brakes for pedestrians cut pedestrian collisions about fourfold for policies trained on the adjusted demonstrations, with the expert's own pedestrian collision rate unchanged \citep[Section~3, ``Expert style'']{hiddenbiases}.
There the reference supplies the action being imitated, whereas here it gates the metric that writes the targets; what carries across the two settings is that a reference-side choice its own score does not register can reach what a policy is trained on.

\paragraph{What this measurement adds.}
What none of them reports is how this rule behaves on this benchmark's released training labels: how often it fires, and how much of a candidate vocabulary it reverses when it does.

\section{Label generation and the training consumer}
\label{sec:background}
\begin{figure}[t]
\centering
\begin{tikzpicture}[x=1pt, y=1pt, font=\scriptsize,
    >={Stealth[length=3.4pt, width=3pt]},
    stage/.style={draw, semithick, rounded corners=1.6pt, align=center,
      inner xsep=2.2pt, inner ysep=2.4pt},
    artifact/.style={stage, rounded corners=0pt, fill=black!12},
    ours/.style={stage, dashed},
    flow/.style={->, semithick},
    ourflow/.style={->, semithick, dashed},
    compare/.style={<->, semithick, dashed},
    note/.style={inner sep=0pt, align=left},
    group/.style={font=\scriptsize\scshape, inner sep=0pt}]
  \node[stage, minimum width=60pt, anchor=west] (scene) at (0, 13)
    {Scene $t$ and its\\human trajectory};
  \node[stage, minimum width=60pt, anchor=west] (vocab) at (0, -13)
    {Vocabulary of\\16,384 candidates $i$};
  \path (scene.east |- 0, 0) coordinate (row);
  \node[stage, right=14pt of row] (scorer)
    {Scorer: human score $h_{tj}$\\and candidate targets $y_{tij}$\\in nine columns $j$:\\NC, DAC, DDC, TLC, EP,\\TTC, LK, HC, \code{pdm_score}};
  \node[stage, right=14pt of scorer] (forgive)
    {Forgiveness, Equation~\ref{eq:forgiveness}:\\if $h_{tj}=0$, column $j$\\becomes 1.0 for\\every candidate $i$};
  \node[artifact, right=14pt of forgive] (labels)
    {Released\\labels $\widetilde y_{tij}$};
  \draw[flow] (scene.east) -- (scene.east -| scorer.west);
  \draw[flow] (vocab.east) -- (vocab.east -| scorer.west);
  \draw[flow] (scorer) -- (forgive);
  \draw[flow] (forgive) -- (labels);
  \path (labels.east |- scene.north) coordinate (top);
  \node[stage, minimum width=79pt, anchor=north west] (drop) at ([xshift=18pt]top)
    {Vocabulary dropout keeps\\8,192 of 16,384 per step};
  \node[stage, minimum width=79pt, below=6pt of drop] (map)
    {Mapping $b_j$: $0.5\to0$\\in NC and DDC};
  \node[stage, minimum width=79pt, below=6pt of map] (loss)
    {BCE over the seven loss\\columns $J_L$: NC, DAC,\\DDC, TLC, EP, TTC, LK};
  \node[artifact, minimum width=79pt, below=6pt of loss] (ckpt)
    {Released checkpoint};
  \draw[flow] (labels.east) -- ++(9, 0) |- (drop.west);
  \draw[flow] (drop) -- (map);
  \draw[flow] (map) -- (loss);
  \draw[flow] (loss) -- (ckpt);
  \node[ours, anchor=west] (pre) at (forgive.west |- ckpt)
    {Section~\ref{sec:method}: pre-overwrite targets $y_{tij}$,\\regenerated with Equation~\ref{eq:forgiveness} disabled};
  \draw[ourflow] ([xshift=-18pt]scorer.south) |- (pre.west);
  \coordinate (c5) at ($(labels.west)!0.5!(pre.east)$);
  \draw[compare] (c5 |- labels.south) -- (c5 |- pre.north);
  \coordinate (n5) at ($(forgive.south)!0.5!(pre.north)$);
  \node[note, anchor=east] at ([xshift=-4pt]c5 |- n5)
    {Section~\ref{sec:results}: $y$ against $\widetilde y$ in each scene:\\changed targets; distinctions erased\\after $b_j$ on $J_L$; distinct label rows};
  \draw[compare] (ckpt.west) -- (ckpt.west -| pre.east);
  \coordinate (c6) at ($(ckpt.west)!0.5!(pre.east)$);
  \node[note, anchor=north] at ([yshift=-3pt]c6 |- pre.south)
    {Section~\ref{sec:probe}: lane-keeping head's ranking\\scored against pre-overwrite LK};
  \node[group, anchor=base west] at (0, 27) {label generation};
  \node[group, anchor=base west] at (drop.west |- 0, 27) {training consumer};
  \node[group, anchor=west] at (0, 0 |- pre) {measured here};
\end{tikzpicture}%
\caption{Label generation and the training consumer as released (solid), and the pre-overwrite side regenerated here with the comparisons made on it (dashed). Shaded boxes are the released label file and checkpoint.}
\label{fig:pipeline}
\end{figure}
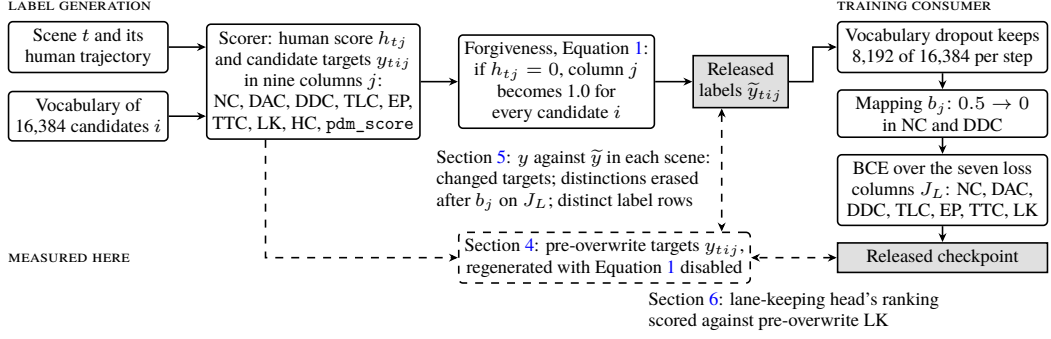

The audited GTRS-Dense file, \code{navtrain_16384.pkl}, has nine columns and 16,384 candidate entries per column: no-at-fault collisions (NC), drivable-area compliance (DAC), driving-direction compliance (DDC), traffic-light compliance (TLC), ego progress (EP), time to collision (TTC), lane keeping (LK), history comfort (HC), and the aggregate \code{pdm_score}.
LK fails a candidate that stays more than \N{flip_lk_limit_m}\,m from the route centerline for \N{flip_lk_steps_required} consecutive steps, skipping steps inside intersections (Appendix~\ref{app:flipped}).
Figure~\ref{fig:pipeline} follows one scene from the scorer that writes these columns to the loss that reads them.
The source revisions are NAVSIM \code{0a380a9} and GTRS \code{92a740de}; we bind NAVSIM \code{v2.0}'s DDC body in place of GTRS's, whose intersection exclusion can only raise DDC, as the released labels follow the earlier body \citep{navsimcode,gtrscode}.

\paragraph{The overwrite.}
The \code{human_penalty_filter} branch in \code{pdm_score_full_v2} executes \code{gt[column]} \texttt{=} \code{np.ones_like(gt[column])} at \code{navsim/evaluate/pdm_score.py:308} when the separate human-scoring call returns zero for that column \citep{gtrscode}.
The assignment follows conversion to float16 and removal of the internal reference row from the candidate arrays.
Writing $y_{tij}$ for the stored-precision pre-overwrite target, $h_{tj}$ for the human score, and $\widetilde y_{tij}$ for the published target, the rule is
\begin{equation}
\widetilde y_{tij}=\begin{cases}1.0,&h_{tj}=0,\\y_{tij},&h_{tj}\ne0.\end{cases}
\label{eq:forgiveness}
\end{equation}
The branch is enabled on every scene of the measured split.
The human reference used by this rule is separate from the internal reference used to normalize vocabulary progress.
The upstream Issue~\#151 fix excludes \code{pdm_score} from human filtering and rebuilds aggregate inputs in the evaluation entry \code{pdm_score_from_interpolated_trajectory} \citep{filterfix}.
The inspected GTRS label entry \code{pdm_score_full_v2} retains Equation~\ref{eq:forgiveness}, so that patch leaves the seven-column target overwrites measured here in place.

\paragraph{Loaded fields and loss terms.}
The loader takes the eight keys of \code{trajectory_pdm_weight} at \path{navsim/agents/gtrs_dense/gtrs_agent.py}:186 of \code{NVlabs/GTRS} commit \code{92a740de} and loads those fields at line~254; \code{pdm_score} is never loaded.
HC is loaded but omitted from the loss sum at lines~129--138, leaving seven loss columns, $J_L=\{\mathrm{NC,DAC,TTC,EP,DDC,LK,TLC}\}$.
The mapping $b_j$ sends 0.5 to 0.0 for NC and DDC and is the identity for the other columns.
A raw mixed column can therefore become constant before the loss reads it.
The loss adds component-wise BCE terms rather than multiplying the component targets \citep[\mbox{\code{hydra_kd_imi_agent_loss_dropout}}]{gtrscode}.

\paragraph{The earlier scene filter.}
NAVSIM documents screening its navtrain list offline by a human-PDMS threshold of 0.8 under a v1 aggregate that contains neither LK nor TLC \citep[Section~3.1, ``Filtering for challenging scenes'']{navsim}; Appendix~\ref{sec:splitcounterexamples} gives the filter's definition and tests it.

\section{Measurement method}
\label{sec:method}
\paragraph{Population and human census.}
The denominator is the benchmark's own \path{navsim/planning/script/config/common/train_test_split/scene_filter/navtrain.yaml} token list, exactly 103,288 scenes.
We use its complete original-scene population and the corresponding released GTRS-Dense labels.
The human-zero mask is the label entry's own human call, scored without history.
Human EP is identically 1.0 under singleton progress normalization, whereas human HC is identically 1.0 because this call omits history, so both human-zero sets are structurally empty.
DAC's empty human-zero set is empirical: the v1 human census contains zero DAC-zero scenes but 20 NC-zero scenes in the published navtrain list.
Both are multiplicative penalties, so enforcing the stated 0.8 criterion would remove both zero sets.
The surviving NC zeros separate DAC's observed emptiness from that filter prediction.

\paragraph{Solver sensitivity of the human census.}
The census uses \code{np.linalg.pinv}, the stock pseudoinverse, for which \citet[solver matrix in Section~4.2]{scorebasis} report a rollout-divergence rate of 99.8\% on their own 450-scene interactive pool.
Under the software configuration they audit, substituting \code{scipy.linalg.pinvh} or direct solve leaves all \N{solver_identical} tokens' nine human values bitwise identical.
Under the deviation-or-step reading, \N{div_tokens} of the \N{div_denominator} stock-solver human rollouts of this split exceed the $(20,10)$, $(100,50)$, and $(200,100)$\,m thresholds; under the path-length reading, at least 5,165 exceed the strict 20\,m limit and none exceeds 100\,m; forgiveness erases a distinction in a loss column on \N{headline_percent}\% of the same scenes (Appendix~\ref{app:solver}).
Neither these distances nor the solver table estimates a prevalence difference between the two corpora, whose channel sets, pools, and measured quantities all differ.

\paragraph{Separate the predicates.}
Let $J$ denote the nine stored columns and define $P_1(t)$ by the existence of $j\in J$ with $h_{tj}=0$.
Define $P_2(t)$ by a forgiven column containing at least one pre-overwrite value different from 1.0, and $P_3(t;K,b)$ by a forgiven column $j\in K$ containing at least two distinct values after mapping $b_j$.
The raw definition uses the identity mapping; the loss-effective definition uses the consumer mapping from Section~\ref{sec:background}.
Restricting to the loss columns removes 237 tokens: $J\setminus J_L$ holds only HC and \code{pdm_score} and the rule never fires on HC, so those 237 are exactly the scenes whose only loss-effective mixed forgiven column is the aggregate.
The main count is the token union $P_3(t;J_L,b)$.
For each forgiven column we record whether it was already all ones, constant at a failing value, or mixed before overwrite.
A column can lose distinctions without the scene's complete label matrix becoming constant.

\paragraph{Pre-overwrite reconstruction and paired rows.}
We regenerate the candidate arrays with the final forgiveness assignment disabled, retain the released float16 precision, and join the arrays to the published labels by token and candidate index.
The assignment is the generator's last step, so these values come from the computation that wrote the released file.
The pre-overwrite values appear in no released artifact, so we check the reconstruction several ways.
Within the $P_1$ pool, every column the overwrite did not touch is compared against the published file, and 97,173 of 97,175 comparisons are equal; both exceptions are the non-forgiven EP and aggregate columns of one token.
That denominator is $12{,}576\times9$ less the \N{forgiven_token_columns} forgiven token-columns.
A 60-token control pool on which the rule cannot fire is bitwise equal on all nine columns, \N{control_pool_comparisons} of \N{control_pool_comparisons} token-column comparisons.
The scorer's own per-metric output, captured in a separate run with forgiveness on, gives the same DDC value-set partition and the same DDC counts at the five ranks as the reconstruction (Appendix~\ref{app:conventions}).
A corpus-wide census of the two arrays finds a difference on all \N{paired_pool} tokens, and the nine tokens that differ without belonging to the $P_1$ pool differ only on ego progress, a column forgiveness never fires on, and the aggregate it feeds.
Run with forgiveness on, the same configuration reproduces all nine published columns byte for byte on \N{reproduced} of \N{population_tokens} tokens (Appendix~\ref{app:cache}), which also localizes the cache dependency behind \N{cache_residuals} of the \N{residual_total} residuals.
We count distinct rows in three column representations: all nine raw columns, the seven raw loss columns, and those seven columns after NC/DDC binarization.

\section{Forgiveness in the released training labels}
\label{sec:results}
\label{sec:targetcounts}
\subsection{Every trigger changes a target, and most also erase a distinction}
Table~\ref{tab:predicates} separates the human-zero trigger from its two possible label consequences.
To scale its last row we measured a second rewrite these labels undergo on the same 103,288 scenes: before the BCE the loss maps 0.5 to 0 in the collision and driving-direction columns, and on 260 scenes, 0.2517\%, that mapping leaves a loss column constant which the published values had left mixed.
On this split, forgiveness erases distinctions on 43.2$\times$ as many scenes as the NC/DDC target mapping.
The two overlap on 73 scenes.

\begin{table}[t]
\centering\small
\caption{Nested token predicates over the official 103,288-scene navtrain split. $P_1$ and $P_2$ share a row because no forgiven column is already all-passing before the overwrite.}
\label{tab:predicates}
\begin{tabularx}{\linewidth}{@{}Xrr@{}}
\toprule
Predicate & Tokens & Fraction (\%) \\
\midrule
$P_1=P_2$: human-zero trigger, and the overwrite changes a value & 12,576 & 12.1757 \\
$P_3$: raw mixed, any stored column & 11,503 & 11.1368 \\
$P_3$: loss-effective mixed, any stored column & 11,474 & 11.1087 \\
\textbf{$P_3$: loss-effective mixed, a loss column} & \textbf{11,237} & \textbf{10.8793} \\
\bottomrule
\end{tabularx}

\end{table}

\paragraph{The trigger union is preserved structurally.}
The 2,967 aggregate-column forgivenesses are not read by this loss, but cannot be the sole trigger: human EP and HC are both 1.0, making the human weighted factor positive, so a zero aggregate requires a zero in NC, DAC, DDC, or TLC, all of which enter the loss.
HC itself cannot trigger forgiveness, so restricting the trigger to the seven loss columns necessarily retains all \N{P1_count} members of $P_1$.

\subsection{Lane keeping accounts for most forgiven loss columns, and firing usually reverses most of a column}
\label{sec:columnreadings}
\begin{table}[t]
\centering\small
\caption{Pre-overwrite columns after the consumer mapping. ``Mixed'' means at least two distinct target values, and the last column pools every forgiven token-column of that column, all-zero and mixed together, and gives the upper median number of candidates that failed it before the overwrite (Appendix~\ref{app:conventions}). The aggregate is a diagnostic field this consumer's loss does not load.}
\label{tab:columns}
\setlength{\tabcolsep}{4pt}\begin{tabular}{@{}lrrrrrr@{}}
\toprule
Column & Forgiven & Already one & All zero & Mixed & \shortstack{Mixed\\(\% of forgiven)} & \shortstack{Pooled median\\failing} \\
\midrule
LK & 9,982 & 0 & 151 & 9,831 & 98.49 & 14,391 \\
TLC & 2,156 & 0 & 1,057 & 1,099 & 50.97 & 16,380 \\
DDC & 876 & 0 & 386 & 490 & 55.94 & 16,317 \\
TTC & 18 & 0 & 9 & 9 & 50.00 & 16,384 \\
NC & 10 & 0 & 5 & 5 & 50.00 & 16,384 \\
DAC (empirical empty set) & 0 & 0 & 0 & 0 & --- & --- \\
EP (structural empty set) & 0 & 0 & 0 & 0 & --- & --- \\
HC (structural empty set) & 0 & 0 & 0 & 0 & --- & --- \\
\midrule
\textit{Seven loss columns} & \textit{13,042} & \textit{0} & \textit{1,608} & \textit{11,434} & & \\
\midrule
\code{pdm_score} (unloaded) & 2,967 & 0 & 1,179 & 1,788 & 60.26 & 16,384 \\
\bottomrule
\end{tabular}

\end{table}

\begin{figure}[t]
\centering
\includegraphics{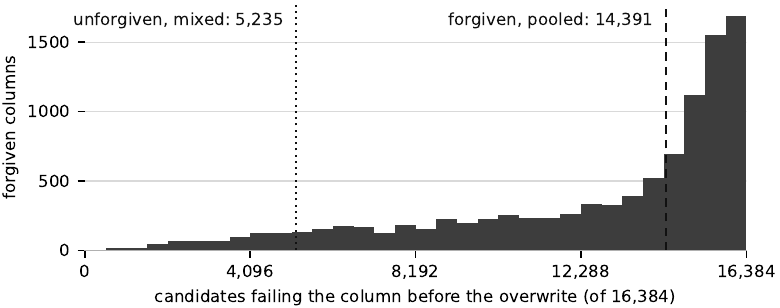}
\caption{Forgiven lane-keeping columns by how many of the 16,384 candidates failed the column before the overwrite, one count per column ($n=\N{LK_forgiven}$, bin width 512). The mass sits against the upper end: where the rule fires on this column, it usually replaces one in which most candidates were failing. The dotted line is the same count where it does not fire, median 5,235 over the 86,107 scenes with a mixed published column; the corresponding mixed-column median on the forgiven side is \N{LK_mixed_median}, since the histogram also carries the \N{LK_all_zero} columns that were already constant. Lane keeping lies below each of the other four firing loss columns at the lower four quantiles (Appendix~\ref{app:conventions}).}
\label{fig:lk}
\end{figure}

LK forgiveness accounts for \N{LK_forgiven_share}\% of the $P_1$ pool.
On \N{LK_mixed} navtrain scenes (\N{LK_mixed_percent}\%), LK is overwritten to passing for all 16,384 candidates despite distinguishing passing and violating candidates before overwrite.
LK's observed value sets are exactly $\{0.0,1.0\}$ for the mixed and $\{0.0\}$ for the all-zero columns of Table~\ref{tab:columns}; LK bypasses binarization, so its raw and loss-effective classifications coincide.
Among mixed LK columns ($n=\N{LK_mixed}$), violating candidates have minimum \N{LK_mixed_min}, median \N{LK_mixed_median} of 16,384 (\N{LK_mixed_median_percent}\%), and maximum \N{LK_mixed_max}; including the \N{LK_all_zero} all-zero columns gives a forgiven-pool median of \N{LK_pooled_median} ($n=\N{LK_forgiven}$).
Figure~\ref{fig:lk} shows the whole distribution.
A candidate that fails lane keeping keeps its failing target on the \N{probe_U_scenes} unforgiven scenes and receives a passing one on the \N{LK_mixed} forgiven ones; in the scorer's own rollouts the entries turned to passing reach a median of \N{flip_T_devC_median_m}\,m from the route centerline at their farthest, against \N{flip_U_devC_median_m}\,m for those kept failing, and their longest run beyond the lane-keeping limit has a median of \N{flip_T_runC_median_steps} against \N{flip_U_runC_median_steps} of \N{flip_rollout_steps} steps (Appendix~\ref{app:flipped}).

\paragraph{Scale in supervision entries.}
Changed LK targets total \N{LK_changed_entries} entries, \N{LK_mixed_changed_entries} in mixed and \N{LK_all_zero_changed_entries} in all-zero columns, out of $103{,}288\times16{,}384\times7=\N{target_entries}$ target entries.
The changed entries in mixed LK columns alone are \N{LK_mixed_entry_percent}\% of all target entries.
The other four triggered loss columns add \N{other_changed_entries_exact}, so the rule changes \N{supervision_entries_exact} target entries in the seven loss columns, \N{supervision_share_exact_percent}\% of them (Appendix~\ref{app:conventions}).
The released BCE weights for NC, DAC, TTC, EP, DDC, LK and TLC are $(3,3,4,2,1,2,3)$, so LK has $2/18$ of their sum; a separate imitation term enters with \mbox{\code{trajectory_imi_weight}} (default 1.0).
These entry counts and loss weights do not determine gradient shares (Section~\ref{sec:related}).

\paragraph{Raw and loss-effective readings.}
Only NC and DDC change under the half-to-zero mapping: mixed NC columns stay at five, while mixed DDC columns fall from 863 of 876 raw (98.52\%) to 490 of 876 loss-effective (55.94\%).
Forgiveness also fires on the v1 channels NC and TTC, on 10 and 18 tokens respectively, so its observed support is not confined to later-added channels.

Under the released vocabulary dropout, both pre-overwrite target values survive with mean probability \N{exposure_mean_probability}, weighting each of the \N{exposure_population} mixed forgiven loss token-columns equally (Appendix~\ref{sec:dropoutexposure}).
The repository's Hydra-MDP configuration takes the same file as its training labels with vocabulary dropout off, so every candidate enters its loss \citep[\mbox{\code{hydra_mdp_vov.yaml}}]{gtrscode}.

\subsection{Overwrite merges label rows on most tokens}
Table~\ref{tab:rows} compares each token with itself, using the reconstructed pre-overwrite and published arrays defined in Section~\ref{sec:method}; pairing within a token is what keeps selection into forgiveness from producing the effect, since both sides of every comparison are the same scene.
The loss-column projection exposes more row mergers than the full stored representation, whose extra coordinates can keep apart rows the projection merges.
The 90,712 untouched tokens have median 1,627 and mean 1,692.52 distinct nine-column rows; the forgiven pool already has approximately 300 fewer at the median before overwrite (Table~\ref{tab:rows}).
This between-pool contrast describes selection into forgiveness.

\begin{table}[t]
\centering\small
\caption{Paired distinct-row counts over the $P_1$ pool of \N{paired_pool} tokens. Medians are computed separately before and after overwrite; the mean drop averages per-token differences. Dropping the one token whose untouched columns differ (Section~\ref{sec:method}) lowers each strict-decrease count by one and moves no entry by more than 0.0089\%.}
\label{tab:rows}
\begin{tabular}{@{}lrrrr@{}}
\toprule
Representation & Pre median & Post median & Mean drop & Strict decreases \\
\midrule
All nine raw columns & 1,318 & 1,293 & 20.50 & 11,499 (91.44\%) \\
Seven raw loss columns & 1,214 & 1,148 & 121.54 & 11,502 (91.46\%) \\
Seven loss-effective columns (binarized) & 1,205 & 1,142 & 118.99 & 11,236 (89.34\%) \\
\bottomrule
\end{tabular}

\end{table}

In all three representations, no token has a single label row before or after overwrite.
After loss binarization, the 11,236 tokens with fewer distinct rows all lie in the \N{headline_count} of Table~\ref{tab:predicates}'s last row; in the remaining token a forgiven loss column is mixed, but its distinction is fully redundant with the other six columns, so the number of distinct rows is unchanged.

\section{What the released weights recover where the rule fired}
\label{sec:probe}
We also read the released GTRS-Dense checkpoint \code{gtrs_dense_vov.ckpt}, the artifact the publisher's training configuration pairs with this label file, ranking every scene's \N{vocab_dense} candidates by the lane-keeping head's logit and scoring that ranking against the pre-overwrite lane-keeping outcome.
Medians over the \N{LK_mixed} mixed LK-forgiven scenes, 95\% intervals bracketed and rounded outward: AUC \N{probe_T_lk_auc} [\N{probe_T_lk_auc_lo}, \N{probe_T_lk_auc_hi}], average precision \N{probe_T_lk_ap}, McClish partial AUC below a 0.1 false-positive rate \N{probe_T_lk_pauc}.
Its fifth and twenty-fifth percentiles both sit exactly at the McClish floor \N{probe_pauc_floor}, which a partial AUC of zero maps to, so at least a quarter of those scenes have a zero partial AUC.
Most of what the forgiven group recovers is matched by a head without lane-keeping supervision: the same pass's drivable-area logits, read against the same labels, reach $\mathrm{AUC}_{\mathrm{DAC}\to\mathrm{LK}}=\N{probe_T_dac_auc}$ where the head reaches \N{probe_T_lk_auc} on the forgiven scenes, and \N{probe_U_dac_auc} where it reaches \N{probe_U_lk_auc} over all \N{probe_U_scenes} unforgiven scenes with a mixed published column, covering \N{probe_T_dac_share}\% and \N{probe_U_dac_share}\% of its excess over chance.
The overwrite changed no drivable-area value on this split, so the shares compare the lane-keeping head with a head of the same network whose own targets it never touched, on the same scenes and base rates.
Matching one-to-one on the failing-candidate count pairs \N{probe_pairs} of those scenes with an unforgiven scene and leaves \N{probe_unmatched} unmatched.
For $D$ the within-scene $\mathrm{AUC}_{\mathrm{DAC}\to\mathrm{DAC}}-\mathrm{AUC}_{\mathrm{LK}\to\mathrm{LK}}$, each head scored against its own column, the primary contrast $D_{\text{forgiven}}-D_{\text{matched}}$, whose drivable-area term is the control for how hard a scene is to rank overall, is $+\N{probe_did_match}$ [\N{probe_did_match_lo}, \N{probe_did_match_hi}] over those pairs and $+\N{probe_did_strat}$ [\N{probe_did_strat_lo}, \N{probe_did_strat_hi}] under direct standardization over strata of that count.
As a secondary reading, over the pairs the unforgiven side reaches AUC \N{probe_matched_lk_auc} [\N{probe_matched_lk_auc_lo}, \N{probe_matched_lk_auc_hi}], average precision \N{probe_matched_lk_ap} and McClish partial AUC \N{probe_matched_lk_pauc}, against an AUC of \N{probe_Tmatched_lk_auc} on the \N{probe_pairs} matched LK-forgiven scenes, a difference of medians of $\N{probe_delta_match}$ [$\N{probe_delta_match_lo}$, $\N{probe_delta_match_hi}$] (Appendix~\ref{sec:probeestimators}).
Discriminability falls on forgiven scenes but stays above chance.
Labels differ in source between the groups: published labels on the matched one and our reconstruction (Section~\ref{sec:method}) on the forgiven one.
The gap holds on held-out navtest scenes, whose forgiven and matched groups take their labels from one regeneration: the median AUC is \N{nt_released_R1} [\N{nt_released_R1_lo}, \N{nt_released_R1_hi}] on all \N{nt_T} mixed LK-forgiven scenes against \N{nt_released_R2} [\N{nt_released_R2_lo}, \N{nt_released_R2_hi}] on the \N{nt_pairs} unforgiven scenes matched to them, and the primary contrast is $+\N{nt_released_R3}$ [\N{nt_released_R3_lo}, \N{nt_released_R3_hi}] (Appendix~\ref{app:retraining}).
The Hydra-MDP checkpoint released with GTRS, whose configuration takes the same file (Section~\ref{sec:columnreadings}), shows the gap too: \N{hmdp_nt_R1} against \N{hmdp_nt_R2} on navtest, with primary contrasts of $+\N{hmdp_nt_R3}$ on navtest and $+\N{hmdp_tr_R3}$ on navtrain, both intervals above zero (Appendix~\ref{sec:probeestimators}).

\section{Label regeneration depends on the OpenBLAS kernel selected for the processor}
\label{sec:xarch}
Under the same code, metric cache and scoring configuration that reproduce the released GTRS-Dense labels bitwise on Broadwell across the full corpus except ten residuals in two of the nine columns, regeneration on Intel Sapphire Rapids matches all nine columns jointly on \N{spr_default_reproduced} of \N{spr_default_tokens} tokens and on \N{spr_sample_reproduced} of a separate \N{spr_sample_tokens}-token sample, whose regenerated arrays a second machine matches bitwise.
With the default kernel, the maximum absolute difference on those \N{spr_default_tokens} tokens reaches 1.0 on every one of the nine columns, whose values lie in $[0, 1]$ and seven of which are restricted to $\{0, 1/2, 1\}$, so in every column at least one label flips rather than differing in the last bit; the share of a token's nine-column entries that differ from the label file has median \N{ob_default_differing_median}\% and maximum \N{ob_default_differing_max}\%.
\citet{schlogl2023causes} find that across 75 platforms the deviations in a trained model's softmax outputs rarely change a label on natural inputs; on the dual-core x86 systems available on Google Cloud, the CPUs fall into equivalence classes aligned with clusters of co-occurring SIMD flags, one holding \code{avx512f}.
Here the object is the released label file, and a token matches only when its complete nine-column array does.
The same symptom has been on the benchmark's issue tracker since 2024, attributed there to \code{np.linalg.pinv} and to an OpenBLAS kernel on newer Intel parts, with the maintainer's question of whether two reports share a cause unanswered; they compare scores, not artifacts, and do not separate the CPU from the accelerator, and the readings above are that comparison made on the label file (Appendix~\ref{sec:upstreamreports}).
Three repeated Sapphire Rapids runs of the same 52 tokens agree bitwise with each other, and the nine columns are invariant across two Sapphire Rapids SKUs, matching the CPU determinism \citeauthor{schlogl2023causes} report; the difference is between platforms and not between runs.
The difference enters through the OpenBLAS library bundled with NumPy, which selects its kernel from the processor at run time and selects a different one on Sapphire Rapids than on Broadwell; with that library alone pinned to the kernel it selects on Broadwell, regeneration on Sapphire Rapids matches all nine columns on \N{ob_pin_reproduced} of the \N{ob_pin_tokens} tokens, and on all \N{ob_allpin_nodispatch_tokens} with every OpenBLAS build pinned and NumPy's AVX-512 dispatch disabled (supplementary material).
On the separate sample, AVX-512 alone does not produce the difference: with NumPy's and SciPy's builds on SkylakeX, also an AVX-512 kernel, regeneration on Sapphire Rapids matches all nine columns on all \N{ob_skx_tokens} tokens, where the default Cooperlake kernel matches none.
Feeding Broadwell's simulator output into the Sapphire Rapids scorer recovers the released labels bitwise on all eight tokens tested.
On the \N{ob_pin_tokens} tokens, which include those eight, the trajectory array entering the simulator is bitwise the same under the default and the pinned kernel while the labels differ on every token (supplementary material); together with the cross-feed on the eight, this places the divergence inside the simulator.

\section{Pre-overwrite targets shift lane keeping and extended comfort, more on forgiven scenes}
\label{sec:retraining}
We continue the released checkpoint from step \N{rt_initial_step}, including its Adam state, for \N{rt_steps} optimizer steps on navtrain with \N{rt_seeds} seeds per supervision: A retains the published targets, M omits forgiven token-columns from their component loss, and P uses pre-overwrite targets.
Evaluation uses single-stage non-reactive simulation on \N{rt_tokens} navtest tokens: OFF disables forgiveness, and ON keeps it and rebuilds the aggregate after human-zero overwrites, as the upstream correction does \citep{filterfix}.
The benchmark's extended comfort (EC) scores consistency between adjacent-frame plans \citep[\code{ego_is_two_frame_extended_comfort}]{gtrscode}, over tokens with an adjacent frame.
F holds the \N{rt_n_human_fail} navtest tokens whose human reference scores zero on LK, DDC, TLC, NC or TTC, and R the other \N{rt_n_rest}.
A contrast with A at the same seed is observed (O) when all seeds share its sign, every \N{rt_confidence}\% interval excludes zero, and its absolute mean exceeds the largest gap between seeds of one supervision, read over seeds \N{rt_seed0}--\N{rt_seed1} and \N{rt_seed0}--\N{rt_seed2} (the two- and three-seed readings).

We observe higher OFF lane keeping and lower EC for P than A, with both changes larger on forgiven scenes (F) than on the rest (R): the LK interaction is positive and the EC interaction negative in every seed, with intervals excluding zero (Table~\ref{tab:retraining}).
\begin{table}[t]
\centering\small
\caption{P--A subset interactions in score points, with \N{rt_confidence}\% intervals: the paired difference on F minus that on R.
Both are observed (O) in the two- and three-seed readings (Appendix~\ref{app:retraining}).}
\label{tab:retraining}
\begin{tabular}{@{}lccc@{}}
\toprule
Metric & $s=\N{rt_seed0}$ & $s=\N{rt_seed1}$ & $s=\N{rt_seed2}$ \\
\midrule
LK OFF & \N{rt_IP0_off_LK} [\N{rt_IP0_off_LK_lo}, \N{rt_IP0_off_LK_hi}] & \N{rt_IP1_off_LK} [\N{rt_IP1_off_LK_lo}, \N{rt_IP1_off_LK_hi}] & \N{rt_IP2_off_LK} [\N{rt_IP2_off_LK_lo}, \N{rt_IP2_off_LK_hi}] \\
EC & \N{rt_IP0_off_EC} [\N{rt_IP0_off_EC_lo}, \N{rt_IP0_off_EC_hi}] & \N{rt_IP1_off_EC} [\N{rt_IP1_off_EC_lo}, \N{rt_IP1_off_EC_hi}] & \N{rt_IP2_off_EC} [\N{rt_IP2_off_EC_lo}, \N{rt_IP2_off_EC_hi}] \\
\bottomrule
\end{tabular}
\end{table}
OFF lane-keeping gains on F are \N{rt_P0A0_off_LK_human_fail}/\N{rt_P1A1_off_LK_human_fail}/\N{rt_P2A2_off_LK_human_fail} points, versus \N{rt_P0A0_off_LK_rest}/\N{rt_P1A1_off_LK_rest}/\N{rt_P2A2_off_LK_rest} on R, in seed order.
On F, where the human reference itself fails a check, these gains come with a centimeter-scale shift from the human trajectory, \N{rt_dist_F_ADE_min}--\N{rt_dist_F_ADE_max}\,m on average (ADE); on R the distance does not separate.
The EC decline also appears on the tokens whose lane-keeping outcome does not change, by \N{rt_unchanged_EC_drop_min}--\N{rt_unchanged_EC_drop_max} points with an interval below zero in every seed.
M's all-token OFF lane-keeping contrast is also observed (O), while its subset contrasts do not separate.
For the positive control L0, with every lane-keeping target passing, we observe lower OFF lane keeping than A at their shared seed: this training budget registers a lane-keeping response.
On Section~\ref{sec:probe}'s probe, the in-sample median AUC on mixed LK-forgiven scenes is \N{prt_P_R1_min}--\N{prt_P_R1_max} for P against \N{prt_A_R1_min}--\N{prt_A_R1_max} for A, and the primary contrast \N{prt_P_R3_min}--\N{prt_P_R3_max} against \N{prt_A_R3_min}--\N{prt_A_R3_max}; on held-out navtest that AUC is \N{nt_PA_R1_min}--\N{nt_PA_R1_max} higher for P than for A, and the primary contrast \N{nt_P_R3_min}--\N{nt_P_R3_max} against \N{nt_A_R3_min}--\N{nt_A_R3_max} (Appendix~\ref{app:retraining}).
At \N{rt530_steps} steps, P--A's OFF lane-keeping and EC contrasts keep their signs and verdicts on all tokens and on F, as do both interactions and the probe contrasts of Section~\ref{sec:probe}, and OFF EPDMS still does not separate P from A (Appendix~\ref{app:retraining}).
Lane keeping and comfort separate P from A, by more on forgiven scenes than on the rest, while OFF EPDMS, the aggregate that includes both, does not separate the supervision policies (Appendix~\ref{app:retraining}).

\section{Discussion}
Because EPDMS applies the same rule when it scores, the overwritten targets agree with the benchmark's own evaluation; what the overwrite removes, where a forgiven column was mixed, is the pre-overwrite ordering among a scene's candidates, which is the quantity measured here.
Consistent with this alignment, we observe higher ON EPDMS under published than pre-overwrite targets at 300 steps in the two-seed reading only, and at 530 steps in neither reading (Appendix~\ref{app:retraining}).
Removing forgiveness has its own cost: \citet[Section~5; Table~4, stock filter-off row]{scorebasis} document zero human-replay and PDM-Closed planner scores on interaction-heavy candidates and near-zero corpus means in their stock filter-off control, whose rollouts diverge (Section~\ref{sec:method}).
The all-excluded $B$ class in Appendix~\ref{sec:labelranking} supplies a label-side instance of this cost on navtrain.
The overwrite is not inert for training: with the pre-overwrite targets restored, we observe higher lane keeping on forgiven navtest scenes, while OFF EPDMS does not separate the supervision policies (Section~\ref{sec:retraining}).

\paragraph{Limitations.}
\label{sec:limitations}
The continuation covers GTRS-Dense on navtest, starting from weights trained with forgiveness.
The v2 label census and solver contrasts run on an Intel Xeon E5-2640 v4 (Broadwell); the separate v1 human census spans the five CPU models recorded in Appendix~\ref{app:v1}.
The navtrain mixed-versus-constant classification counts were taken on Broadwell only, not on AMD Rome, where Appendix~\ref{app:flipped}'s lane-keeping recomputation equals the stored pre-overwrite column on all \N{flip_rome_tokens} tokens run there, or on the Sapphire Rapids parts of Section~\ref{sec:xarch}, where the published labels do not reproduce at all under the default kernel. With forgiveness disabled there, none of \N{spr_noforgiveness_tokens} tokens matches the Broadwell reconstruction on all nine columns.
The label census covers navtrain, not navtest, navhard, or synthetic frames.
The exposure result (Appendix~\ref{sec:dropoutexposure}) is a combinatorial expectation over the sampling distribution; no actual training run's dropout indices were read.
The rollout-divergence readings cover the human-reference rollout under the stock solver only.

\section*{AI use statement}
In this work, we used generative AI tools to search for, identify and summarize relevant literature, and to edit and polish the text of this paper for readability.
We have not used generative AI tools to help develop theoretical models or conceptual frameworks, to interpret results, to propose or refine hypotheses, to design or give feedback on research methodology or experiments, or to implement methods.
Generating synthetic data sets, formulating mathematical claims, providing critical ingredients for proving mathematical claims, assisting in the writing of proofs, translation, cleaning and reformatting datasets, and supporting qualitative and thematic data analysis are not applicable to this work.
Additionally, we used generative AI tools for the creation of artifacts.
We have reviewed all AI-assisted work.
The papers we cite were read in full, and a citation was checked at the place in the cited paper that supports the claim it carries; the authors edited the text and verified it against the results it reports.
We take responsibility for the final content of this work, including text, claims or artifacts produced with the aid of generative AI.

\section*{Reproducibility statement}
The supplementary material gives the settings behind its numbers that the paper does not state: the public inputs with their digests, the configurations of the label-regeneration, metric-cache and checkpoint-reading runs, the token samples and processors of Section~\ref{sec:xarch}, the continued-training and evaluation configurations of Section~\ref{sec:retraining}, the order-statistic conventions, the rollout settings of Appendix~\ref{app:flipped}, and the resampling, tie treatments and held-out labels behind the readings of Section~\ref{sec:probe}.
Appendix~\ref{app:cache} reports the full-corpus certification of the published labels and the AMD Rome sample, and Section~\ref{sec:xarch} the Sapphire Rapids differences.

\bibliographystyle{iclr2027_conference}
{\microtypesetup{protrusion=false}
\bibliography{references}
}
\appendix
\section{Reconstructed v1 human scores}
\label{app:v1}
The census scores all 103,288 tokens, and the eight consistency checks listed in the supplementary material hold on every token.
It spans Intel E5-2640 v4 and Gold 5118, and AMD EPYC 7502, 7543, and 8224P.
The paired means, EP 0.8799 and PDMS 0.9495, sit near the published navtest human row's 0.875 and 0.948 \citep[Table~1, human row]{navsim}.
This anchor supports the reconstruction, not the filtering.
The cache was built by v2, whose observation-construction chain differs from v1's.
Its missing forecast second requires a TTC clamp.
The measured TTC is consequently an upper bound on the intended v1 value.
The v2 census also removes agents overlapping the ego at frame zero, whereas v1 uses non-reactive replay without that removal.
On the 12 of 103,288 tokens where the v1 and v2 NC values differ, v1 is the lower; that input difference, rather than changed NC semantics, accounts for this direction.

\paragraph{The v1 driving-direction member (Appendix~\ref{sec:splitcounterexamples}) is inert on this corpus.}
The v1 loop slices the proposal axis rather than summing the intended one-second temporal window.
DDC equals 1.0 on all 103,288 human trajectories, including all 2,003 that v2 scores below 1.0.
In the artificial batch $[\mathrm{ref},\mathrm{ref}]$, the second row's accumulated vector is exactly twice the first row's on all 876 tested tokens, with 513 nonzero cases; the doubling crosses a scoring threshold on 2 of 876.
The human DDC score changes between the single-row and paired batches on zero tokens.
Thus v1's driving-direction member cannot fire on this human corpus, while the mechanism that would make it batch-dependent is measured in the artificial configuration.

\section{Counterexamples to the documented split criterion}
\label{sec:splitcounterexamples}
\paragraph{The documented filter.}
NAVSIM's split-generation paragraph names navtrain and navtest, removes scenes with human PDMS below 0.8, also removes scenes with constant-velocity PDMS above 0.8, and reserves adjustment of these thresholds to obtain the desired dataset size
\citep[Section~3.1, ``Filtering for challenging scenes'']{navsim}.
Its published v1 equation uses $\{\mathrm{NC,DAC,EP,TTC,C}\}$ and omits DDC, whereas the released v1 code includes DDC as a multiplicative member; LK and TLC are absent from both v1 aggregates.
A cited description of v1 records DDC instead as a weighted term of weight zero \citep[Appendix~E.3]{flowr2a}; the two readings agree on this corpus because DDC is 1.0 on every human trajectory in it (Appendix~\ref{app:v1}).
The distributed \code{SceneFilter} has twelve fields, none expressing a score, metric, or threshold; its token-list consumption path checks set membership \citep[\code{SceneFilter}; \code{filter_scenes}]{navsimcode}.
The documented score screening is an offline step embodied in the supplied list, outside the distributed split-consumption code, which cannot reapply it.
No split-list generation code was found in the \code{autonomousvision/navsim} repository at \code{0a380a9} (supplementary material).
New fields falling outside an older validation predicate are a familiar data-validation problem \citep{breck2019validation}.
The relationship between NAVSIM's original split filter and its later metric channels is an instance of this problem.

\paragraph{Counterexamples.}
Scoring all 103,288 navtrain human trajectories with the reconstructed v1 pipeline yields 34 scenes (0.0329\%) below 0.8 even under the most permissive of the four aggregations, NAVSIM's published formula scored on the human trajectory alone.
These counterexamples hold under both the published and the code aggregation and any choice of reference row.
With the cached PDM-Closed trajectory as the reference row, the code aggregation gives 364 (0.3524\%).
Both counts are inconsistent with the described removal of every scene below 0.8; neither establishes what threshold, if any, was executed.
The additional 330 are EP-only cases; their reference trajectory is the cached v2 PDM-Closed rollout, and the median factor needed to raise normalized EP to the threshold is 1.102.
A 10\% reduction of the denominator can move that median case across the threshold; the 34 reference-independent cases carry the conclusion without this dependence.
Across the paired counterexamples, the nonexclusive failing-member counts are TTC zero 33, NC zero 20, NC half-valued 3, DAC zero 0, and DDC below one 0.
The counts are lower bounds under three one-sided relaxations: missing TTC anchors in the v2 cache, truncation from v1's anchor count to v2's, and omission of the DDC factor in the published-formula one-row reading.
The nearest measured score is $3.21\times10^{-6}$ from 0.8, with none within $10^{-6}$.
NC's at-fault classifier is AST-identical between versions, and its score function differs only cosmetically.
The changed definitions are DDC's aggregation axis and TTC's anchor count (41 versus 32).
The \code{COMFORTABLE} member of NAVSIM v1.0 (\code{9f3ea68}) and v2's TLC, LK, HC, and \code{two_frame_extended_comfort} are not mapped between versions.
Appendix~\ref{app:v1} records the reconstruction controls and input boundaries.

\section{Full-corpus certification and the cache behind eight of its residuals}
\label{app:cache}
\paragraph{Full-corpus certification.}
With forgiveness enabled, the full-corpus certification reproduces all nine published columns byte for byte on \N{reproduced} of \N{population_tokens} tokens.
The \N{residual_total} mismatches split into \N{ties} storage-precision ties and \N{cache_residuals} cache-content residuals.
This appendix localizes the cache dependency behind the latter.
A storage-precision tie is a token whose \code{pdm_score} column is bitwise equal and whose every differing entry differs by one float16 unit in the last place at its value.
That certification ran on Broadwell, and two further microarchitectures were tested on a separate 260-token sample: AMD Rome reproduces all nine columns on \N{rome_tokens}/\N{rome_expected} tokens, and Sapphire Rapids is the subject of Section~\ref{sec:xarch}.
Pinning every run of a comparison to one processor keeps the processor out of the differences between those runs, which is a separate matter from whether the labels reproduce on it.

The eight cache-content residuals trace to accumulated red-light identifiers consumed during leading-agent selection for the Intelligent Driver Model (IDM) in \code{autonomousvision/navsim@0a380a9}, \path{navsim/planning/simulation/planner/pdm_planner/proposal/pdm_generator.py}:263.
Clearing the accumulated field or short-circuiting that red-light use recovers seven of the eight; the remaining residual, \code{9430eba18cb05d7c}, is reproduced by injecting its measured identifier set, and a log-grouped construction, one chunk per log, reproduces all eight.
The log-grouped construction identifies a configuration because it changes both chunk size and composition; the field and consumption-site interventions localize the mechanism.

\section{Label matrices and candidate-set overlap}
\label{sec:labelranking}
Before applying any selection statistic, the two label matrices are bitwise identical on all nine columns for 90,703 of 103,278 tokens after excluding the ten reproduction residuals.
The affected pool here is exactly the complement, 12,575 tokens: the $P_1$ pool less its one reproduction residual.
Over the full 103,288-token comparison, differing-column counts are LK 9,982, \code{pdm_score} 2,975, TLC 2,156, DDC 876, TTC 18, NC 10, DAC 0, and HC 0; EP differs on the ten reproduction residuals.
The two accounts of \code{pdm_score} add rather than overlap: the 2,967 forgiven columns and the eight residuals whose \code{pdm_score} differs are disjoint sets, and their union is exactly the 2,975 columns on which the two matrices differ.
The column reproduces bitwise on \N{pdm_reproduced} of \N{population_tokens} tokens, failing on \N{cache_residuals} of the ten and on neither storage-precision tie.

\paragraph{Selection rule and set overlap.}
The source consumes labels as seven per-column BCE targets and contains no argmax over the label matrix.
Its vocabulary-wide argmax acts on student predictions at inference.
We substitute the loss-mapped labels $t_j$ into that rule's metric part, a diagnostic construction rather than an operation performed by the training code:
\begin{equation}
\begin{split}
S(i)={}&0.1\ln t_{\mathrm{TLC},i}+0.1\ln t_{\mathrm{NC},i}
       +0.9\ln t_{\mathrm{DAC},i}+0.2\ln t_{\mathrm{DDC},i}\\
       &+6\ln(7t_{\mathrm{TTC},i}+7t_{\mathrm{EP},i}+3t_{\mathrm{LK},i}).
\end{split}
\label{eq:labelrule}
\end{equation}
We promote stored float16 labels to float64, map NC/DDC as the loss does, retain $\ln0=-\infty$, and compare exact maximizer sets $M_A,M_B$ for published ($A$) and pre-overwrite ($B$) files without breaking ties.
Evaluated on the label matrices with the metric part of the rule the released agent applies at inference (\path{navsim/agents/gtrs_dense/hydra_model.py}:264--271), its imitation term absent because no label column corresponds to it, the published and the pre-overwrite label files place different candidate sets first on 10,193 of 103,278 navtrain tokens (9.87\%), and share no top-ranked candidate on 967 (0.94\%).
The set-different total is 81.06\% of the affected pool; the decomposition follows with both denominators:
\begin{center}
\small
\begin{tabular}{@{}lrrr@{}}
\toprule
Relationship & Tokens & \% of 12,575 & \% of 103,278\\
\midrule
$M_B\subsetneq M_A$ & 7,637 & 60.73 & 7.39\\
$M_A\subsetneq M_B$ & 1,589 & 12.64 & 1.54\\
Crossing & 0 & 0.00 & 0.00\\
Disjoint & 967 & 7.69 & 0.94\\
\bottomrule
\end{tabular}
\end{center}
The first class leaves a smaller jointly optimal set under $B$, refining the published tie without implying a unique candidate.
In the second, $B$ excludes all 16,384 candidates: its formal maximizer set is the whole vocabulary at $-\infty$, and it ranks nothing.
That class is 1,589 of the 10,193, so the set-different total excluding it is 8,604 tokens, 68.42\% of the affected pool.

\paragraph{Ties and the separate unique-candidate row.}
The analysis is tie-dominated: both sets are singletons on 542 of 12,575 affected tokens (4.31\%), both are nonsingletons on 86.85\%, and their median sizes are 660 ($A$) and 130 ($B$).
Tokens on which both sets are singletons and differ:
\begin{center}
\small
\begin{tabular}{@{}lrrrr@{}}
\toprule
Case & Tokens & \% of 542 & \% of 12,575 & \% of 103,278\\
\midrule
Both unique, different & 167 & 30.81 & 1.33 & 0.1617\\
\bottomrule
\end{tabular}
\end{center}
For these tokens, scoring both candidates with $A$'s aggregate gives median $\Delta^{(A)}=+11.91$ points, positive on 164/167; scoring both with $B$'s gives median $\Delta^{(B)}=+11.91$, positive on 165/167.
The two medians are equal before rounding.
Each difference is $B$-ranked minus $A$-ranked, on the 0--100 label scale, per scene and candidate.

\paragraph{Other selection statistics.}
Under the loss's own weights $w_j$ (Section~\ref{sec:columnreadings}), $S_{\mathrm{lossw}}=\sum_jw_jt_j$, the split by whether the two files' maximizer sets differ agrees with that of $S$ on 10,971 of the 12,575 affected tokens (87.24\%), and under $S_{\mathrm{pdm}}=\code{pdm_score}$ on 3,180 (25.29\%); the set-overlap counts above are those of $S$.

\section{The erased distinction would usually survive vocabulary dropout}
\label{sec:dropoutexposure}
\paragraph{Expected exposure under vocabulary dropout.}
The released training configuration uses \mbox{\code{randperm}} and \code{num_total // 2} to retain $s=8{,}192$ of $N=16{,}384$ candidates without replacement; the labels are sliced by the same indices.
For a mixed binary column with $k$ candidates in one class, the standard hypergeometric calculation gives
\begin{equation}
p(k)=1-\frac{\binom{k}{s}+\binom{N-k}{s}}{\binom{N}{s}},\qquad \binom{a}{s}=0\ \text{for }a<s.
\label{eq:exposure}
\end{equation}
This is the probability that a sampled sub-vocabulary would contain both pre-overwrite target values.
We pool it over mixed forgiven loss columns, weighting each token-column equally; a column already constant before overwrite has no distinction for dropout to remove.
The constant published column remains constant under the same sampling.
Across those \N{exposure_population} mixed forgiven token-columns in the loss, the mean probability in Equation~\ref{eq:exposure} is \N{exposure_mean_probability}.
For a mixed binary column, retaining half the candidates gives $p(k)\ge s/N=0.5$, with equality at a singleton minority; probabilities below 0.5 are structurally excluded.
For \N{exposure_p_one} of these \N{exposure_population} columns, the probability equals 1.0 to double precision.
Averaged over these columns, the erased distinction would therefore be present on essentially every sampled appearance of the corresponding scene during training.
The calculation closes one alternative: if dropout usually removed the distinction anyway, the overwrite would be erasing something training rarely sampled.

\section{What the overwrite turns from failing to passing in lane keeping}
\label{app:flipped}
We roll out all \N{vocab_dense} candidates of the \N{LK_forgiven} forgiven LK columns and of the \N{probe_U_scenes} unforgiven mixed ones with the simulator that \code{pdm_score_full_v2} calls, and the human reference exactly as the forgiveness trigger does, reading positions at the bounding-box center that the lane-keeping check uses.
The check recomputed from these rollouts equals the stored pre-overwrite column on all \N{flip_rollout_candidates} candidates; it fails the human on every forgiven column and passes it on every unforgiven one.
Table~\ref{tab:flipped} compares the \N{LK_mixed_changed_entries} flipped entries of the \N{LK_mixed} mixed forgiven columns with the failing entries of the unforgiven ones.
On the flipped entries, the largest distance from the route centerline outside intersections has median \N{flip_T_devC_median_m}\,m, and \N{flip_T_devC_gt_human_percent}\% of them reach further from it than the human rollout of the same scene, whose own largest distance has median \N{flip_T_human_devC_median_m}\,m over these columns.
Their longest run above the \N{flip_lk_limit_m}\,m limit spans a median of \N{flip_T_runC_median_steps} of the rollout's \N{flip_rollout_steps} steps; the check fails a candidate at a run of \N{flip_lk_steps_required}.
Rerun with the human rollout's path as its reference line, the same check fails \N{flip_T_lkH_fail_percent}\% of the flipped entries, and \N{flip_T_lkH_dac_fail_percent}\% of all flipped entries fail it together with drivable-area compliance; \N{flip_T_within_half_m_percent}\% stay within \N{flip_lk_limit_m}\,m of the human rollout laterally at every step.
Of the flipped entries, \N{flip_T_dac_fail_percent}\% also fail drivable-area compliance, against \N{flip_U_dac_fail_percent}\% of the failing entries in the unforgiven columns.
The human rollout's own largest distance from the route centerline is at most \N{flip_human_bin_a_m}\,m on \N{flip_T_human_bin_a_percent}\% of the mixed forgiven columns, between \N{flip_human_bin_a_m} and \N{flip_human_bin_b_m}\,m on \N{flip_T_human_bin_b_percent}\%, between \N{flip_human_bin_b_m} and \N{flip_human_bin_c_m}\,m on \N{flip_T_human_bin_c_percent}\%, and above \N{flip_human_bin_c_m}\,m on \N{flip_T_human_bin_d_percent}\%.

\begin{table}[ht]
\centering\small
\caption{Flipped entries of the mixed forgiven LK columns against the failing entries of the mixed unforgiven ones. ``Against the human path'' reruns the lane-keeping check with the human rollout's path as the reference line; ``within \N{flip_lk_limit_m}\,m'' bounds the offset from the human rollout, perpendicular to its heading, at every \N{flip_step_s}\,s step; the run is the check's longest run above \N{flip_lk_limit_m}\,m outside intersections, of \N{flip_rollout_steps} steps. ``Another check'' is any of DAC, DDC, NC, TTC and TLC, read before the overwrite and after $b_j$. Percentages are over entries.}
\label{tab:flipped}
\begin{tabular}{@{}lrr@{}}
\toprule
& Forgiven, flipped & Unforgiven, failing \\
\midrule
Columns & \N{LK_mixed} & \N{probe_U_scenes} \\
Entries & \N{LK_mixed_changed_entries} & \N{flip_U_fail_entries} \\
Fails LK against the human path (\%) & \N{flip_T_lkH_fail_percent} & \N{flip_U_lkH_fail_percent} \\
Within \N{flip_lk_limit_m}\,m of the human rollout (\%) & \N{flip_T_within_half_m_percent} & \N{flip_U_within_half_m_percent} \\
Fails DAC (\%) & \N{flip_T_dac_fail_percent} & \N{flip_U_dac_fail_percent} \\
Fails another check (\%) & \N{flip_T_other_fail_percent} & \N{flip_U_other_fail_percent} \\
Largest distance from centerline, median (m) & \N{flip_T_devC_median_m} & \N{flip_U_devC_median_m} \\
Longest run above \N{flip_lk_limit_m}\,m, median (steps) & \N{flip_T_runC_median_steps} & \N{flip_U_runC_median_steps} \\
\bottomrule
\end{tabular}
\end{table}

\section{Reports of the same symptom on the upstream tracker}
\label{sec:upstreamreports}
Users of the benchmark have reported machine-dependent scoring since 2024; the threads below were read on 21 September 2026.
In \code{navsim} issue~\#28 (2024-10), the reporter writes ``I changed a machine to run the evaluation, then the PDMS is 0.96, and every metric item is normal'', and localizes it to ``the bug comes from \code{np.linalg.pinv()}, which generate different output given the same input''.
A second reporter there resolved it ``by changing to another computer''.
A third attributes it to ``some micro kernel selection issues for openblas \dots\ when running codes on CPUs with newer arch like intel 8457C'', and prescribes a NumPy upgrade together with ``purge all metric\_caches and rebuild them''.
That part number is a Sapphire Rapids SKU, the same generation as the Sapphire Rapids parts used here.
Issue~\#141 (2025-08) reports a zero PDM score on one accelerator and reasonable scores on another; a second reporter there is on a ``Gen4 Xeon Scalable CPU'', also Sapphire Rapids, and a third traces the zero to the drivable-area column and reports that fixing seeds, enabling deterministic algorithms and disabling TF32 ``did not change anything''.
The maintainer asked in 2025-09 whether \#141 ``related to \#28''; the thread records no answer.
Issue~\#136 concerns instability in the same \code{pinv} call, and issue~\#81 concerns a metric mismatch attributed to library versions.
These threads compare scores rather than stored artifacts, report no bitwise comparison and no reference artifact to compare against, and do not separate the CPU from the accelerator.
The prescription in \#28 to rebuild every metric cache after a library change already assumes that the cached artifacts depend on the machine; what is absent is a measurement of how far.

\section{Estimators and intervals of Section~\ref{sec:probe}}
\label{sec:probeestimators}
Section~\ref{sec:probe} reports the difference of median AUCs under one-to-one matching, $\N{probe_delta_match}$ [$\N{probe_delta_match_lo}$, $\N{probe_delta_match_hi}$]; direct standardization over strata of the failing-candidate count gives $\N{probe_delta_strat}$ [$\N{probe_delta_strat_lo}$, $\N{probe_delta_strat_hi}$].

\paragraph{How the interval and the matched contrast are formed.}
Each scene contributes one AUC per column; a trapezoid-rule computation of the
same quantity agrees to within \N{probe_rank_trapezoid_max}. Ranks are
averaged over ties, which scores a tie one half, and the AUC is the rank sum
$(\sum_{\text{positive ranks}} - n_1(n_1+1)/2)/(n_1 n_0)$.
AP and the partial AUC take the candidates in \code{torch.argsort} order with ties not grouped.
On a rerun of the forward pass, grouping tied candidates, or breaking ties at random in each of \N{ties_draws} draws, leaves the median and the fifth and twenty-fifth percentiles of both, and the share of zero partial AUCs, unchanged at five decimals for the released model and each continued one, and every treatment finds the same scenes with a zero partial AUC; for the released model they are the \N{ties_zero_released} scenes behind Section~\ref{sec:probe}'s ``at least a quarter''.
Matching is one-to-one on the failing-candidate count, without replacement,
inside a caliper of 250 candidates. Treatment scenes take their partner in
order of increasing supply, so the scenes with fewest admissible partners
choose first; a treatment scene with no free
partner inside the caliper is returned unmatched and counted. Balance after matching is
read as the absolute standardized difference against a threshold of 0.1 and
comes out at \N{probe_asd_after}.
The drivable-area column is mixed on all \N{LK_mixed} forgiven scenes and on every matched partner, so all \N{probe_pairs} pairs enter $D$.
Direct standardization keeps every scene on both sides instead: strata are
500 candidates wide on the same count, each stratum is weighted by its number
of treatment scenes, and every stratum holding treatment scenes also holds control scenes.
The two estimators therefore read different forgiven scenes: the \N{probe_pairs} matched ones have a median lane-keeping AUC of \N{probe_Tmatched_lk_auc}, and with the \N{probe_unmatched} that matching leaves out the median over all \N{LK_mixed} falls to \N{probe_T_lk_auc}.
The contrast is a difference of medians, the median over the treatment group
minus the median over the control group, not the median of per-pair differences.
Its interval is a percentile bootstrap with $10{,}000$ resamples at the
$2.5$ and $97.5$ percentiles, drawn over pair indices with the same indices
used on both groups so that a scene stays with its match; the matching itself is
formed once and is not redone inside the resampling. The seed is given in the supplementary material.

\paragraph{The Hydra-MDP checkpoint.}
Table~\ref{tab:hmdp} reads the Hydra-MDP checkpoint released with GTRS with the scenes and pairs of Section~\ref{sec:probe} and the analyzer of Table~\ref{tab:rt-probe} on navtrain and those of the held-out probe of Appendix~\ref{app:retraining} on navtest, beside the released GTRS-Dense checkpoint: R1 is the lane-keeping head's median AUC on the forgiven scenes, R2 the same on the matched unforgiven scenes and R3 the primary contrast $D_{\text{forgiven}}-D_{\text{matched}}$.

\begin{table}[htbp]
\centering\small
\caption{Section~\ref{sec:probe}'s readings for the Hydra-MDP checkpoint released with GTRS and for the released GTRS-Dense checkpoint, with 95\% intervals.}
\label{tab:hmdp}
\begin{tabular}{@{}llcc@{}}
\toprule
Probe & Reading & Hydra-MDP & GTRS-Dense \\
\midrule
navtrain & R1 & \N{hmdp_tr_R1} [\N{hmdp_tr_R1_lo}, \N{hmdp_tr_R1_hi}] & \N{probe_T_lk_auc} [\N{probe_T_lk_auc_lo}, \N{probe_T_lk_auc_hi}] \\
 & R2 & \N{hmdp_tr_R2} [\N{hmdp_tr_R2_lo}, \N{hmdp_tr_R2_hi}] & \N{probe_matched_lk_auc} [\N{probe_matched_lk_auc_lo}, \N{probe_matched_lk_auc_hi}] \\
 & R3 & \N{hmdp_tr_R3} [\N{hmdp_tr_R3_lo}, \N{hmdp_tr_R3_hi}] & \N{probe_did_match} [\N{probe_did_match_lo}, \N{probe_did_match_hi}] \\
navtest & R1 & \N{hmdp_nt_R1} [\N{hmdp_nt_R1_lo}, \N{hmdp_nt_R1_hi}] & \N{nt_released_R1} [\N{nt_released_R1_lo}, \N{nt_released_R1_hi}] \\
 & R2 & \N{hmdp_nt_R2} [\N{hmdp_nt_R2_lo}, \N{hmdp_nt_R2_hi}] & \N{nt_released_R2} [\N{nt_released_R2_lo}, \N{nt_released_R2_hi}] \\
 & R3 & \N{hmdp_nt_R3} [\N{hmdp_nt_R3_lo}, \N{hmdp_nt_R3_hi}] & \N{nt_released_R3} [\N{nt_released_R3_lo}, \N{nt_released_R3_hi}] \\
\bottomrule
\end{tabular}
\end{table}

\section{Sources of the two devices in Section~\ref{sec:probe}}
\label{sec:probeprecedents}
Section~\ref{sec:probe} uses two established devices.
\paragraph{Reading a main result against a control channel.}
\citet[Section~2]{hewitt2019control} pair a linguistic task with a control task that assigns each word type a random output, and read the gap between the two accuracies \citep[Section~1, Figure~2]{hewitt2019control}, which puts task accuracy in context with ``the probe's capacity to memorize from word types'' \citep[abstract]{hewitt2019control}.
The control there is a random relabeling and the quantity it bounds is a fitted probe's memorization.
The control channel in Section~\ref{sec:probe} is a second output head of the same forward pass, scored against the same labels ($\mathrm{AUC}_{\mathrm{DAC}\to\mathrm{LK}}$), and what it measures is the order a column retains without the head whose supervision was overwritten; in $D$ each head is instead scored against its own column ($\mathrm{AUC}_{\mathrm{DAC}\to\mathrm{DAC}}$ and $\mathrm{AUC}_{\mathrm{LK}\to\mathrm{LK}}$).
\paragraph{Withholding an input to see what the score still returns.}
Three audits of driving benchmarks remove an input and report what the score retains.
\citet[Section~4.1]{scorebasis} run two actor-blind probes on navtest and record EPDMS scores of 79.6 and 79.2 against human replay at 74.0.
\citet[Section~1]{egostatus} reproduce a planner that ``relies solely on the ego status'' and find it on par with UniAD and VAD on L2 distance and collision rate \citep[Section~4, ``Ego status plays a key role'']{egostatus}.
\citet[Section~1]{drivingonmemory} replace a planner's camera input with memories retrieved from earlier traversals, leaving it able to ``only access static and quasi-static information'', and report that this is nearly sufficient on NAVSIM.
Each of the three bounds what a benchmark score requires as input.
The reading in Section~\ref{sec:probe} holds the input fixed and varies the head instead: what it compares is how much of one column's surviving order another head of the same pass already carries.

\section{Order statistics and the seven-column share}
\label{app:conventions}

\paragraph{Order statistics and the median.}
For each loss column, the number of candidates failing it before the overwrite is
counted once per forgiven token-column. The $m$ counts are sorted ascending and
read at five fixed ranks, indices $0$, $\lfloor m/4 \rfloor$, $\lfloor m/2
\rfloor$, $\lfloor 3m/4 \rfloor$ and $m-1$. The median column of
Table~\ref{tab:columns} is the value at rank $\lfloor m/2 \rfloor$, for even $m$ the upper of the two central values.
The pool behind each row is every
forgiven token-column of that column, the all-zero and the mixed together, which
is why a row can reach the vocabulary width exactly.

\paragraph{The seven-column share, column by column.}
The numerator counts target entries over the seven loss columns whose written value the overwrite changed; the denominator is $103{,}288 \times 16{,}384 \times 7 = \N{target_entries}$. In a forgiven column the overwrite changes every entry that was not already passing: all $16{,}384$ in a column constant at zero after $b_j$, and one per failing candidate in a mixed column. DAC and EP have no forgiven column (Table~\ref{tab:columns}). Per column:

\begin{center}
\begin{tabular}{@{}lrl@{}}
\toprule
Column & Changed entries & Counts at the five ranks \\
\midrule
LK & \N{LK_changed_entries} & \N{ranks_LK} \\
TLC & \N{exact_TLC} & \N{ranks_TLC} \\
DDC & \N{exact_DDC} & \N{ranks_DDC} \\
TTC & \N{exact_TTC} & \N{ranks_TTC} \\
NC & \N{exact_NC} & \N{ranks_NC} \\
\bottomrule
\end{tabular}
\end{center}

The five add to \N{supervision_entries_exact}, \N{supervision_share_exact_percent}\% of the denominator; the share is the full quotient rounded to the nearest at four decimals.

\paragraph{DDC before and after target mapping.}
The 876 forgiven DDC columns partition into 490 three-valued, 373 two-valued and 13 one-valued; the raw and loss-effective mixed-column counts in Section~\ref{sec:columnreadings} are cells of that partition, $490+373$ and $490$, and the 386 columns in which no candidate scored 1.0 are the 373 and the 13 together.
Under the mapping the 490 retain both binary labels and the 386 become all zero; the raw uniform columns are the 13.

\section{Continued-training comparison}
\label{app:retraining}
\paragraph{Supervision and scoring.}
L0, the positive control, continues at seed \N{rt_seed0} on the published targets with every lane-keeping target set to passing.
M averages each component's BCE over retained candidate entries of its unmasked tokens, with unchanged component weights and zero contribution if none remain in a batch.
The imitation term and vocabulary dropout are unchanged; within each seed the arms share sample order and dropout random-number schedules.
Checkpoints are selected by optimizer step independently of evaluation scores.
Each model's stored trajectories are scored twice: OFF disables forgiveness; ON rebuilds the multiplicative and weighted aggregate inputs after human-zero overwrites, following the upstream correction cited in Section~\ref{sec:retraining}.
On the released model, GTRS's own scoring with forgiveness on returns the OFF aggregate with the ON components, so ON rebuilds the aggregate as the upstream correction does.
EC is identical under both switches and uses \N{rt_ec_n_all} adjacent-frame tokens: \N{rt_ec_n_human_fail} in F and \N{rt_ec_n_rest} in R (Section~\ref{sec:retraining}).

\paragraph{Pairing, intervals and decision rule.}
We pair by token and seed, bootstrap the \N{rt_logs} logs \N{rt_bootstrap} times with replacement and shared draws for both subsets and all contrasts, and take percentile \N{rt_confidence}\% intervals of token-weighted means, using only finite EC values.
For each metric, switch and population, write $\Delta_s^X=\bar q_{X_s}-\bar q_{A_s}$ and $I_s^X=\Delta_s^X[F]-\Delta_s^X[R]$, with $X\in\{P,M\}$.
A contrast is observed (O) if every seed has the same nonzero sign, every interval excludes zero, and the absolute seed-mean difference exceeds the within-arm seed scale $V_{\mathcal S}(q)=\max_{X\in\{A,M,P\},\,i,j\in\mathcal S}|\bar q_{X_i}-\bar q_{X_j}|$.
For interactions, apply the same rule to $I_s^X$, replacing $\bar q_{X_s}$ in the scale by $D_s^X=\bar q_{X_s}[F]-\bar q_{X_s}[R]$.
When the full rule fails, the contrast is unresolved: S (on the seed scale) if every interval excludes zero, U otherwise.
Verdicts give $\mathcal S=\{\N{rt_seed0},\N{rt_seed1}\}$ followed by $\mathcal S=\{\N{rt_seed0},\N{rt_seed1},\N{rt_seed2}\}$; the primary contrast is all-token OFF EPDMS for P--A.
Point estimates round to nearest and interval endpoints outward at three decimals; scores are percentages and differences are score points, with no simultaneous-coverage adjustment.

\begin{table}[htbp]
\centering\scriptsize
\setlength{\tabcolsep}{3pt}
\caption{Absolute scores on all evaluated tokens, with EC restricted to adjacent-frame tokens; final rows give the positive-control and continuation differences [intervals].}
\label{tab:rt-absolute}
\begin{tabular}{@{}lrrrrr@{}}
\toprule
Arm & EPDMS OFF & EPDMS ON & LK OFF & LK ON & EC \\
\midrule
Released & \N{rt_abs_released_off_EPDMS} & \N{rt_abs_released_on_EPDMS} & \N{rt_abs_released_off_LK} & \N{rt_abs_released_on_LK} & \N{rt_abs_released_off_EC} \\
A$_{\N{rt_seed0}}$ & \N{rt_abs_A0_off_EPDMS} & \N{rt_abs_A0_on_EPDMS} & \N{rt_abs_A0_off_LK} & \N{rt_abs_A0_on_LK} & \N{rt_abs_A0_off_EC} \\
M$_{\N{rt_seed0}}$ & \N{rt_abs_M0_off_EPDMS} & \N{rt_abs_M0_on_EPDMS} & \N{rt_abs_M0_off_LK} & \N{rt_abs_M0_on_LK} & \N{rt_abs_M0_off_EC} \\
P$_{\N{rt_seed0}}$ & \N{rt_abs_P0_off_EPDMS} & \N{rt_abs_P0_on_EPDMS} & \N{rt_abs_P0_off_LK} & \N{rt_abs_P0_on_LK} & \N{rt_abs_P0_off_EC} \\
A$_{\N{rt_seed1}}$ & \N{rt_abs_A1_off_EPDMS} & \N{rt_abs_A1_on_EPDMS} & \N{rt_abs_A1_off_LK} & \N{rt_abs_A1_on_LK} & \N{rt_abs_A1_off_EC} \\
M$_{\N{rt_seed1}}$ & \N{rt_abs_M1_off_EPDMS} & \N{rt_abs_M1_on_EPDMS} & \N{rt_abs_M1_off_LK} & \N{rt_abs_M1_on_LK} & \N{rt_abs_M1_off_EC} \\
P$_{\N{rt_seed1}}$ & \N{rt_abs_P1_off_EPDMS} & \N{rt_abs_P1_on_EPDMS} & \N{rt_abs_P1_off_LK} & \N{rt_abs_P1_on_LK} & \N{rt_abs_P1_off_EC} \\
A$_{\N{rt_seed2}}$ & \N{rt_abs_A2_off_EPDMS} & \N{rt_abs_A2_on_EPDMS} & \N{rt_abs_A2_off_LK} & \N{rt_abs_A2_on_LK} & \N{rt_abs_A2_off_EC} \\
M$_{\N{rt_seed2}}$ & \N{rt_abs_M2_off_EPDMS} & \N{rt_abs_M2_on_EPDMS} & \N{rt_abs_M2_off_LK} & \N{rt_abs_M2_on_LK} & \N{rt_abs_M2_off_EC} \\
P$_{\N{rt_seed2}}$ & \N{rt_abs_P2_off_EPDMS} & \N{rt_abs_P2_on_EPDMS} & \N{rt_abs_P2_off_LK} & \N{rt_abs_P2_on_LK} & \N{rt_abs_P2_off_EC} \\
L$_{\N{rt_seed0}}$ & \N{rt_abs_L0_off_EPDMS} & \N{rt_abs_L0_on_EPDMS} & \N{rt_abs_L0_off_LK} & \N{rt_abs_L0_on_LK} & \N{rt_abs_L0_off_EC} \\
\midrule
L0--A0 & \shortstack{\N{rt_L0A0_off_EPDMS_all}\\{}[\N{rt_L0A0_off_EPDMS_all_lo}, \N{rt_L0A0_off_EPDMS_all_hi}]} & \shortstack{\N{rt_L0A0_on_EPDMS_all}\\{}[\N{rt_L0A0_on_EPDMS_all_lo}, \N{rt_L0A0_on_EPDMS_all_hi}]} & \shortstack{\N{rt_L0A0_off_LK_all}\\{}[\N{rt_L0A0_off_LK_all_lo}, \N{rt_L0A0_off_LK_all_hi}]} & \shortstack{\N{rt_L0A0_on_LK_all}\\{}[\N{rt_L0A0_on_LK_all_lo}, \N{rt_L0A0_on_LK_all_hi}]} & \shortstack{\N{rt_L0A0_off_EC_all}\\{}[\N{rt_L0A0_off_EC_all_lo}, \N{rt_L0A0_off_EC_all_hi}]} \\
A0--Released & \shortstack{\N{rt_A0released_off_EPDMS_all}\\{}[\N{rt_A0released_off_EPDMS_all_lo}, \N{rt_A0released_off_EPDMS_all_hi}]} & \shortstack{\N{rt_A0released_on_EPDMS_all}\\{}[\N{rt_A0released_on_EPDMS_all_lo}, \N{rt_A0released_on_EPDMS_all_hi}]} & \shortstack{\N{rt_A0released_off_LK_all}\\{}[\N{rt_A0released_off_LK_all_lo}, \N{rt_A0released_off_LK_all_hi}]} & \shortstack{\N{rt_A0released_on_LK_all}\\{}[\N{rt_A0released_on_LK_all_lo}, \N{rt_A0released_on_LK_all_hi}]} & \shortstack{\N{rt_A0released_off_EC_all}\\{}[\N{rt_A0released_off_EC_all_lo}, \N{rt_A0released_off_EC_all_hi}]} \\
\bottomrule
\end{tabular}
\end{table}

{\small\noindent Interactions: Table~\ref{tab:retraining} gives O/O; all other $I^P$ and all $I^M$ of the scores in Table~\ref{tab:rt-contrasts} are U/U.}

\noindent Table~\ref{tab:rt-contrasts} lists the EPDMS, LK and EC contrasts; DDC, TLC, NC, DAC, EP, TTC and HC do not separate either P--A or M--A under any scoring switch, subset or seed reading.

\begin{table}[htbp]
\centering\scriptsize
\setlength{\tabcolsep}{3pt}
\caption{Paired contrasts: point estimates [intervals] and two-/three-seed verdicts. Populations: all tokens, F and R as in Section~\ref{sec:retraining}; EC uses adjacent-frame tokens and is shared by both switches. R's per-token scores are the same under both switches.}
\label{tab:rt-contrasts}
\begin{tabular}{@{}llllccc c@{}}
\toprule
Contrast & Metric & Mode & Set & $s=\N{rt_seed0}$ & $s=\N{rt_seed1}$ & $s=\N{rt_seed2}$ & Verdict \\
\midrule
P--A & EPDMS & OFF & All & \N{rt_P0A0_off_EPDMS_all} [\N{rt_P0A0_off_EPDMS_all_lo}, \N{rt_P0A0_off_EPDMS_all_hi}] & \N{rt_P1A1_off_EPDMS_all} [\N{rt_P1A1_off_EPDMS_all_lo}, \N{rt_P1A1_off_EPDMS_all_hi}] & \N{rt_P2A2_off_EPDMS_all} [\N{rt_P2A2_off_EPDMS_all_lo}, \N{rt_P2A2_off_EPDMS_all_hi}] & U/U \\
P--A & EPDMS & OFF & F & \N{rt_P0A0_off_EPDMS_human_fail} [\N{rt_P0A0_off_EPDMS_human_fail_lo}, \N{rt_P0A0_off_EPDMS_human_fail_hi}] & \N{rt_P1A1_off_EPDMS_human_fail} [\N{rt_P1A1_off_EPDMS_human_fail_lo}, \N{rt_P1A1_off_EPDMS_human_fail_hi}] & \N{rt_P2A2_off_EPDMS_human_fail} [\N{rt_P2A2_off_EPDMS_human_fail_lo}, \N{rt_P2A2_off_EPDMS_human_fail_hi}] & U/U \\
P--A & EPDMS & OFF & R & \N{rt_P0A0_off_EPDMS_rest} [\N{rt_P0A0_off_EPDMS_rest_lo}, \N{rt_P0A0_off_EPDMS_rest_hi}] & \N{rt_P1A1_off_EPDMS_rest} [\N{rt_P1A1_off_EPDMS_rest_lo}, \N{rt_P1A1_off_EPDMS_rest_hi}] & \N{rt_P2A2_off_EPDMS_rest} [\N{rt_P2A2_off_EPDMS_rest_lo}, \N{rt_P2A2_off_EPDMS_rest_hi}] & U/U \\
P--A & EPDMS & ON & All & \N{rt_P0A0_on_EPDMS_all} [\N{rt_P0A0_on_EPDMS_all_lo}, \N{rt_P0A0_on_EPDMS_all_hi}] & \N{rt_P1A1_on_EPDMS_all} [\N{rt_P1A1_on_EPDMS_all_lo}, \N{rt_P1A1_on_EPDMS_all_hi}] & \N{rt_P2A2_on_EPDMS_all} [\N{rt_P2A2_on_EPDMS_all_lo}, \N{rt_P2A2_on_EPDMS_all_hi}] & O/U \\
P--A & EPDMS & ON & F & \N{rt_P0A0_on_EPDMS_human_fail} [\N{rt_P0A0_on_EPDMS_human_fail_lo}, \N{rt_P0A0_on_EPDMS_human_fail_hi}] & \N{rt_P1A1_on_EPDMS_human_fail} [\N{rt_P1A1_on_EPDMS_human_fail_lo}, \N{rt_P1A1_on_EPDMS_human_fail_hi}] & \N{rt_P2A2_on_EPDMS_human_fail} [\N{rt_P2A2_on_EPDMS_human_fail_lo}, \N{rt_P2A2_on_EPDMS_human_fail_hi}] & U/U \\
P--A & EPDMS & ON & R & \N{rt_P0A0_on_EPDMS_rest} [\N{rt_P0A0_on_EPDMS_rest_lo}, \N{rt_P0A0_on_EPDMS_rest_hi}] & \N{rt_P1A1_on_EPDMS_rest} [\N{rt_P1A1_on_EPDMS_rest_lo}, \N{rt_P1A1_on_EPDMS_rest_hi}] & \N{rt_P2A2_on_EPDMS_rest} [\N{rt_P2A2_on_EPDMS_rest_lo}, \N{rt_P2A2_on_EPDMS_rest_hi}] & U/U \\
P--A & LK & OFF & All & \N{rt_P0A0_off_LK_all} [\N{rt_P0A0_off_LK_all_lo}, \N{rt_P0A0_off_LK_all_hi}] & \N{rt_P1A1_off_LK_all} [\N{rt_P1A1_off_LK_all_lo}, \N{rt_P1A1_off_LK_all_hi}] & \N{rt_P2A2_off_LK_all} [\N{rt_P2A2_off_LK_all_lo}, \N{rt_P2A2_off_LK_all_hi}] & O/O \\
P--A & LK & OFF & F & \N{rt_P0A0_off_LK_human_fail} [\N{rt_P0A0_off_LK_human_fail_lo}, \N{rt_P0A0_off_LK_human_fail_hi}] & \N{rt_P1A1_off_LK_human_fail} [\N{rt_P1A1_off_LK_human_fail_lo}, \N{rt_P1A1_off_LK_human_fail_hi}] & \N{rt_P2A2_off_LK_human_fail} [\N{rt_P2A2_off_LK_human_fail_lo}, \N{rt_P2A2_off_LK_human_fail_hi}] & O/O \\
P--A & LK & OFF & R & \N{rt_P0A0_off_LK_rest} [\N{rt_P0A0_off_LK_rest_lo}, \N{rt_P0A0_off_LK_rest_hi}] & \N{rt_P1A1_off_LK_rest} [\N{rt_P1A1_off_LK_rest_lo}, \N{rt_P1A1_off_LK_rest_hi}] & \N{rt_P2A2_off_LK_rest} [\N{rt_P2A2_off_LK_rest_lo}, \N{rt_P2A2_off_LK_rest_hi}] & O/O \\
P--A & LK & ON & All & \N{rt_P0A0_on_LK_all} [\N{rt_P0A0_on_LK_all_lo}, \N{rt_P0A0_on_LK_all_hi}] & \N{rt_P1A1_on_LK_all} [\N{rt_P1A1_on_LK_all_lo}, \N{rt_P1A1_on_LK_all_hi}] & \N{rt_P2A2_on_LK_all} [\N{rt_P2A2_on_LK_all_lo}, \N{rt_P2A2_on_LK_all_hi}] & O/O \\
P--A & LK & ON & F & \N{rt_P0A0_on_LK_human_fail} [\N{rt_P0A0_on_LK_human_fail_lo}, \N{rt_P0A0_on_LK_human_fail_hi}] & \N{rt_P1A1_on_LK_human_fail} [\N{rt_P1A1_on_LK_human_fail_lo}, \N{rt_P1A1_on_LK_human_fail_hi}] & \N{rt_P2A2_on_LK_human_fail} [\N{rt_P2A2_on_LK_human_fail_lo}, \N{rt_P2A2_on_LK_human_fail_hi}] & O/U \\
P--A & LK & ON & R & \N{rt_P0A0_on_LK_rest} [\N{rt_P0A0_on_LK_rest_lo}, \N{rt_P0A0_on_LK_rest_hi}] & \N{rt_P1A1_on_LK_rest} [\N{rt_P1A1_on_LK_rest_lo}, \N{rt_P1A1_on_LK_rest_hi}] & \N{rt_P2A2_on_LK_rest} [\N{rt_P2A2_on_LK_rest_lo}, \N{rt_P2A2_on_LK_rest_hi}] & O/O \\
P--A & EC & Both & All & \N{rt_P0A0_off_EC_all} [\N{rt_P0A0_off_EC_all_lo}, \N{rt_P0A0_off_EC_all_hi}] & \N{rt_P1A1_off_EC_all} [\N{rt_P1A1_off_EC_all_lo}, \N{rt_P1A1_off_EC_all_hi}] & \N{rt_P2A2_off_EC_all} [\N{rt_P2A2_off_EC_all_lo}, \N{rt_P2A2_off_EC_all_hi}] & O/O \\
P--A & EC & Both & F & \N{rt_P0A0_off_EC_human_fail} [\N{rt_P0A0_off_EC_human_fail_lo}, \N{rt_P0A0_off_EC_human_fail_hi}] & \N{rt_P1A1_off_EC_human_fail} [\N{rt_P1A1_off_EC_human_fail_lo}, \N{rt_P1A1_off_EC_human_fail_hi}] & \N{rt_P2A2_off_EC_human_fail} [\N{rt_P2A2_off_EC_human_fail_lo}, \N{rt_P2A2_off_EC_human_fail_hi}] & O/O \\
P--A & EC & Both & R & \N{rt_P0A0_off_EC_rest} [\N{rt_P0A0_off_EC_rest_lo}, \N{rt_P0A0_off_EC_rest_hi}] & \N{rt_P1A1_off_EC_rest} [\N{rt_P1A1_off_EC_rest_lo}, \N{rt_P1A1_off_EC_rest_hi}] & \N{rt_P2A2_off_EC_rest} [\N{rt_P2A2_off_EC_rest_lo}, \N{rt_P2A2_off_EC_rest_hi}] & O/O \\
\midrule
M--A & EPDMS & OFF & All & \N{rt_M0A0_off_EPDMS_all} [\N{rt_M0A0_off_EPDMS_all_lo}, \N{rt_M0A0_off_EPDMS_all_hi}] & \N{rt_M1A1_off_EPDMS_all} [\N{rt_M1A1_off_EPDMS_all_lo}, \N{rt_M1A1_off_EPDMS_all_hi}] & \N{rt_M2A2_off_EPDMS_all} [\N{rt_M2A2_off_EPDMS_all_lo}, \N{rt_M2A2_off_EPDMS_all_hi}] & U/U \\
M--A & EPDMS & OFF & F & \N{rt_M0A0_off_EPDMS_human_fail} [\N{rt_M0A0_off_EPDMS_human_fail_lo}, \N{rt_M0A0_off_EPDMS_human_fail_hi}] & \N{rt_M1A1_off_EPDMS_human_fail} [\N{rt_M1A1_off_EPDMS_human_fail_lo}, \N{rt_M1A1_off_EPDMS_human_fail_hi}] & \N{rt_M2A2_off_EPDMS_human_fail} [\N{rt_M2A2_off_EPDMS_human_fail_lo}, \N{rt_M2A2_off_EPDMS_human_fail_hi}] & U/U \\
M--A & EPDMS & OFF & R & \N{rt_M0A0_off_EPDMS_rest} [\N{rt_M0A0_off_EPDMS_rest_lo}, \N{rt_M0A0_off_EPDMS_rest_hi}] & \N{rt_M1A1_off_EPDMS_rest} [\N{rt_M1A1_off_EPDMS_rest_lo}, \N{rt_M1A1_off_EPDMS_rest_hi}] & \N{rt_M2A2_off_EPDMS_rest} [\N{rt_M2A2_off_EPDMS_rest_lo}, \N{rt_M2A2_off_EPDMS_rest_hi}] & U/U \\
M--A & EPDMS & ON & All & \N{rt_M0A0_on_EPDMS_all} [\N{rt_M0A0_on_EPDMS_all_lo}, \N{rt_M0A0_on_EPDMS_all_hi}] & \N{rt_M1A1_on_EPDMS_all} [\N{rt_M1A1_on_EPDMS_all_lo}, \N{rt_M1A1_on_EPDMS_all_hi}] & \N{rt_M2A2_on_EPDMS_all} [\N{rt_M2A2_on_EPDMS_all_lo}, \N{rt_M2A2_on_EPDMS_all_hi}] & U/U \\
M--A & EPDMS & ON & F & \N{rt_M0A0_on_EPDMS_human_fail} [\N{rt_M0A0_on_EPDMS_human_fail_lo}, \N{rt_M0A0_on_EPDMS_human_fail_hi}] & \N{rt_M1A1_on_EPDMS_human_fail} [\N{rt_M1A1_on_EPDMS_human_fail_lo}, \N{rt_M1A1_on_EPDMS_human_fail_hi}] & \N{rt_M2A2_on_EPDMS_human_fail} [\N{rt_M2A2_on_EPDMS_human_fail_lo}, \N{rt_M2A2_on_EPDMS_human_fail_hi}] & U/U \\
M--A & EPDMS & ON & R & \N{rt_M0A0_on_EPDMS_rest} [\N{rt_M0A0_on_EPDMS_rest_lo}, \N{rt_M0A0_on_EPDMS_rest_hi}] & \N{rt_M1A1_on_EPDMS_rest} [\N{rt_M1A1_on_EPDMS_rest_lo}, \N{rt_M1A1_on_EPDMS_rest_hi}] & \N{rt_M2A2_on_EPDMS_rest} [\N{rt_M2A2_on_EPDMS_rest_lo}, \N{rt_M2A2_on_EPDMS_rest_hi}] & U/U \\
M--A & LK & OFF & All & \N{rt_M0A0_off_LK_all} [\N{rt_M0A0_off_LK_all_lo}, \N{rt_M0A0_off_LK_all_hi}] & \N{rt_M1A1_off_LK_all} [\N{rt_M1A1_off_LK_all_lo}, \N{rt_M1A1_off_LK_all_hi}] & \N{rt_M2A2_off_LK_all} [\N{rt_M2A2_off_LK_all_lo}, \N{rt_M2A2_off_LK_all_hi}] & O/O \\
M--A & LK & OFF & F & \N{rt_M0A0_off_LK_human_fail} [\N{rt_M0A0_off_LK_human_fail_lo}, \N{rt_M0A0_off_LK_human_fail_hi}] & \N{rt_M1A1_off_LK_human_fail} [\N{rt_M1A1_off_LK_human_fail_lo}, \N{rt_M1A1_off_LK_human_fail_hi}] & \N{rt_M2A2_off_LK_human_fail} [\N{rt_M2A2_off_LK_human_fail_lo}, \N{rt_M2A2_off_LK_human_fail_hi}] & U/U \\
M--A & LK & OFF & R & \N{rt_M0A0_off_LK_rest} [\N{rt_M0A0_off_LK_rest_lo}, \N{rt_M0A0_off_LK_rest_hi}] & \N{rt_M1A1_off_LK_rest} [\N{rt_M1A1_off_LK_rest_lo}, \N{rt_M1A1_off_LK_rest_hi}] & \N{rt_M2A2_off_LK_rest} [\N{rt_M2A2_off_LK_rest_lo}, \N{rt_M2A2_off_LK_rest_hi}] & U/U \\
M--A & LK & ON & All & \N{rt_M0A0_on_LK_all} [\N{rt_M0A0_on_LK_all_lo}, \N{rt_M0A0_on_LK_all_hi}] & \N{rt_M1A1_on_LK_all} [\N{rt_M1A1_on_LK_all_lo}, \N{rt_M1A1_on_LK_all_hi}] & \N{rt_M2A2_on_LK_all} [\N{rt_M2A2_on_LK_all_lo}, \N{rt_M2A2_on_LK_all_hi}] & S/U \\
M--A & LK & ON & F & \N{rt_M0A0_on_LK_human_fail} [\N{rt_M0A0_on_LK_human_fail_lo}, \N{rt_M0A0_on_LK_human_fail_hi}] & \N{rt_M1A1_on_LK_human_fail} [\N{rt_M1A1_on_LK_human_fail_lo}, \N{rt_M1A1_on_LK_human_fail_hi}] & \N{rt_M2A2_on_LK_human_fail} [\N{rt_M2A2_on_LK_human_fail_lo}, \N{rt_M2A2_on_LK_human_fail_hi}] & U/U \\
M--A & LK & ON & R & \N{rt_M0A0_on_LK_rest} [\N{rt_M0A0_on_LK_rest_lo}, \N{rt_M0A0_on_LK_rest_hi}] & \N{rt_M1A1_on_LK_rest} [\N{rt_M1A1_on_LK_rest_lo}, \N{rt_M1A1_on_LK_rest_hi}] & \N{rt_M2A2_on_LK_rest} [\N{rt_M2A2_on_LK_rest_lo}, \N{rt_M2A2_on_LK_rest_hi}] & U/U \\
M--A & EC & Both & All & \N{rt_M0A0_off_EC_all} [\N{rt_M0A0_off_EC_all_lo}, \N{rt_M0A0_off_EC_all_hi}] & \N{rt_M1A1_off_EC_all} [\N{rt_M1A1_off_EC_all_lo}, \N{rt_M1A1_off_EC_all_hi}] & \N{rt_M2A2_off_EC_all} [\N{rt_M2A2_off_EC_all_lo}, \N{rt_M2A2_off_EC_all_hi}] & U/U \\
M--A & EC & Both & F & \N{rt_M0A0_off_EC_human_fail} [\N{rt_M0A0_off_EC_human_fail_lo}, \N{rt_M0A0_off_EC_human_fail_hi}] & \N{rt_M1A1_off_EC_human_fail} [\N{rt_M1A1_off_EC_human_fail_lo}, \N{rt_M1A1_off_EC_human_fail_hi}] & \N{rt_M2A2_off_EC_human_fail} [\N{rt_M2A2_off_EC_human_fail_lo}, \N{rt_M2A2_off_EC_human_fail_hi}] & U/U \\
M--A & EC & Both & R & \N{rt_M0A0_off_EC_rest} [\N{rt_M0A0_off_EC_rest_lo}, \N{rt_M0A0_off_EC_rest_hi}] & \N{rt_M1A1_off_EC_rest} [\N{rt_M1A1_off_EC_rest_lo}, \N{rt_M1A1_off_EC_rest_hi}] & \N{rt_M2A2_off_EC_rest} [\N{rt_M2A2_off_EC_rest_lo}, \N{rt_M2A2_off_EC_rest_hi}] & U/U \\
\bottomrule
\end{tabular}
\end{table}

\noindent The positive control's OFF LK interval is below zero and its OFF EPDMS interval spans zero; continuation under A has an EPDMS interval spanning zero under both switches and a positive EC interval (Table~\ref{tab:rt-absolute}).

{\scriptsize\noindent All-token seed scales (two/three; column order of Table~\ref{tab:rt-absolute}): \N{rt_spread_two_seed_off_EPDMS_all}/\N{rt_spread_three_seed_off_EPDMS_all}; \N{rt_spread_two_seed_on_EPDMS_all}/\N{rt_spread_three_seed_on_EPDMS_all}; \N{rt_spread_two_seed_off_LK_all}/\N{rt_spread_three_seed_off_LK_all}; \N{rt_spread_two_seed_on_LK_all}/\N{rt_spread_three_seed_on_LK_all}; \N{rt_spread_two_seed_off_EC_all}/\N{rt_spread_three_seed_off_EC_all}.}

\paragraph{Section~\ref{sec:probe}'s probe on the continued models.}
Table~\ref{tab:rt-probe} reruns the probe of Section~\ref{sec:probe} on the step-\N{rt_steps} checkpoints over the same scenes and matched pairs: R1 is the lane-keeping head's median AUC on the \N{LK_mixed} forgiven scenes, R2 the same on the \N{probe_pairs} matched unforgiven scenes and R3 Section~\ref{sec:probe}'s primary contrast $D_{\text{forgiven}}-D_{\text{matched}}$, with verdicts by the rule above and L0--A0 read against both seed scales.
On the matched unforgiven scenes every A, M and P model lies between \N{prt_AMP_R2_min} and \N{prt_AMP_R2_max}, against \N{probe_matched_lk_auc} for the released model.

\begin{table}[htbp]
\centering\scriptsize
\caption{The probe of Section~\ref{sec:probe} on the continued models: point estimates of R1, R2 and R3, and two-/three-seed verdicts.}
\label{tab:rt-probe}
\begin{tabular}{@{}lccc@{}}
\toprule
Model & R1 & R2 & R3 \\
\midrule
Released & \N{probe_T_lk_auc} & \N{probe_matched_lk_auc} & \N{probe_did_match} \\
A$_{\N{rt_seed0}}$ & \N{prt_A0_R1} & \N{prt_A0_R2} & \N{prt_A0_R3} \\
M$_{\N{rt_seed0}}$ & \N{prt_M0_R1} & \N{prt_M0_R2} & \N{prt_M0_R3} \\
P$_{\N{rt_seed0}}$ & \N{prt_P0_R1} & \N{prt_P0_R2} & \N{prt_P0_R3} \\
A$_{\N{rt_seed1}}$ & \N{prt_A1_R1} & \N{prt_A1_R2} & \N{prt_A1_R3} \\
M$_{\N{rt_seed1}}$ & \N{prt_M1_R1} & \N{prt_M1_R2} & \N{prt_M1_R3} \\
P$_{\N{rt_seed1}}$ & \N{prt_P1_R1} & \N{prt_P1_R2} & \N{prt_P1_R3} \\
A$_{\N{rt_seed2}}$ & \N{prt_A2_R1} & \N{prt_A2_R2} & \N{prt_A2_R3} \\
M$_{\N{rt_seed2}}$ & \N{prt_M2_R1} & \N{prt_M2_R2} & \N{prt_M2_R3} \\
P$_{\N{rt_seed2}}$ & \N{prt_P2_R1} & \N{prt_P2_R2} & \N{prt_P2_R3} \\
L$_{\N{rt_seed0}}$ & \N{prt_L0_R1} & \N{prt_L0_R2} & \N{prt_L0_R3} \\
\midrule
P--A & O/O & O/U & O/O \\
M--A & O/O & S/S & O/O \\
L0--A0 & O/O & O/O & O/O \\
\bottomrule
\end{tabular}
\end{table}

\paragraph{Section~\ref{sec:probe}'s probe on held-out navtest.}
Table~\ref{tab:rt-navtest} repeats the probe on the \N{nt_T} navtest scenes whose human lane-keeping score is zero and whose lane-keeping column is mixed, and on \N{nt_pairs} matched pairs, with R4 the matched minus the forgiven median over the pairs, the sign opposite to Section~\ref{sec:probe}'s difference of medians.
P--A is observed on every reading and P's R3 and R4 stay above zero in every seed; M--A is observed only on R4, and L0's R1 and R2 lie below 0.5.

\begin{table}[htbp]
\centering\scriptsize
\caption{The probe of Section~\ref{sec:probe} on held-out navtest scenes: point estimates of R1 to R4 and two-/three-seed verdicts.}
\label{tab:rt-navtest}
\begin{tabular}{@{}lcccc@{}}
\toprule
Model & R1 & R2 & R3 & R4 \\
\midrule
Released & \N{nt_released_R1} & \N{nt_released_R2} & \N{nt_released_R3} & \N{nt_released_R4} \\
A$_{\N{rt_seed0}}$ & \N{nt_A0_R1} & \N{nt_A0_R2} & \N{nt_A0_R3} & \N{nt_A0_R4} \\
M$_{\N{rt_seed0}}$ & \N{nt_M0_R1} & \N{nt_M0_R2} & \N{nt_M0_R3} & \N{nt_M0_R4} \\
P$_{\N{rt_seed0}}$ & \N{nt_P0_R1} & \N{nt_P0_R2} & \N{nt_P0_R3} & \N{nt_P0_R4} \\
A$_{\N{rt_seed1}}$ & \N{nt_A1_R1} & \N{nt_A1_R2} & \N{nt_A1_R3} & \N{nt_A1_R4} \\
M$_{\N{rt_seed1}}$ & \N{nt_M1_R1} & \N{nt_M1_R2} & \N{nt_M1_R3} & \N{nt_M1_R4} \\
P$_{\N{rt_seed1}}$ & \N{nt_P1_R1} & \N{nt_P1_R2} & \N{nt_P1_R3} & \N{nt_P1_R4} \\
A$_{\N{rt_seed2}}$ & \N{nt_A2_R1} & \N{nt_A2_R2} & \N{nt_A2_R3} & \N{nt_A2_R4} \\
M$_{\N{rt_seed2}}$ & \N{nt_M2_R1} & \N{nt_M2_R2} & \N{nt_M2_R3} & \N{nt_M2_R4} \\
P$_{\N{rt_seed2}}$ & \N{nt_P2_R1} & \N{nt_P2_R2} & \N{nt_P2_R3} & \N{nt_P2_R4} \\
L$_{\N{rt_seed0}}$ & \N{nt_L0_R1} & \N{nt_L0_R2} & \N{nt_L0_R3} & \N{nt_L0_R4} \\
\midrule
P--A & O/O & O/O & O/O & O/O \\
M--A & U/U & S/S & S/S & O/O \\
L0--A0 & O/O & O/O & O/O & O/O \\
\bottomrule
\end{tabular}
\end{table}

\paragraph{Distance to the human trajectory.}
Table~\ref{tab:rt-distance} compares each model's navtest plans with the human trajectory at eight shared times, with the pairing, intervals and rule above.

\begin{table}[htbp]
\centering\scriptsize
\caption{Planned-to-human distance in meters, P$_s$ minus A$_s$ on F, on R and their difference: point estimates [intervals] and two-/three-seed verdicts. ADE averages the eight aligned poses, FDE takes the last, and the lateral deviation is the largest distance from a planned pose to the human path.}
\label{tab:rt-distance}
\setlength{\tabcolsep}{3pt}
\begin{tabular}{@{}llcccc@{}}
\toprule
Metric & Set & $s=\N{rt_seed0}$ & $s=\N{rt_seed1}$ & $s=\N{rt_seed2}$ & Verdict \\
\midrule
ADE & F & \N{rt_dist_P0A0_ADE_F} [\N{rt_dist_P0A0_ADE_F_lo}, \N{rt_dist_P0A0_ADE_F_hi}] & \N{rt_dist_P1A1_ADE_F} [\N{rt_dist_P1A1_ADE_F_lo}, \N{rt_dist_P1A1_ADE_F_hi}] & \N{rt_dist_P2A2_ADE_F} [\N{rt_dist_P2A2_ADE_F_lo}, \N{rt_dist_P2A2_ADE_F_hi}] & O/O \\
ADE & R & \N{rt_dist_P0A0_ADE_R} [\N{rt_dist_P0A0_ADE_R_lo}, \N{rt_dist_P0A0_ADE_R_hi}] & \N{rt_dist_P1A1_ADE_R} [\N{rt_dist_P1A1_ADE_R_lo}, \N{rt_dist_P1A1_ADE_R_hi}] & \N{rt_dist_P2A2_ADE_R} [\N{rt_dist_P2A2_ADE_R_lo}, \N{rt_dist_P2A2_ADE_R_hi}] & U/U \\
ADE & F$-$R & \N{rt_dist_P0A0_ADE_FR} [\N{rt_dist_P0A0_ADE_FR_lo}, \N{rt_dist_P0A0_ADE_FR_hi}] & \N{rt_dist_P1A1_ADE_FR} [\N{rt_dist_P1A1_ADE_FR_lo}, \N{rt_dist_P1A1_ADE_FR_hi}] & \N{rt_dist_P2A2_ADE_FR} [\N{rt_dist_P2A2_ADE_FR_lo}, \N{rt_dist_P2A2_ADE_FR_hi}] & O/O \\
FDE & F & \N{rt_dist_P0A0_FDE_F} [\N{rt_dist_P0A0_FDE_F_lo}, \N{rt_dist_P0A0_FDE_F_hi}] & \N{rt_dist_P1A1_FDE_F} [\N{rt_dist_P1A1_FDE_F_lo}, \N{rt_dist_P1A1_FDE_F_hi}] & \N{rt_dist_P2A2_FDE_F} [\N{rt_dist_P2A2_FDE_F_lo}, \N{rt_dist_P2A2_FDE_F_hi}] & O/O \\
FDE & R & \N{rt_dist_P0A0_FDE_R} [\N{rt_dist_P0A0_FDE_R_lo}, \N{rt_dist_P0A0_FDE_R_hi}] & \N{rt_dist_P1A1_FDE_R} [\N{rt_dist_P1A1_FDE_R_lo}, \N{rt_dist_P1A1_FDE_R_hi}] & \N{rt_dist_P2A2_FDE_R} [\N{rt_dist_P2A2_FDE_R_lo}, \N{rt_dist_P2A2_FDE_R_hi}] & U/U \\
FDE & F$-$R & \N{rt_dist_P0A0_FDE_FR} [\N{rt_dist_P0A0_FDE_FR_lo}, \N{rt_dist_P0A0_FDE_FR_hi}] & \N{rt_dist_P1A1_FDE_FR} [\N{rt_dist_P1A1_FDE_FR_lo}, \N{rt_dist_P1A1_FDE_FR_hi}] & \N{rt_dist_P2A2_FDE_FR} [\N{rt_dist_P2A2_FDE_FR_lo}, \N{rt_dist_P2A2_FDE_FR_hi}] & O/U \\
Lateral & F & \N{rt_dist_P0A0_MAXLAT_F} [\N{rt_dist_P0A0_MAXLAT_F_lo}, \N{rt_dist_P0A0_MAXLAT_F_hi}] & \N{rt_dist_P1A1_MAXLAT_F} [\N{rt_dist_P1A1_MAXLAT_F_lo}, \N{rt_dist_P1A1_MAXLAT_F_hi}] & \N{rt_dist_P2A2_MAXLAT_F} [\N{rt_dist_P2A2_MAXLAT_F_lo}, \N{rt_dist_P2A2_MAXLAT_F_hi}] & O/U \\
Lateral & R & \N{rt_dist_P0A0_MAXLAT_R} [\N{rt_dist_P0A0_MAXLAT_R_lo}, \N{rt_dist_P0A0_MAXLAT_R_hi}] & \N{rt_dist_P1A1_MAXLAT_R} [\N{rt_dist_P1A1_MAXLAT_R_lo}, \N{rt_dist_P1A1_MAXLAT_R_hi}] & \N{rt_dist_P2A2_MAXLAT_R} [\N{rt_dist_P2A2_MAXLAT_R_lo}, \N{rt_dist_P2A2_MAXLAT_R_hi}] & U/U \\
Lateral & F$-$R & \N{rt_dist_P0A0_MAXLAT_FR} [\N{rt_dist_P0A0_MAXLAT_FR_lo}, \N{rt_dist_P0A0_MAXLAT_FR_hi}] & \N{rt_dist_P1A1_MAXLAT_FR} [\N{rt_dist_P1A1_MAXLAT_FR_lo}, \N{rt_dist_P1A1_MAXLAT_FR_hi}] & \N{rt_dist_P2A2_MAXLAT_FR} [\N{rt_dist_P2A2_MAXLAT_FR_lo}, \N{rt_dist_P2A2_MAXLAT_FR_hi}] & O/O \\
\bottomrule
\end{tabular}
\end{table}

\paragraph{Step \N{rt530_steps}.}
Table~\ref{tab:rt-s530} reads the same ten runs at optimizer step \N{rt530_steps} with the pairing, intervals and rule above; L0--A0's EPDMS intervals lie below zero under both switches.

\begin{table}[htbp]
\centering\scriptsize
\setlength{\tabcolsep}{3pt}
\caption{Step \N{rt530_steps}: P$_s$ and M$_s$ minus A$_s$ on all tokens, F and R, and the interactions $I^X_s$ (F$-$R), as point estimates with two-/three-seed verdicts (V); the last rows give L0--A0 [intervals].}
\label{tab:rt-s530}
\begin{tabular}{@{}lll rrrc rrrc@{}}
\toprule
& & & \multicolumn{4}{c}{P--A} & \multicolumn{4}{c}{M--A} \\
\cmidrule(lr){4-7}\cmidrule(l){8-11}
Metric & Mode & Set & $s=\N{rt_seed0}$ & $s=\N{rt_seed1}$ & $s=\N{rt_seed2}$ & V & $s=\N{rt_seed0}$ & $s=\N{rt_seed1}$ & $s=\N{rt_seed2}$ & V \\
\midrule
EPDMS & OFF & All & \N{rt530_P0A0_off_EPDMS_all} & \N{rt530_P1A1_off_EPDMS_all} & \N{rt530_P2A2_off_EPDMS_all} & U/U & \N{rt530_M0A0_off_EPDMS_all} & \N{rt530_M1A1_off_EPDMS_all} & \N{rt530_M2A2_off_EPDMS_all} & U/U \\
 & & F & \N{rt530_P0A0_off_EPDMS_human_fail} & \N{rt530_P1A1_off_EPDMS_human_fail} & \N{rt530_P2A2_off_EPDMS_human_fail} & U/U & \N{rt530_M0A0_off_EPDMS_human_fail} & \N{rt530_M1A1_off_EPDMS_human_fail} & \N{rt530_M2A2_off_EPDMS_human_fail} & U/U \\
 & & R & \N{rt530_P0A0_off_EPDMS_rest} & \N{rt530_P1A1_off_EPDMS_rest} & \N{rt530_P2A2_off_EPDMS_rest} & U/U & \N{rt530_M0A0_off_EPDMS_rest} & \N{rt530_M1A1_off_EPDMS_rest} & \N{rt530_M2A2_off_EPDMS_rest} & U/U \\
 & & F$-$R & \N{rt530_IP0_off_EPDMS} & \N{rt530_IP1_off_EPDMS} & \N{rt530_IP2_off_EPDMS} & U/U & \N{rt530_IM0_off_EPDMS} & \N{rt530_IM1_off_EPDMS} & \N{rt530_IM2_off_EPDMS} & U/U \\
EPDMS & ON & All & \N{rt530_P0A0_on_EPDMS_all} & \N{rt530_P1A1_on_EPDMS_all} & \N{rt530_P2A2_on_EPDMS_all} & U/U & \N{rt530_M0A0_on_EPDMS_all} & \N{rt530_M1A1_on_EPDMS_all} & \N{rt530_M2A2_on_EPDMS_all} & U/U \\
 & & F & \N{rt530_P0A0_on_EPDMS_human_fail} & \N{rt530_P1A1_on_EPDMS_human_fail} & \N{rt530_P2A2_on_EPDMS_human_fail} & U/U & \N{rt530_M0A0_on_EPDMS_human_fail} & \N{rt530_M1A1_on_EPDMS_human_fail} & \N{rt530_M2A2_on_EPDMS_human_fail} & U/U \\
 & & R & \N{rt530_P0A0_on_EPDMS_rest} & \N{rt530_P1A1_on_EPDMS_rest} & \N{rt530_P2A2_on_EPDMS_rest} & U/U & \N{rt530_M0A0_on_EPDMS_rest} & \N{rt530_M1A1_on_EPDMS_rest} & \N{rt530_M2A2_on_EPDMS_rest} & U/U \\
 & & F$-$R & \N{rt530_IP0_on_EPDMS} & \N{rt530_IP1_on_EPDMS} & \N{rt530_IP2_on_EPDMS} & U/U & \N{rt530_IM0_on_EPDMS} & \N{rt530_IM1_on_EPDMS} & \N{rt530_IM2_on_EPDMS} & U/U \\
LK & OFF & All & \N{rt530_P0A0_off_LK_all} & \N{rt530_P1A1_off_LK_all} & \N{rt530_P2A2_off_LK_all} & O/O & \N{rt530_M0A0_off_LK_all} & \N{rt530_M1A1_off_LK_all} & \N{rt530_M2A2_off_LK_all} & U/U \\
 & & F & \N{rt530_P0A0_off_LK_human_fail} & \N{rt530_P1A1_off_LK_human_fail} & \N{rt530_P2A2_off_LK_human_fail} & O/O & \N{rt530_M0A0_off_LK_human_fail} & \N{rt530_M1A1_off_LK_human_fail} & \N{rt530_M2A2_off_LK_human_fail} & U/U \\
 & & R & \N{rt530_P0A0_off_LK_rest} & \N{rt530_P1A1_off_LK_rest} & \N{rt530_P2A2_off_LK_rest} & O/O & \N{rt530_M0A0_off_LK_rest} & \N{rt530_M1A1_off_LK_rest} & \N{rt530_M2A2_off_LK_rest} & U/U \\
 & & F$-$R & \N{rt530_IP0_off_LK} & \N{rt530_IP1_off_LK} & \N{rt530_IP2_off_LK} & O/O & \N{rt530_IM0_off_LK} & \N{rt530_IM1_off_LK} & \N{rt530_IM2_off_LK} & U/U \\
LK & ON & All & \N{rt530_P0A0_on_LK_all} & \N{rt530_P1A1_on_LK_all} & \N{rt530_P2A2_on_LK_all} & O/O & \N{rt530_M0A0_on_LK_all} & \N{rt530_M1A1_on_LK_all} & \N{rt530_M2A2_on_LK_all} & U/U \\
 & & F & \N{rt530_P0A0_on_LK_human_fail} & \N{rt530_P1A1_on_LK_human_fail} & \N{rt530_P2A2_on_LK_human_fail} & U/U & \N{rt530_M0A0_on_LK_human_fail} & \N{rt530_M1A1_on_LK_human_fail} & \N{rt530_M2A2_on_LK_human_fail} & U/U \\
 & & R & \N{rt530_P0A0_on_LK_rest} & \N{rt530_P1A1_on_LK_rest} & \N{rt530_P2A2_on_LK_rest} & O/O & \N{rt530_M0A0_on_LK_rest} & \N{rt530_M1A1_on_LK_rest} & \N{rt530_M2A2_on_LK_rest} & U/U \\
 & & F$-$R & \N{rt530_IP0_on_LK} & \N{rt530_IP1_on_LK} & \N{rt530_IP2_on_LK} & O/O & \N{rt530_IM0_on_LK} & \N{rt530_IM1_on_LK} & \N{rt530_IM2_on_LK} & U/U \\
EC & Both & All & \N{rt530_P0A0_off_EC_all} & \N{rt530_P1A1_off_EC_all} & \N{rt530_P2A2_off_EC_all} & O/O & \N{rt530_M0A0_off_EC_all} & \N{rt530_M1A1_off_EC_all} & \N{rt530_M2A2_off_EC_all} & U/U \\
 & & F & \N{rt530_P0A0_off_EC_human_fail} & \N{rt530_P1A1_off_EC_human_fail} & \N{rt530_P2A2_off_EC_human_fail} & O/O & \N{rt530_M0A0_off_EC_human_fail} & \N{rt530_M1A1_off_EC_human_fail} & \N{rt530_M2A2_off_EC_human_fail} & U/U \\
 & & R & \N{rt530_P0A0_off_EC_rest} & \N{rt530_P1A1_off_EC_rest} & \N{rt530_P2A2_off_EC_rest} & U/U & \N{rt530_M0A0_off_EC_rest} & \N{rt530_M1A1_off_EC_rest} & \N{rt530_M2A2_off_EC_rest} & U/U \\
 & & F$-$R & \N{rt530_IP0_off_EC} & \N{rt530_IP1_off_EC} & \N{rt530_IP2_off_EC} & O/O & \N{rt530_IM0_off_EC} & \N{rt530_IM1_off_EC} & \N{rt530_IM2_off_EC} & U/U \\
\midrule
EPDMS & OFF & L0--A0 & \multicolumn{8}{l}{\N{rt530_L0A0_off_EPDMS} [\N{rt530_L0A0_off_EPDMS_lo}, \N{rt530_L0A0_off_EPDMS_hi}]} \\
EPDMS & ON & L0--A0 & \multicolumn{8}{l}{\N{rt530_L0A0_on_EPDMS} [\N{rt530_L0A0_on_EPDMS_lo}, \N{rt530_L0A0_on_EPDMS_hi}]} \\
LK & OFF & L0--A0 & \multicolumn{8}{l}{\N{rt530_L0A0_off_LK} [\N{rt530_L0A0_off_LK_lo}, \N{rt530_L0A0_off_LK_hi}]} \\
LK & ON & L0--A0 & \multicolumn{8}{l}{\N{rt530_L0A0_on_LK} [\N{rt530_L0A0_on_LK_lo}, \N{rt530_L0A0_on_LK_hi}]} \\
EC & Both & L0--A0 & \multicolumn{8}{l}{\N{rt530_L0A0_off_EC} [\N{rt530_L0A0_off_EC_lo}, \N{rt530_L0A0_off_EC_hi}]} \\
\bottomrule
\end{tabular}
\end{table}

\paragraph{The probes at step \N{rt530_steps}.}
Table~\ref{tab:rt-s530-probe} repeats both probes on the same scenes and pairs at step \N{rt530_steps}, and P--A keeps its step-\N{rt_steps} sign and verdict on R1 and R3 of both.
On navtest, P's R1 exceeds A's by \N{p530nt_PA_R1_min}--\N{p530nt_PA_R1_max}, and P's R3 is \N{p530nt_P_R3_min}--\N{p530nt_P_R3_max} against \N{p530nt_A_R3_min}--\N{p530nt_A_R3_max} for A.

\begin{table}[htbp]
\centering\scriptsize
\caption{The probes of Section~\ref{sec:probe} at step \N{rt530_steps} for the A, M and P models: point estimates on navtrain (R1 to R3) and on held-out navtest (R1 to R4), and two-/three-seed verdicts.}
\label{tab:rt-s530-probe}
\begin{tabular}{@{}lccccccc@{}}
\toprule
& \multicolumn{3}{c}{navtrain} & \multicolumn{4}{c}{navtest} \\
\cmidrule(lr){2-4}\cmidrule(l){5-8}
Model & R1 & R2 & R3 & R1 & R2 & R3 & R4 \\
\midrule
A$_{\N{rt_seed0}}$ & \N{p530tr_A0_R1} & \N{p530tr_A0_R2} & \N{p530tr_A0_R3} & \N{p530nt_A0_R1} & \N{p530nt_A0_R2} & \N{p530nt_A0_R3} & \N{p530nt_A0_R4} \\
M$_{\N{rt_seed0}}$ & \N{p530tr_M0_R1} & \N{p530tr_M0_R2} & \N{p530tr_M0_R3} & \N{p530nt_M0_R1} & \N{p530nt_M0_R2} & \N{p530nt_M0_R3} & \N{p530nt_M0_R4} \\
P$_{\N{rt_seed0}}$ & \N{p530tr_P0_R1} & \N{p530tr_P0_R2} & \N{p530tr_P0_R3} & \N{p530nt_P0_R1} & \N{p530nt_P0_R2} & \N{p530nt_P0_R3} & \N{p530nt_P0_R4} \\
A$_{\N{rt_seed1}}$ & \N{p530tr_A1_R1} & \N{p530tr_A1_R2} & \N{p530tr_A1_R3} & \N{p530nt_A1_R1} & \N{p530nt_A1_R2} & \N{p530nt_A1_R3} & \N{p530nt_A1_R4} \\
M$_{\N{rt_seed1}}$ & \N{p530tr_M1_R1} & \N{p530tr_M1_R2} & \N{p530tr_M1_R3} & \N{p530nt_M1_R1} & \N{p530nt_M1_R2} & \N{p530nt_M1_R3} & \N{p530nt_M1_R4} \\
P$_{\N{rt_seed1}}$ & \N{p530tr_P1_R1} & \N{p530tr_P1_R2} & \N{p530tr_P1_R3} & \N{p530nt_P1_R1} & \N{p530nt_P1_R2} & \N{p530nt_P1_R3} & \N{p530nt_P1_R4} \\
A$_{\N{rt_seed2}}$ & \N{p530tr_A2_R1} & \N{p530tr_A2_R2} & \N{p530tr_A2_R3} & \N{p530nt_A2_R1} & \N{p530nt_A2_R2} & \N{p530nt_A2_R3} & \N{p530nt_A2_R4} \\
M$_{\N{rt_seed2}}$ & \N{p530tr_M2_R1} & \N{p530tr_M2_R2} & \N{p530tr_M2_R3} & \N{p530nt_M2_R1} & \N{p530nt_M2_R2} & \N{p530nt_M2_R3} & \N{p530nt_M2_R4} \\
P$_{\N{rt_seed2}}$ & \N{p530tr_P2_R1} & \N{p530tr_P2_R2} & \N{p530tr_P2_R3} & \N{p530nt_P2_R1} & \N{p530nt_P2_R2} & \N{p530nt_P2_R3} & \N{p530nt_P2_R4} \\
\midrule
P--A & O/O & U/U & O/O & O/O & O/O & O/O & O/O \\
M--A & O/O & O/U & O/O & U/U & O/O & U/U & U/U \\
\bottomrule
\end{tabular}
\end{table}

\section{Solver sensitivity of the human census}
\label{app:solver}
The census uses \code{np.linalg.pinv}; \citet[solver matrix in Section~4.2]{scorebasis} report human-reference failure rates of 100.0\%, 12.9\%, and 9.1\% on their own 450-scene interactive pool for stock pseudoinverse, direct solve, and Hermitian pseudoinverse, alongside rollout-divergence rates of 99.8\%, 0\%, and 0\%.
Under the software configuration they audit (supplementary material), substituting \code{scipy.linalg.pinvh} or direct solve leaves all \N{solver_identical} tokens' nine human values bitwise identical, preserves every per-column human-zero set and their union, and retains agreement on all \N{solver_worst_conditioned} worst-conditioned tokens, although the per-token relative matrix difference between stock and Hermitian pseudoinverses reaches $1.159\times10^{-12}$.
Their fallback definition uses deviation from the proposal and single-step distance, while their threshold-ladder description says ``path and step limits'' \citep[Sections~3.2 and~4.3]{scorebasis}; we report both readings.
Under the path-length reading, the 95th-percentile path is 40.67\,m, so at least 5,165 rollouts exceed the strict 20\,m limit, and none exceeds 100\,m, the longest path being 65.85\,m.
Both readings detect their kilometer-scale failure, a median path length of 32.11\,km in the affected configuration of their 32-token diagnostic, whereas the choice of quantity changes the strict-threshold conclusion on our corpus.
Their overwrite rates are measured in the divergent condition their own solver table reports; under the deviation-or-step reading, the \N{div_denominator} stock-solver human rollouts of this split record \N{div_tokens} exceedances at $(20,10)$, $(100,50)$, and $(200,100)$\,m, with every rollout within \N{div_worst_deviation}\,m of its proposal and no single step above \N{div_worst_step}\,m, and forgiveness still erases a distinction in a loss column on \N{headline_percent}\% of the same scenes.

\end{document}